\documentclass[10pt,twocolumn,letterpaper]{article}

\PassOptionsToPackage{nameinlink}{cleveref}

\usepackage{cvpr}              

\definecolor{cvprblue}{rgb}{0.21,0.49,0.74}
\usepackage[pagebackref,breaklinks,colorlinks,allcolors=cvprblue]{hyperref}

\usepackage{shortex-light} 

\usepackage{safecolours}

\colorlet{scBlue}{tol-colour1}
\colorlet{scRed}{tol-colour2}
\colorlet{scGreen}{tol-colour3}
\colorlet{scYellow}{tol-colour4}
\colorlet{scCyan}{tol-colour5}
\colorlet{scPurple}{tol-colour6}

\usepackage{tikz,pgfplots,pgf}
\usetikzlibrary{spy,calc, datavisualization.formats.functions}
\pgfplotsset{compat=1.17}
\usepackage{afterpage}
\usepackage{multirow,array,colortbl}
\usepackage{pifont}

\usetikzlibrary{positioning}
\usepgflibrary{shadings}

\usepackage{xcolor}
\colorlet{highlight}{cvprblue} 

\newcommand{\cmark}{\textcolor{green!50!black}{\ding{51}}}%
\newcommand{\xmark}{\textcolor{black!30}{\ding{55}}}%

\newcolumntype{H}{>{\columncolor{highlight!10!white}}c}

\pgfplotsset{/pgf/number format/1000 sep={}}

\newlength{\figurewidth}
\newlength{\figureheight}

\colorlet{first}{red!50!white}
\colorlet{second}{orange!50!white}
\colorlet{third}{yellow!50!white}
\newcommand{\val}[2]{\cellcolor{#1} $#2$}
\renewcommand{\colorbox}[2]{\protect\tikz[baseline]\protect\node[anchor=base,inner sep=2pt,rounded corners=1pt,fill=#1]{#2};}
\newcommand{\first}[1]{\colorbox{first}{#1}}
\newcommand{\second}[1]{\colorbox{second}{#1}}
\renewcommand{\third}[1]{\colorbox{third}{#1}}

\newcommand{\ours}{DeSplat\xspace}

\renewcommand{\paragraph}[1]{\medskip\noindent\textbf{#1}~~}

\def\paperID{11604} 
\def\confName{CVPR}
\def\confYear{2025}

\title{\ours: Decomposed Gaussian Splatting for Distractor-Free Rendering
}

\author{Yihao Wang$^{1,*}$, Marcus Klasson$^2$, Matias Turkulainen$^2$, Shuzhe Wang$^2$, Juho Kannala$^{2,3}$, Arno Solin$^2$ \\
$^1$Technical University of Munich \quad\quad\quad $^2$Aalto University \quad\quad\quad $^3$University of Oulu \\ 
{\tt\small johanna.wang@tum.de} \quad\quad\quad\quad\quad {\tt\small \{firstname.lastname\}@aalto.fi} \quad\quad\quad\quad
}

\begin{document}

\crefname{table}{Table}{Tables}
\crefname{appendix}{App.}{Apps.}

\twocolumn[{%
\renewcommand\twocolumn[1][]{#1}%
\maketitle
\begin{center}
    \centering
    \captionsetup{type=figure}
    \vspace*{-2em}
    \begin{tikzpicture}
      
      \node[minimum width=\textwidth,minimum height=.2\textwidth,rounded corners=4pt](bg){};
      \node[anchor=west] (left) at (bg.west) {\includegraphics[width=.4\textwidth]{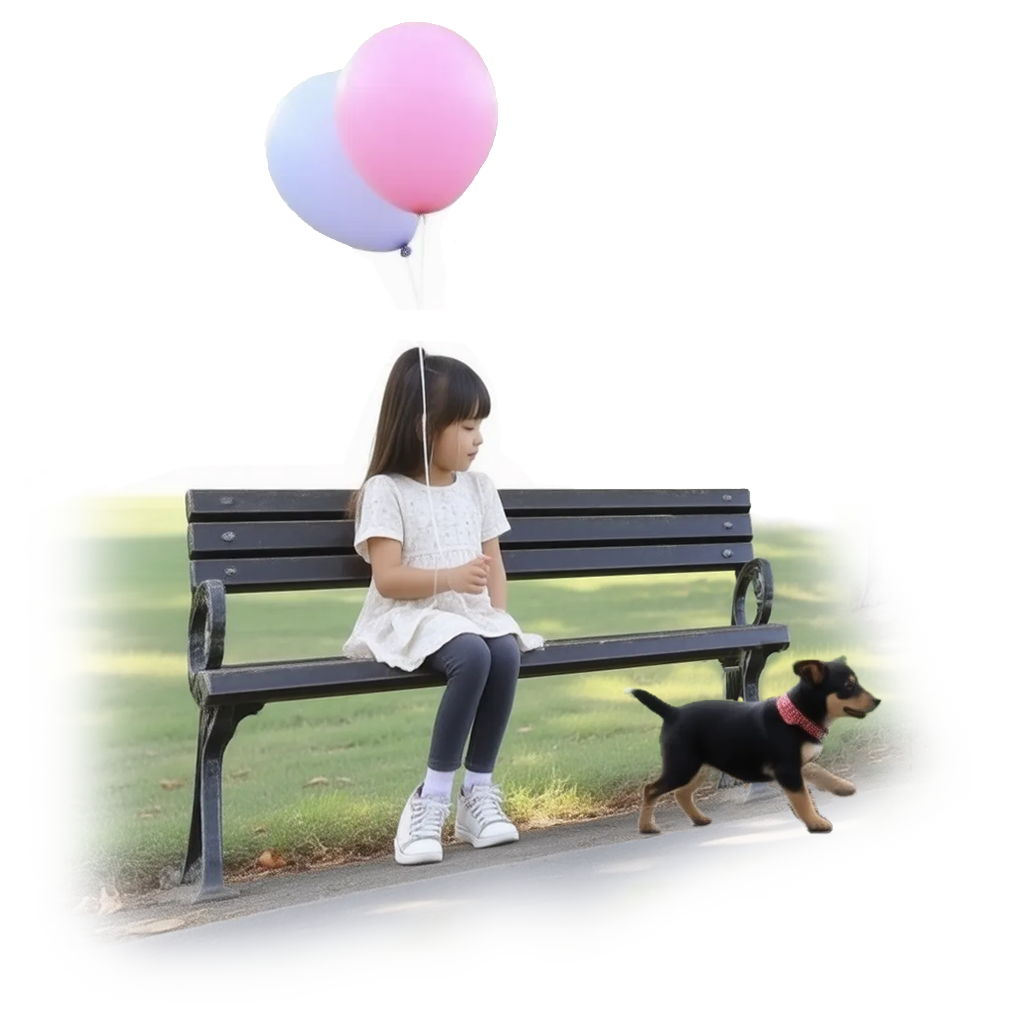}};
      \node[anchor=east] (right) at ($(bg.east)-(.5,0)$) {\includegraphics[width=.4\textwidth]{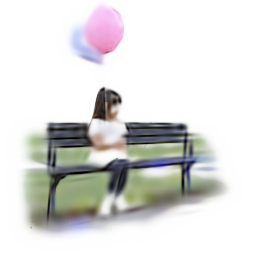}};
      \node at (right) {\includegraphics[width=.4\textwidth]{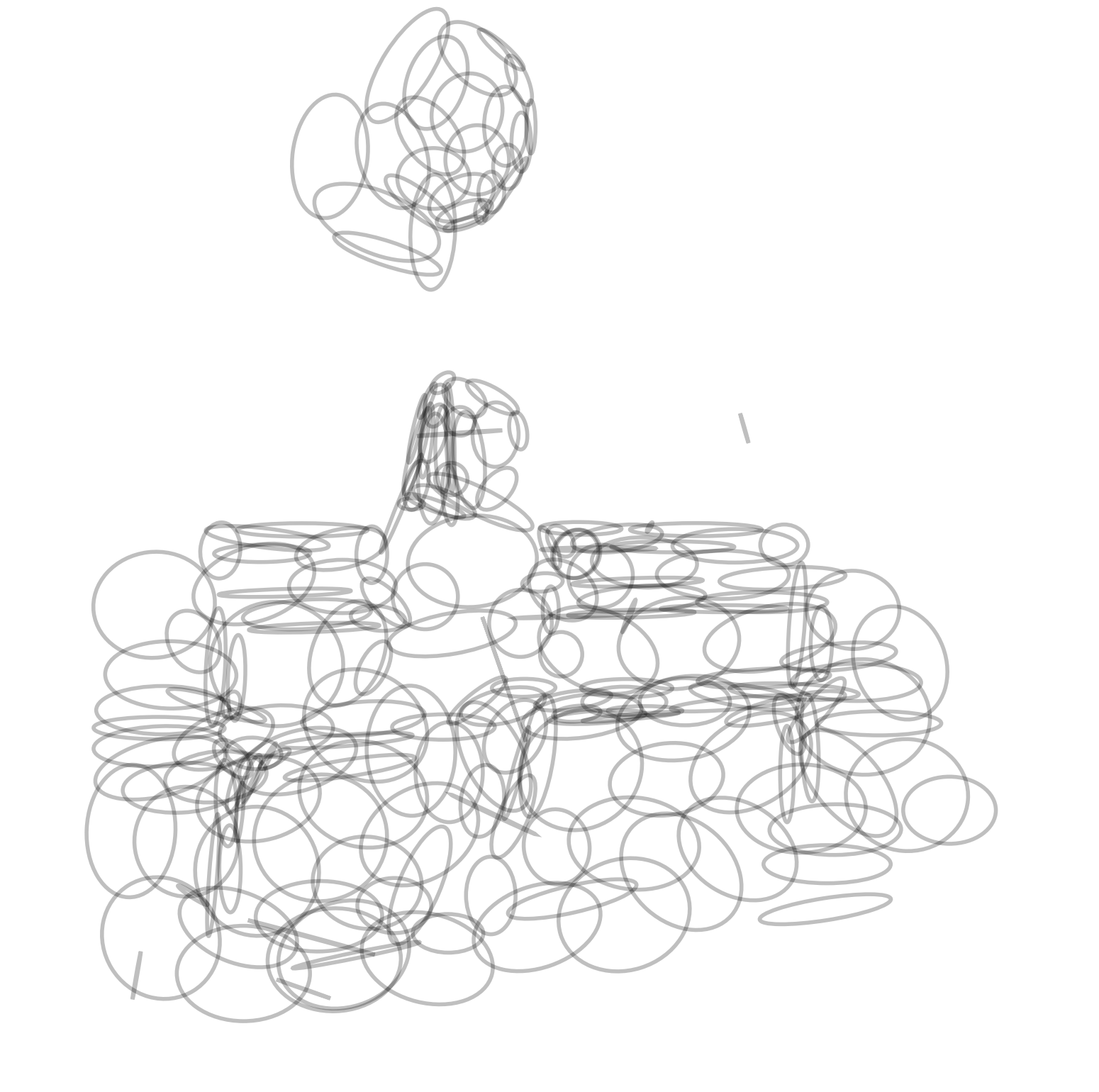}};
      \node at (right) {\includegraphics[width=.4\textwidth]{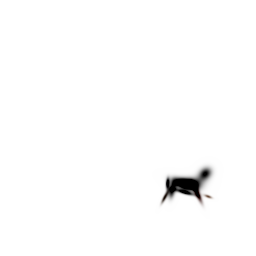}};
      \node at (right) {\includegraphics[width=.4\textwidth]{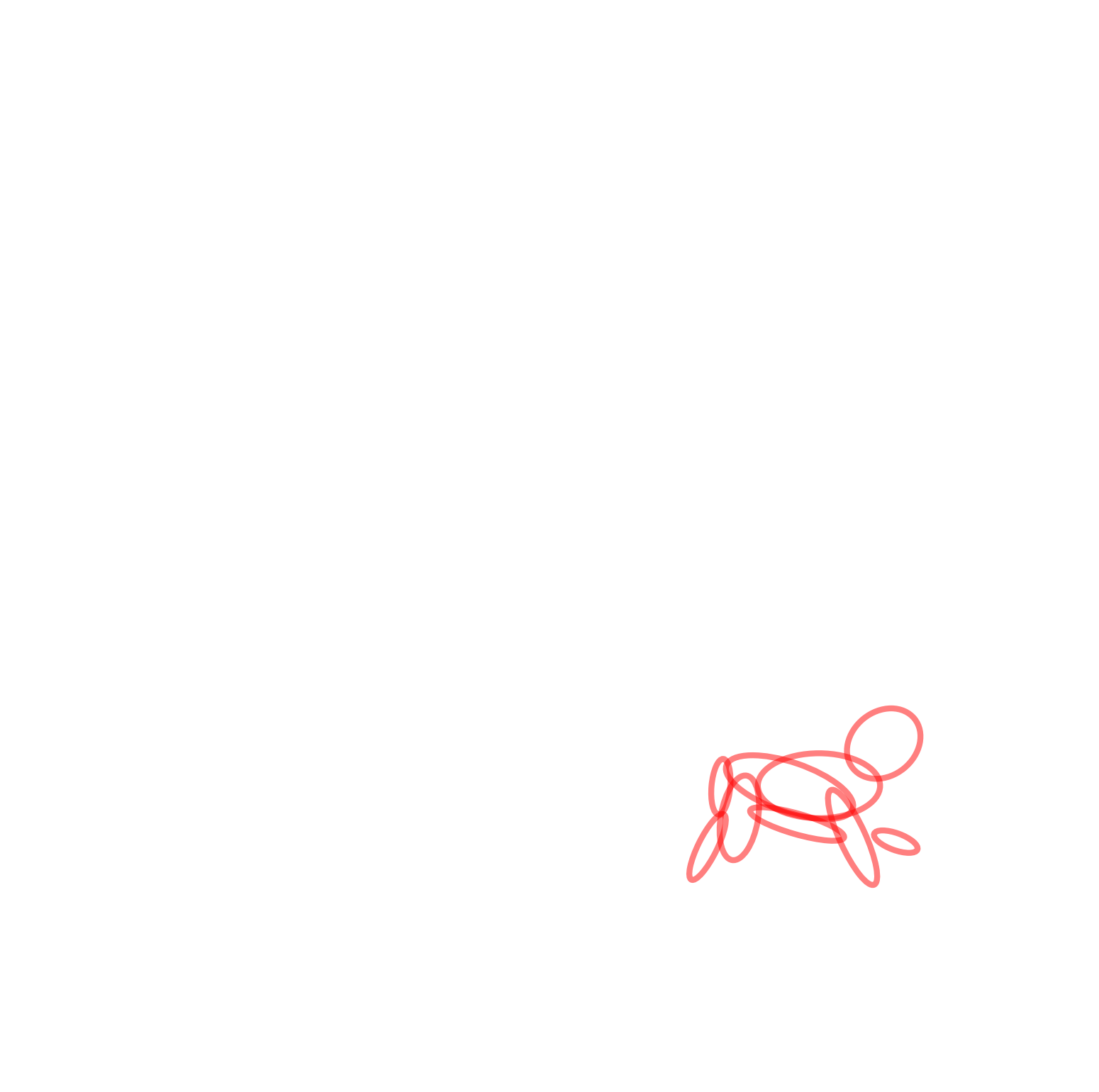}};
          
    \newcommand{\frustum}[2]{%
    \begin{scope}[scale=#1, shift={#2}]
        \coordinate (F) at (0, .2);
        
        \coordinate (A) at (-1.9, -.1);
        \coordinate (B) at (-1.8, 1.8);
        \coordinate (C) at (.3, 2.4);
        \coordinate (D) at (.25, .65);
        
        \draw[thick] (A) -- (B) -- (C) -- (D) -- cycle;
        
        \draw[thick] (F) -- (A);
        \draw[thick] (F) -- (B);
        \draw[thick] (F) -- (C);
        \draw[thick] (F) -- (D);
        
        \draw[fill=black!10,opacity=.8] (A) -- (B) -- (F) -- cycle;
        \draw[fill=black!10,opacity=.8] (B) -- (C) -- (F) -- cycle;
        \draw[fill=black!10,opacity=.8] (C) -- (D) -- (F) -- cycle;
                        
    \end{scope}}
    
    \frustum{.5}{($(left.south east) + (0,1.5)$)}
    \frustum{.5}{($(right.south east) + (0,1.5)$)}    
             
    \node[align=left,font=\footnotesize,color=red!80!black,text width=8em,inner sep=3pt] (box) at ($(left)!.5!(right) - (0,.5)$) {Moving objects in the views easily break multi-view consistency};
    \draw[thick,red!80!black] (box.south west) -- (box.north west);
    \draw[thick,red!80!black] (box.west) edge[->,bend right=20] ++(-1,-.5); 

    \node[align=center,font=\footnotesize,color=red!80!black,text width=3.1em,inner sep=3pt] (box2) at ($(right) + (2.4,0.5)$) {Distractor\\Gaussians};
    \draw[thick,red!80!black,align=center] (box2.south west) -- (box2.south east);
    \draw[thick,red!80!black] (box2.south) edge[->,bend left=20] ++(-0.1,-1.1);

    \end{tikzpicture}
    \vspace*{-2.8em}
    \captionof{figure}{\textbf{\ours:} Gaussian Splatting 
    struggles with floaters and artifacts when image sequences violate photometric consistency assumptions. 
    Unlike existing distractor-free methods 
    which rely on external semantic features, we propose a fully splatting-based solution grounded in photometric consistency 
    that decomposes 3DGS scenes into static components and per-view distractors. 
    }
    \label{fig:teaser}
    \vspace*{1em}
\end{center}%
}]

\begin{abstract}
Gaussian splatting enables fast novel view synthesis in static 3D environments. However, reconstructing real-world environments remains challenging as distractors or occluders break the multi-view consistency assumption required for accurate 3D reconstruction. Most existing methods rely on external semantic information from pre-trained models, introducing additional computational overhead as pre-processing steps or during optimization. In this work, we propose a novel method, \ours, that directly separates distractors and static scene elements purely based on volume rendering of Gaussian primitives. We initialize Gaussians within each camera view for reconstructing the view-specific distractors to separately model the static 3D scene and distractors in the alpha compositing stages. \ours yields an explicit scene separation of static elements and distractors, achieving comparable results to prior distractor-free approaches without sacrificing rendering speed. We demonstrate \ours's effectiveness on three benchmark data sets for distractor-free novel view synthesis. 
See the project website at {\footnotesize \url{https://aaltoml.github.io/desplat/}}.

\end{abstract}\vspace*{-12pt}


\section{Introduction}
\label{sec:intro}
\renewcommand{\thefootnote}{} 
\footnotetext{\hspace{-5mm} $^{*}$Work partially done during internship at Aalto University.}
\renewcommand{\thefootnote}{\arabic{footnote}} 

%
3D Gaussian Splatting (3DGS, \cite{kerbl20233d}) is a popular framework for novel view synthesis for its fast rendering and training speeds with high-quality results. 
However, learning accurate scene reconstructions with 3DGS from images of non-static scenes containing distractors~\cite{sabour2023robustnerf,klasson2024sources,ren2024nerf}---\eg, moving people, vegetation, or transient effects~\cite{martinbrualla2020nerfw} remains challenging.
Vanilla 3DGS is optimized to satisfy a static volumetric rendering constraint even though reference images can contain multi-view inconsistent distractors. This results in spurious floaters being generated close to camera views or as thin view-specific effects appearing in only a few views. 
This lack of robustness limits applying 3DGS to unstructured image collections, \eg crowd-sourced images~\cite{snavely2006photo}, and casually captured videos~\cite{park2021nerfies,warburg2023nerfbusters}.

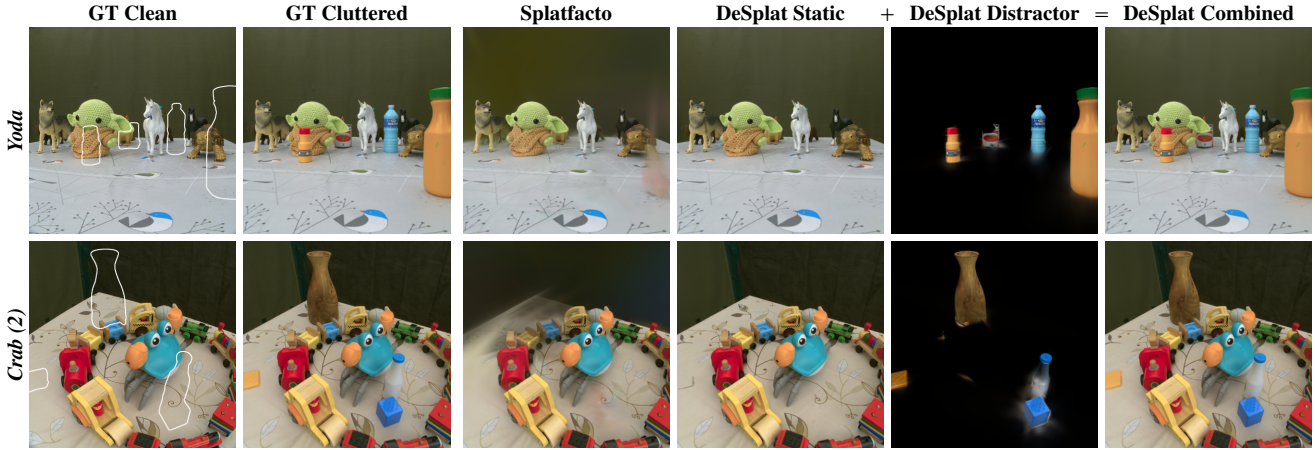
\begin{figure*}[t]
    \centering
    \setlength{\figurewidth}{0.157\textwidth}
    \begin{tikzpicture}[
        image/.style = {inner sep=0pt, outer sep=1pt, minimum width=\figurewidth, anchor=north west, text width=\figurewidth}, 
        node distance = 1pt and 1pt, every node/.style={font= {\tiny}}, 
        label/.style = {font={\footnotesize\bf\vphantom{p}},anchor=south,inner sep=0pt}, 
        spy using outlines={rectangle, red, magnification=1.5, size=0.75cm, connect spies},
        ] 
    
        \node [image] (img-00) {\includegraphics[width=\figurewidth]{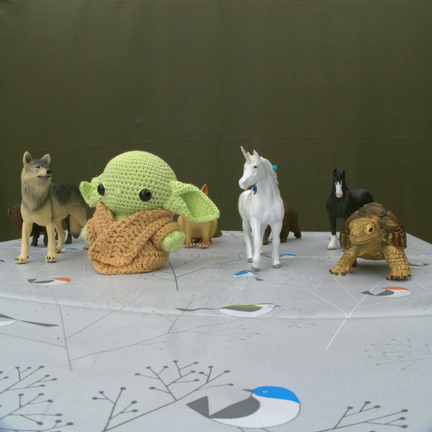}};
        \node [image,right=of img-00] (img-01) {\includegraphics[width=\figurewidth]{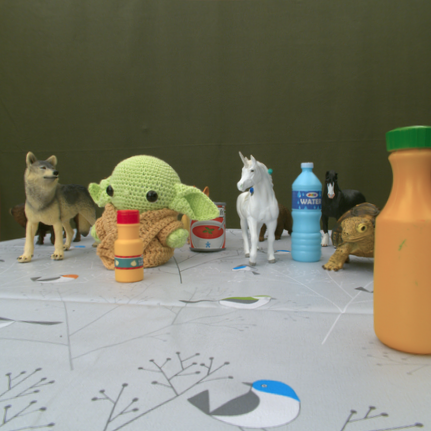}};
        
        \node [image,right=of img-01,xshift=2pt] (img-02) {\includegraphics[width=\figurewidth]{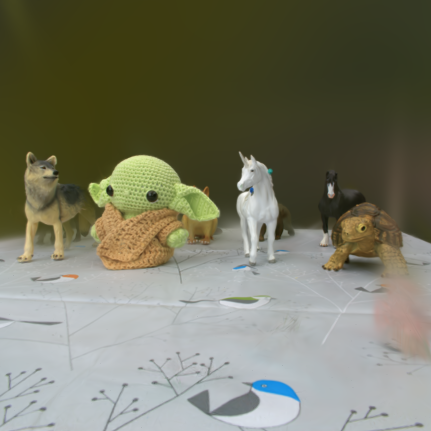}};
        \node [image,right=of img-02] (img-03) {\includegraphics[width=\figurewidth]{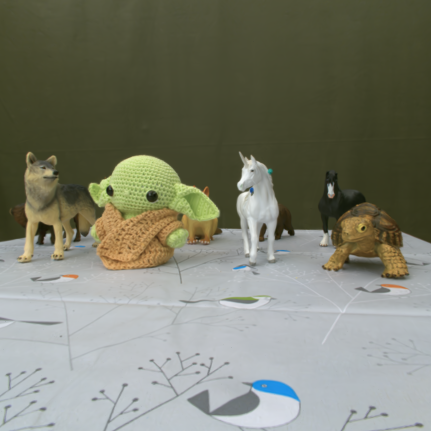}};
        \node [image,right=of img-03] (img-04) {\includegraphics[width=\figurewidth]{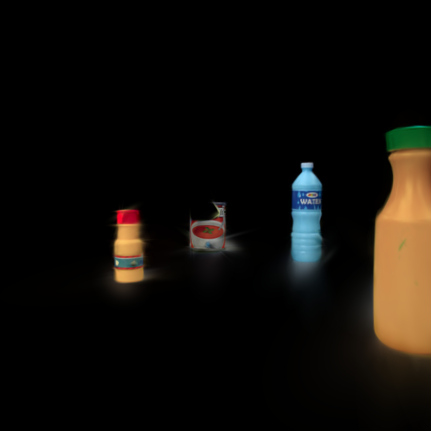}};
        \node[image, right=of img-04] (img-05){\includegraphics[width=\figurewidth]{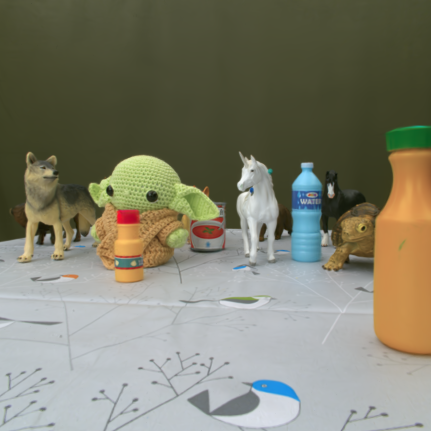}};

        \node [image,below=of img-00] (img-10) {\includegraphics[width=\figurewidth]{        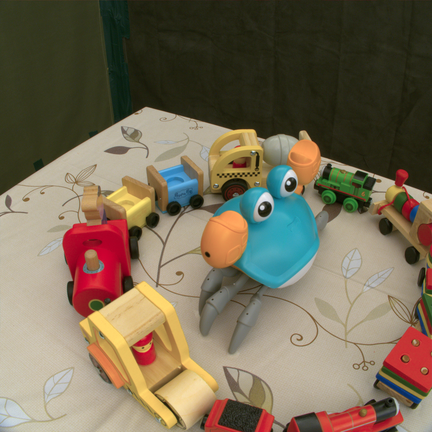}};
        \node [image,right=of img-10] (img-11) {\includegraphics[width=\figurewidth]{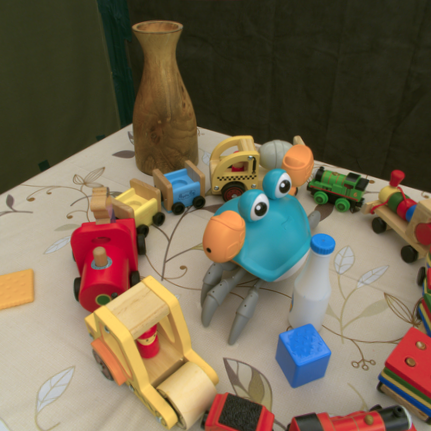}};
        
        \node [image,right=of img-11,xshift=2pt] (img-12) {\includegraphics[width=\figurewidth]{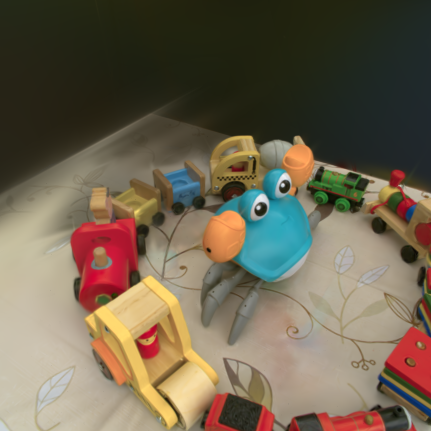}};
        \node [image,right=of img-12] (img-13) {\includegraphics[width=\figurewidth]{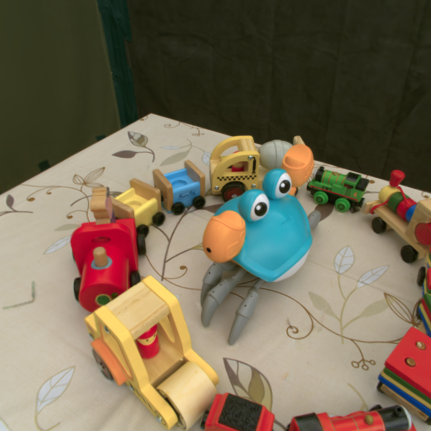}};
        \node [image,right=of img-13] (img-14) {\includegraphics[width=\figurewidth]{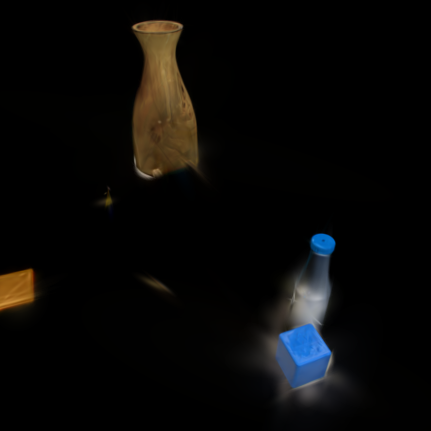}};
        \node[image, right=of img-14] (img-15){\includegraphics[width=\figurewidth]{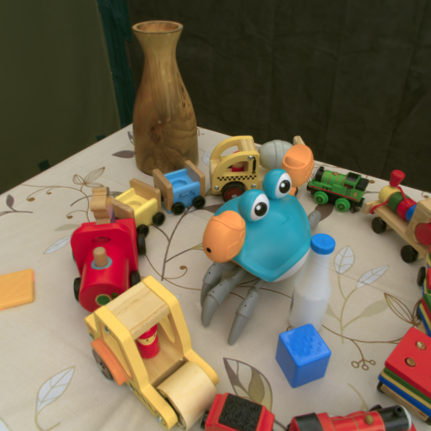}};

        \node[label] (label1) at (img-00.north) {GT Clean};
        \node[label] (label1) at (img-01.north) {GT Cluttered};
        \node[label] (label2) at (img-02.north) {Splatfacto};
        \node[label] (label3) at (img-03.north) {\ours Static};
        \node[label] (label4) at (img-04.north) {\ours Distractor};
        \node[label] (label5) at (img-05.north) {\ours Combined};

        \node at ($(label3)!.5!(label4)$) {$\bm{+}$};
        \node at ($(label4)!.5!(label5)$) {$\bm{=}$};

        \node at (img-00) {\includegraphics[width=\figurewidth]{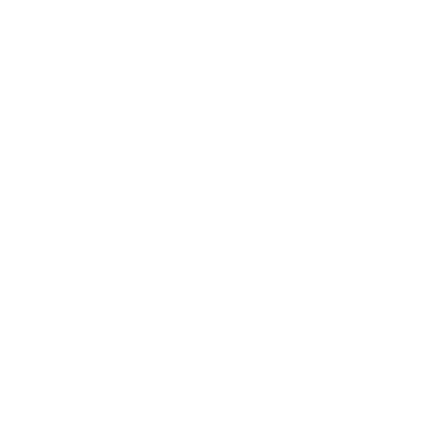}};        
        \node at (img-10) {\includegraphics[width=\figurewidth]{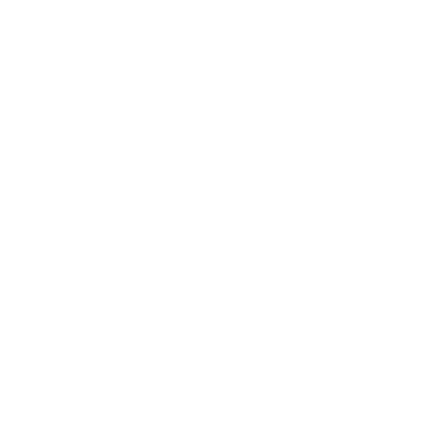}};

        \node[label,rotate=90] (scene1) at (img-00.west) {\itshape Yoda};
        \node[label,rotate=90] (scene2) at (img-10.west) {\itshape Crab~(2)};
    \end{tikzpicture}
    \vspace*{-.1in}
    \caption{\textbf{Qualitative visualization of static and distractor elements achieved by our method, \ours (\cref{sec:decomposed_3dgs}).} In the \textit{Yoda} and \textit{Crab~(2)} scenes~\cite{sabour2023robustnerf}, both clean and cluttered images are captured from the same viewpoints. By explicitly modelling the scene using static and distractor Gaussians, our approach enables clear distractor segmentation and reduced artifacts compared to the Splatfacto baseline~\cite{nerfstudio}.}
    \label{fig:teaser-examples}
\end{figure*}

Prior work in 3DGS and Neural Radiance Fields (NeRFs, \cite{mildenhall2020nerf}) for distractor-free rendering aims to detect which pixels in reference images belong to distractors and to reduce their influence during optimization. 
RobustNeRF~\cite{sabour2023robustnerf} computes robust masks that detect where distractors are present in images from pixel residuals and weigh the reconstruction loss accordingly. 
The same approach has recently been applied to 3DGS where pre-trained networks are utilized for improving the learned masks~\cite{sabour2024spotlesssplats,kulhanek2024wildgaussians}. 
However, it is currently under-explored how the errors from the pre-trained networks affect the 3DGS framework, especially in the densification step~\cite{kerbl20233d,kulhanek2024wildgaussians}. 
Furthermore, recent works closely follow NeRF-W~\cite{martinbrualla2020nerfw} and introduce separate networks for modelling transient effects which introduce additional computational cost resulting in lower rendering speeds~\cite{dahmani2024swag,zhang2024gaussian, kulhanek2024wildgaussians}. 
It remains an open question if the 3DGS inverse rendering framework could be modified to directly recognize what portions of reference images belong to a static scene and what belongs to spurious distractors.


In this paper, we propose decomposing the 3D scene representation into two distinct sets of Gaussian primitives responsible for reconstructing the underlying static scene and distractors respectively. 
Our approach named \ours
involves initializing distractor Gaussians for per-view transient effects which are jointly optimized with the static scene (see \cref{fig:teaser}). 
The explicit modelling of transient and static Gaussians allows decomposing the volumetric rendering equation into two portions and separate optimization into learning per-view transient effects and the static 3D scene utilizing the same photometric reconstruction loss as proposed in Kerbl \etal \cite{kerbl20233d}. See \cref{fig:teaser-examples} for a demonstration. 
This separation maintains fast rendering capabilities without introducing additional pre-processing steps. \looseness-1

\paragraph{Contributions} Our contributions can be summarized as:
\begin{itemize}[leftmargin=*]
    \item We propose decomposing the 3DGS image formation model and training objective to explicitly learn both transient per-view distractors and the static scene relying only on photometric supervision.  
    \item Our approach is efficient and entirely splatting-based allowing for effective robustification of 3DGS which is fully compatible with existing 3DGS tools and pipelines.
    \item We perform experiments on RobustNeRF~\cite{sabour2023robustnerf}, On-the-go~\cite{ren2024nerf}, and Photo Tourism~\cite{snavely2006photo} data sets containing confounding distractors and show that our decomposed 3DGS method performs well 
    against baselines that require pre-trained networks for detecting distractors. 
\end{itemize}

\section{Related Work}
\label{sec:related_work}
\textbf{Robustness in the Presence of Distractors}~~
Novel view synthesis from non-static images is an active research area. 
Early attempts have aimed to separate static scene elements from transient parts~\cite{martinbrualla2020nerfw} and occluders~\cite{sabour2023robustnerf} using photometric-based loss functions. 
For NeRF-based approaches, NeRF-W~\cite{martinbrualla2020nerfw} uses separate MLPs to model static and transient components, applying weights to the per-pixel reconstruction loss based on uncertainty estimates. Other methods utilize learned visibility masks~\cite{chen2022hallucinated,yang2023cross} to modulate the reconstruction loss. RobustNeRF~\cite{sabour2023robustnerf} introduces a kernel-based robust estimator, trained with iteratively re-weighted least squares, to detect distractors using photometric error for loss weighting. Recently, pre-trained models~\cite{kirillov2023segment,oquab2023dinov2} have been adopted to enhance occluder detection through semantic cues. NeRF On-the-go~\cite{ren2024nerf} refines the pixel-wise loss using an uncertainty predictor trained with DINOv2~\cite{oquab2023dinov2} features, while NeRF-HuGS~\cite{chen2024nerf} learns static maps based on SfM features and segmentation from SAM~\cite{kirillov2023segment}.

Several 3DGS-based methods that are robust to distractors have also been introduced. 
SpotlessSplats~\cite{sabour2024spotlesssplats} uses semantic features from Stable Diffusion~\cite{rombach2022high,tang2023emergent} for spatial and spatio-temporal clustering combined with robust masking to identify distractors.
WildGaussians~\cite{kulhanek2024wildgaussians} predicts uncertainty similarly as~\cite{ren2024nerf}, but computes a binary mask from the uncertainties for gradient scaling to stabilize optimization. 
For unstructured photo collections, 
Splatfacto-W~\cite{xu2024splatfactow} uses MLPs for appearance and background modelling combined with robust masks for handling transient effects and objects,
SWAG~\cite{dahmani2024swag} predicts per-Gaussian opacity shifts with an MLP combined with image-conditioned embeddings, while \cite{xu2024wildgs,zhang2024gaussian} learns visibility masks with a U-Net~\cite{ronneberger2015u} and feature embeddings for static and transient object handling. 
The mentioned methods require neural networks to model transient effects and occluders, either with pre-trained models~\cite{kulhanek2024wildgaussians,sabour2024spotlesssplats} or learning feature embeddings~\cite{dahmani2024swag,xu2024splatfactow,xu2024wildgs,zhang2024gaussian} during optimization, which introduce additional computational costs.
In contrast, we propose a pure splatting-based framework that learns to reconstruct static and distractor objects separately by decomposing the volume rendering function and minimizing the photometric loss from 3DGS~\cite{kerbl20233d}.

\begin{table}[t!]
    \centering\small
    \caption{\textbf{Comparison to prior art.} We briefly summarize prior methods based on three criteria: need for external pre-trained models for feature or segmentation generation, rendering performance compared to the original 3DGS work \cite{kerbl20233d}, and explicit scene separation. Explicit scene separation refers to the ability to distinctly separate a scene into static and transient distractor elements.}
    \vspace*{-.1in}
    \setlength{\tabcolsep}{5pt}  
    \newcommand{\rb}[1]{\tikz[baseline]\node[anchor=base,rotate=90,font=\bf,align=left,anchor=west,inner sep=0]{#1};}
    \begin{tabular}{l c c c c c c H}
        \toprule
        & \rb{SWAG \cite{dahmani2024swag}} & \rb{GS-W \cite{zhang2024gaussian}} & \rb{Wild-GS \cite{xu2024wildgs}}   & \rb{Splatfacto-W \cite{xu2024splatfactow}}  & \rb{WildGaussians \cite{kulhanek2024wildgaussians}} & \rb{SpotlessSplats \cite{sabour2024spotlesssplats}}  & \rb{\ours (ours)} \\ 
        \midrule
        {No pre-trained model} & \cmark & \xmark   & \xmark & \cmark & \xmark & \xmark & \cmark \\ 
        {Fast 3DGS rendering speed} & \xmark & \xmark & \xmark & \cmark & \cmark & \cmark & \cmark\\ 
        {Explicit scene separation} & \xmark & \xmark & \xmark & \xmark & \xmark & \xmark & \cmark \\ 
        \bottomrule
    \end{tabular}  
    \label{tab:prior-comparison}
    \vspace{-1em}
\end{table}

\paragraph{Scene Decomposition} 
Learning scene decomposition of static and dynamic objects is an active field in computer vision and graphics~\cite{yunus2024recent}, where generative models have been popular for single- and multi-object decomposition~\cite{bielski2019emergence,chen2019unsupervised,liao2020towards,lin2020space,niemeyer2021giraffe}. 
Several composite radiance fields for decoupling static and dynamic elements have been proposed with NeRFs~\cite{li2021neural,wu2022d} and 3DGS~\cite{yang2024deformable,zhou2024hugs} learned from monocular video. 
Large-scale novel view synthesis with NeRF and 3DGS in urban scenes have incorporated scene semantics~\cite{otonari2024entity,rematas2022urban,tancik2022block} and 3D bounding boxes~\cite{tonderski2024neurad,yan2024street,zhou2024drivinggaussian} to specify moving objects.
Moreover, separate modelling of backgrounds to disentangle static objects from lighting and weather conditions have been proposed~\cite{kulhanek2024wildgaussians,martinbrualla2020nerfw,xu2024splatfactow}. 
In this paper, we propose a decomposition of 3DGS where two sets of Gaussian points are used for fitting the static elements and distractors separately. 
We draw inspiration from representing 2D images with Gaussian Splatting~\cite{zhang2024gaussianimage} and initialize view-specific Gaussian points for modelling per-view distractors near the camera plane.

\paragraph{Comparison to Previous Methods} In \cref{tab:prior-comparison}, we present the main differences between our proposed method and previous works. 
Our method relies solely on volume rendering and a photometric loss, without the need for introducing additional neural networks. GS-W and Wild-GS on the other hand use a pre-trained ResNet-18~\cite{he2016deep} for encoding appearance features, while WildGaussians and SpotLessSplats use features from foundation models~\cite{oquab2023dinov2,rombach2022high} for detecting distractors. 
However, our method is still fully compatible with prior approaches using neural networks for modelling scene appearance \cite{kulhanek2024wildgaussians,sabour2024spotlesssplats}, background \cite{xu2024splatfactow}, and/or distractors due to its generality.
Regarding rendering speed, our method achieves a similar speed as 3DGS since Gaussians are rasterized directly without having to cache image features to maintain fast rendering  times~\cite{xu2024wildgs,zhang2024gaussian}. 
Finally, we achieve explicit scene separation between the underlying static scene and distractor elements by decomposing 3DGS, whilst previous methods aim to mitigate overfitting to distractors via loss masking by detecting distractor pixels from the reference images. Instead, our method jointly reconstructs distractor elements, rather than simply masking them. 

\begin{figure*}[t!]
    \centering
    \setlength{\figurewidth}{1.5cm}
    \input{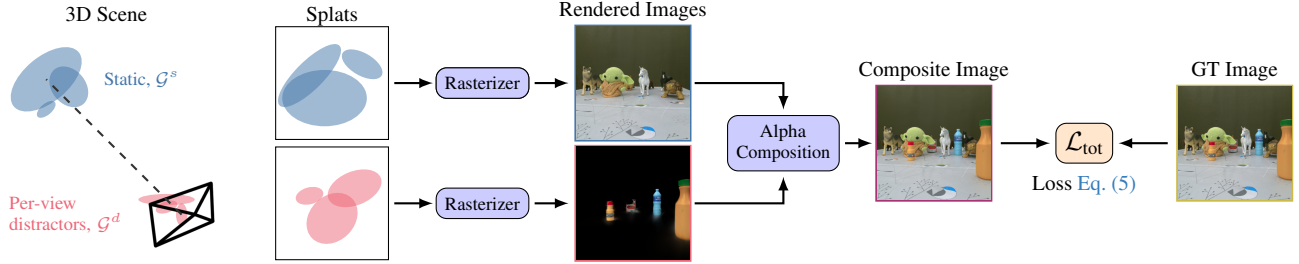}
    \vspace*{-1em}
    \caption{\textbf{Method overview of \ours:} We decompose 3DGS to model the static scene and per-view distractors explicitly. The static scene $\mcG_{s}$ is optimized for all camera views but we allow learning of per-view distractor Gaussians $\mcG_{d}$ to model spurious transient effects which are jointly optimized with the static scene via alpha-compositing. We show how this formulation allows implicit learning of distractor segmentation masks and decomposition of the 3DGS scene into static and distractor elements.}
    \label{fig:pipeline}
\end{figure*}

\section{Method}
\label{sec:method}
%
3D Gaussian splatting \cite{kerbl20233d} represents a scene with a set of 3D Gaussian points $\mcG$. 
Each Gaussian point is associated with a position $\bmu \in \bbR^3$, covariance matrix $\bSigma \in \bbR^{3\times 3}$, opacity $o \in [0, 1]$, and view-dependent colour $\vc$ parameterized by spherical harmonics (SH) coefficients.  

When rendering images, Gaussian points are splatted into screen space as 2D Gaussians with means $\bmu' \in \bbR^2$ 
obtained via projective transformation and covariances approximated by $\bSigma' = \mbJ \mbW \bSigma \mbW^{\top} \mbJ^{\top} \in \bbR^{2 \times 2}$ where $\mbJ \in \bbR^{2 \times 3}$ is the Jacobian of the affine approximation of the projective transformation, and $\mbW \in \bbR^{3 \times 3}$ is the view transformation.
The Gaussians overlapping with the target pixel are sorted according to depth and rendered using alpha-blending: 
\begin{align}
    \vc_{\text{GS}} & = \sum_{i=1}^{N} \vc_i \alpha_i \prod_{j=1}^{i-1} (1 - \alpha_j), \label{eq:gs-volume-rendering} \\
    \alpha_i & = o_i \cdot \exp{(- \frac{1}{2} \Delta_i^{\top} \bSigma_i'^{-1} \Delta_i)},
\end{align}
where $N$ is the number of depth-sorted Gaussians, $\alpha$ denotes alpha-blending weights and $\Delta_i = (\bx_i' - \bmu_i')$ is the offset between the pixel center $\bx_i'$ and the 2D Gaussian mean $\bmu_i'$. 
Gaussian points are optimized by minimizing the following reconstruction loss between rendered pixels and ground truth reference images:
\begin{equation}\label{eq:gs-loss}
    \mcL_{\text{GS}} = (1-\lambda) \mcL_{1} + \lambda \mcL_{\text{D-SSIM}},
\end{equation}
where $\mcL_{1}$ is an $L_1$ loss, $\mcL_{\text{D-SSIM}}$ is a SSIM \cite{dssim} loss, and $\lambda$ is a weighting factor.

The Gaussian points are initialized using a sparse point cloud obtained from a Structure-from-Motion approach, \eg, COLMAP~\cite{schoenberger2016sfm}. 
Gaussians are densified via Adaptive Density Control (ADC, ~\cite{kerbl20233d}) that performs cloning of Gaussians in under-reconstructed regions and splitting large Gaussians in over-reconstructed parts of the scene.

\subsection{Decomposed 3D Gaussian Splatting}
\label{sec:decomposed_3dgs}
We propose decomposing the 3DGS representation to model the static scene and distractors explicitly. 
The idea is to initialize two sets of Gaussian points $\mcG_{s}$ and $\mcG_{d}$, 
which are optimized to reconstruct the static scene and 
distractors respectively. 
We render images using volume rendering \cite{porter1984compositing} (ref.\ \cref{eq:gs-volume-rendering}) by decomposing the output pixel colour for a given view into two distinct blended images, one for the static scene and one for distractors using
\begin{equation}\label{eq:desplat-alpha-composition}
    \vc_{\text{comp}} = \vc_{d} + (1 {-} \alpha_{d}) \vc_{s} ~\text{and}~ 
    \alpha_{d} = \!\sum_{i=1}^{N_d} \!\alpha_{d,i} \prod_{j=1}^{i-1} (1 {-} \alpha_{d,j}), 
\end{equation}
where $\vc_{d}$ and $\vc_{s}$ 
are alpha-composited colours pre-multiplied with alpha-blending weights for the distractor and static Gaussians respectively. $\alpha_{d}$ 
represents the blending weight of the $N_d$ distractor Gaussians overlapping with the target pixel. 
This separation allows rendering the static scene elements and distractors independently.

\paragraph{Per-view Distractor Gaussians} Decomposing a scene into static and distractor Gaussians is challenging when distractors appear in several views or shift slightly between frames. A globally shared set of distractor Gaussians may not be effective since distractors often behave as view-dependent effects. Therefore, we initialize a set of distractor Gaussians for each $n$-th view as $\mcG_{d,n}$ where $n=1,..., N_{\text{train}}$ and $N_{\text{train}}$ is the total number of training images. 
The per-view distractor Gaussians are optimized only 
when the associated camera view is selected during optimization.

\paragraph{Initialization of Distractor Gaussians} We initialize the locations of the distractor Gaussians per-view by placing them on a 2D plane in front of camera views. 
For a single view with camera-to-world rotation matrix $\mbR$ and translation vector $\vt$, the distractor Gaussian positions are initialized as $\bmu_d = \vt - \mbR^{\top} \vu$, where $\vu = [\rho u, \rho v, \rho]^{\top}$. Here, $u, v \sim \mcU(0, 1)$ are sampled from a uniform distribution, and the constant $\rho$ controls the depth, with $\rho = 0.02$ performing well in our experiments. Other Gaussian parameters follow the standard initialization from 3DGS \cite{kerbl20233d}. A fixed number of distractor Gaussians are initialized for each training camera view.\looseness-1

\paragraph{Per-view Densification} 
We apply Adaptive Density Control (ADC, \cite{kerbl20233d}) to the per-view distractor Gaussians to handle varying number and area of distractors in the training images. Unlike the ADC in 3DGS that densifies Gaussian points after every $T$ iteration, we densify the distractor Gaussians when their corresponding image has been trained $S$ times. The culling, splitting, and duplication procedures are the same as in 3DGS for both the distractor and static Gaussians.

\begin{figure*}[t]
    \centering
    \setlength{\figurewidth}{0.19\textwidth}
    \input{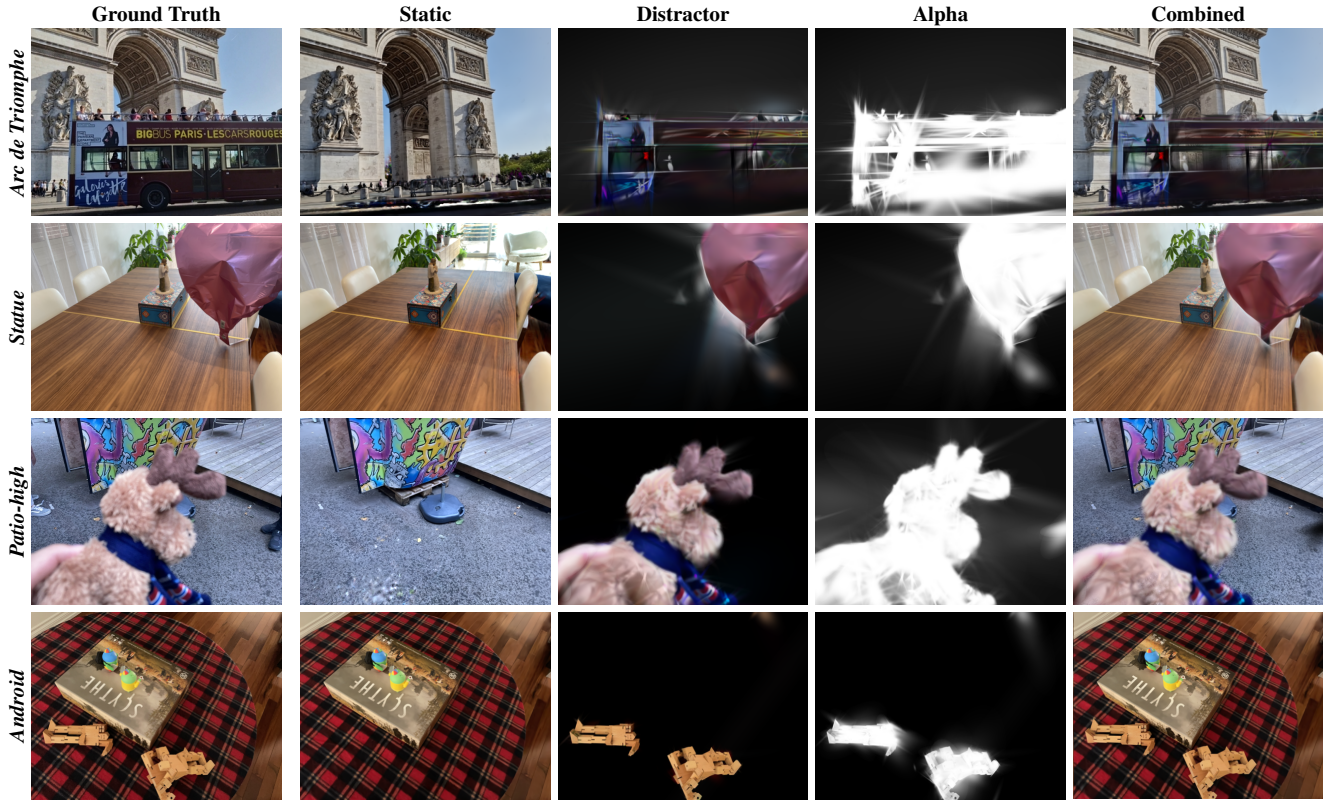}
    \vspace*{-1em}
    \caption{\textbf{Handling occluders.} Examples of static and transient components alongside the composited images. The transient components exhibit well-defined boundaries, while the quality of the static scene is preserved. This showcases the ability of our model to learn transient components without requiring semantic supervision or external priors from pre-trained networks. 
    }
    \label{fig:handling-occluders}
\end{figure*}

\paragraph{Method Pipeline} \cref{fig:pipeline} gives an overview of the rasterization and loss computation of \ours. 
The static Gaussian points are initialized globally using the sparse point cloud obtained from COLMAP as commonly done in 3DGS~\cite{kerbl20233d}, whereas distractor Gaussian points are initialized locally in front of each camera view. 
The static and view-specific distractor Gaussians are splatted onto the screen space and rasterized separately with \cref{eq:gs-volume-rendering} to render two images representing the static elements and distractors respectively. 
We obtain a composite image via alpha-composition of the two images as in \cref{eq:desplat-alpha-composition}, which is compared against the ground-truth image to compute a photometric loss. 
By reconstructing both static elements and distractors, we obtain an explicit scene separation that allows rendering the static and distractor components independently.  

\paragraph{Loss Function} 
We apply regularization on the opacities and alpha maps of the distractor Gaussians to enhance the distractor separation. The idea is to reconstruct the distractors with as few Gaussians as possible, which has previously been applied to balance memory and compute in 3DGS~\cite{kheradmand20243d,klasson2024sources}.  
The final loss function is defined as 
\begin{equation}\label{eq:loss-total}
    \mcL_{\text{tot}} = \mcL_{\text{GS}} + \lambda_{s} |1 - \alpha_s| + \lambda_{d} |\alpha_d|,
\end{equation}
where the second and third terms are $L_1$ regularizers on the static and distractor blending weights $\alpha_s$ and $\alpha_d$ with multipliers $\lambda_{d}$ and $\lambda_{s}$, respectively. 
These terms discourage 
from using distractor Gaussians to reconstruct static objects 
and encourage the accumulation of static Gaussians being equal to one, 
thereby reducing the risk of holes in the background.

\begin{figure*}[t!]
    \centering
    \setlength{\figurewidth}{0.135\textwidth}
    \begin{tikzpicture}[
        image/.style = {inner sep=0pt, outer sep=1pt, minimum width=\figurewidth, anchor=north west, text width=\figurewidth}, 
        node distance = 1pt and 1pt, every node/.style={font= {\tiny}}, 
        label/.style = {font={\footnotesize\bf\vphantom{p}},anchor=south,inner sep=0pt}, 
        spy using outlines={rectangle, magnification=2, size=0.36\figurewidth}
        ] 

        \node [image] (img-00) {\includegraphics[width=\figurewidth]{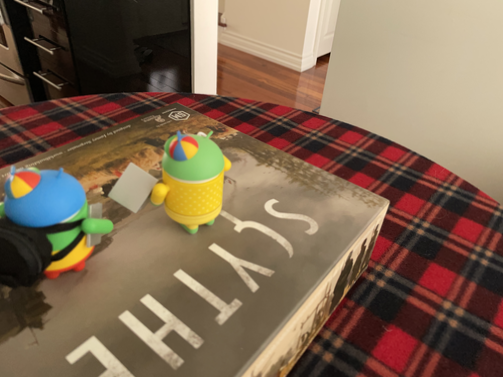}};
        \spy [red,magnification=1.5] on ($(img-00.north) + (0.2\figurewidth,-0.15\figurewidth)$) in node [right] at ($(img-00.east) + (0,0.52*0.36\figurewidth)$); 
        \spy [blue] on ($(img-00) + (0.25\figurewidth, - 0.20\figurewidth)$) in node [right] at ($(img-00.east) - (0,0.52*0.36\figurewidth)$);

        \node [image,right=of img-00,xshift=0.42\figurewidth] (img-01) {\includegraphics[width=\figurewidth]{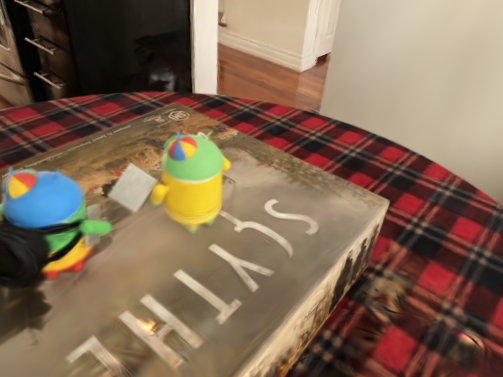}};
        \spy [red,magnification=1.5] on ($(img-01.north) + (0.2\figurewidth,-0.15\figurewidth)$) in node [right] at ($(img-01.east) + (0,0.52*0.36\figurewidth)$); 
        \spy [blue] on ($(img-01) + (0.25\figurewidth, - 0.2\figurewidth)$) in node [right] at ($(img-01.east) - (0,0.52*0.36\figurewidth)$);

        \node [image,right=of img-01,xshift=0.42\figurewidth] (img-02) {\includegraphics[width=\figurewidth]{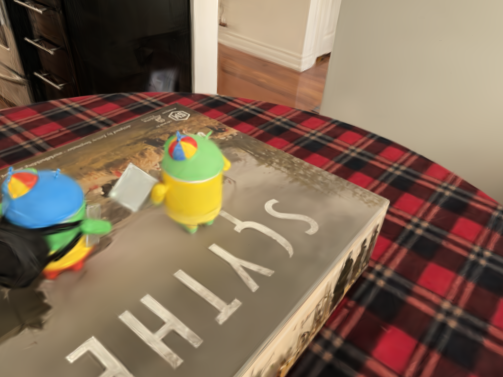}};
        \spy [red,magnification=1.5] on ($(img-02.north) + (0.2\figurewidth,-0.15\figurewidth)$) in node [right] at ($(img-02.east) + (0,0.52*0.36\figurewidth)$); 
        \spy [blue] on ($(img-02) + (0.25\figurewidth, - 0.20\figurewidth)$) in node [right] at ($(img-02.east) - (0,0.52*0.36\figurewidth)$);
        
        \node [image,right=of img-02,xshift=0.42\figurewidth] (img-03) {\includegraphics[width=\figurewidth]{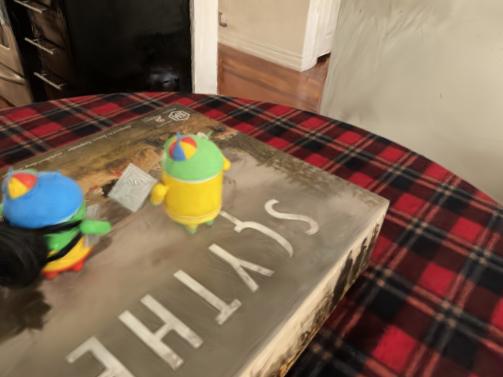}};
        \spy [red,magnification=1.5] on ($(img-03.north) + (0.2\figurewidth,-0.15\figurewidth)$) in node [right] at ($(img-03.east) + (0,0.52*0.36\figurewidth)$); 
        \spy [blue] on ($(img-03) + (0.25\figurewidth, - 0.20\figurewidth)$) in node [right] at ($(img-03.east) - (0,0.52*0.36\figurewidth)$);

        \node [image,right=of img-03,xshift=0.42\figurewidth] (img-04) {\includegraphics[width=\figurewidth]{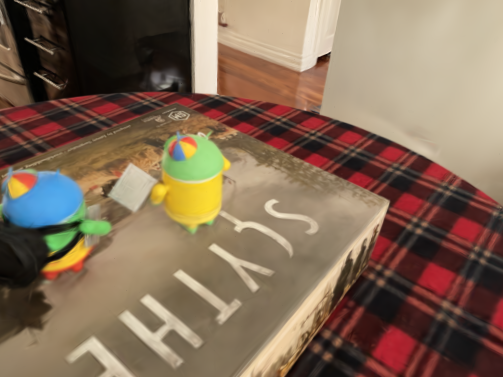}};
        \spy [red,magnification=1.5] on ($(img-04.north) + (0.2\figurewidth,-0.15\figurewidth)$) in node [right] at ($(img-04.east) + (0,0.52*0.36\figurewidth)$); 
        \spy [blue] on ($(img-04) + (0.25\figurewidth, - 0.20\figurewidth)$) in node [right] at ($(img-04.east) - (0,0.52*0.36\figurewidth)$);

        \node [image, below=0.05\figurewidth of img-00] (img-10) {\includegraphics[width=\figurewidth]{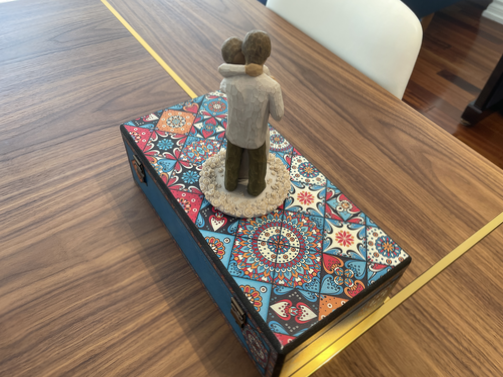}};
        \spy [red,magnification=2] on ($(img-10.north) + (+0.35\figurewidth,-0.2\figurewidth)$) in node [right] at ($(img-10.east) + (0,0.52*0.36\figurewidth)$); 
        \spy [blue,magnification=1.5] on ($(img-10.north) + (0.23\figurewidth, - 0.60\figurewidth)$) in node [right] at ($(img-10.east) - (0,0.52*0.36\figurewidth)$);

        \node [image,right=of img-10,xshift=0.42\figurewidth] (img-11) {\includegraphics[width=\figurewidth]{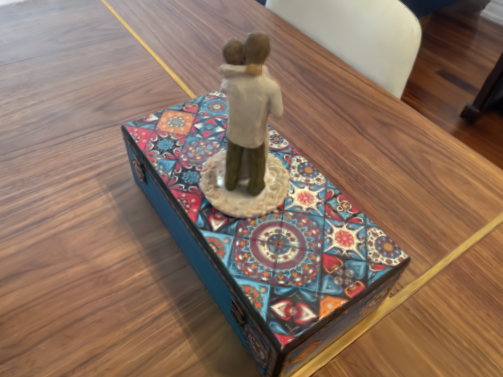}};
        \spy [red,magnification=2] on ($(img-11.north) + (+0.35\figurewidth,-0.2\figurewidth)$) in node [right] at ($(img-11.east) + (0,0.52*0.36\figurewidth)$); 
        \spy [blue,magnification=1.5] on ($(img-11.north) + (0.23\figurewidth, - 0.60\figurewidth)$) in node [right] at ($(img-11.east) - (0,0.52*0.36\figurewidth)$);

        \node [image,right=of img-11,xshift=0.42\figurewidth] (img-12) {\includegraphics[width=\figurewidth]{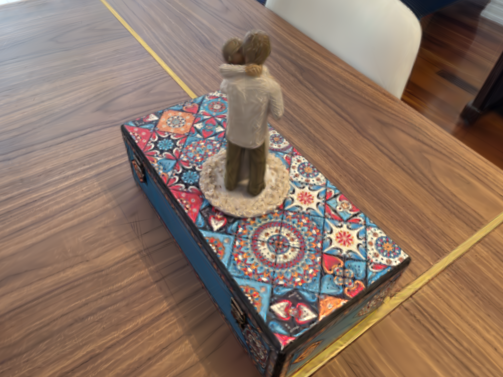}};
        \spy [red,magnification=2] on ($(img-12.north) + (+0.35\figurewidth,-0.2\figurewidth)$) in node [right] at ($(img-12.east) + (0,0.52*0.36\figurewidth)$); 
        \spy [blue,magnification=1.5] on ($(img-12.north) + (0.23\figurewidth, - 0.60\figurewidth)$) in node [right] at ($(img-12.east) - (0,0.52*0.36\figurewidth)$);
        
        \node [image,right=of img-12,xshift=0.42\figurewidth] (img-13) {\includegraphics[width=\figurewidth]{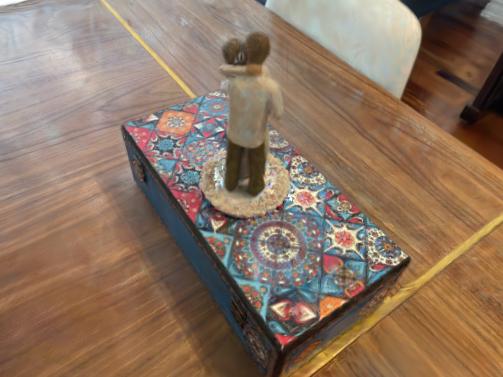}};
        \spy [red,magnification=2] on ($(img-13.north) + (+0.35\figurewidth,-0.2\figurewidth)$) in node [right] at ($(img-13.east) + (0,0.52*0.36\figurewidth)$); 
        \spy [blue,magnification=1.5] on ($(img-13.north) + (0.23\figurewidth, - 0.60\figurewidth)$) in node [right] at ($(img-13.east) - (0,0.52*0.36\figurewidth)$);

        \node [image,right=of img-13,xshift=0.42\figurewidth] (img-14) {\includegraphics[width=\figurewidth]{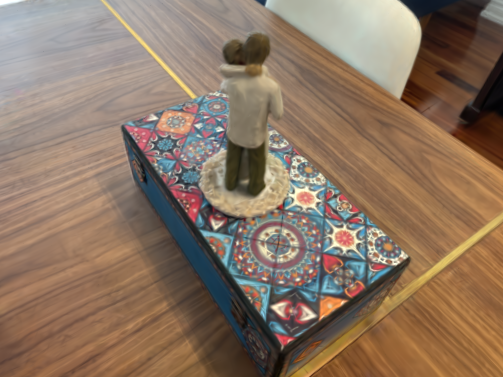}};
        \spy [red,magnification=2] on ($(img-14.north) + (+0.35\figurewidth,-0.2\figurewidth)$) in node [right] at ($(img-14.east) + (0,0.52*0.36\figurewidth)$); 
        \spy [blue,magnification=1.5] on ($(img-14.north) + (0.23\figurewidth, - 0.60\figurewidth)$) in node [right] at ($(img-14.east) - (0,0.52*0.36\figurewidth)$);
        
        \node[label, xshift=0.25\figurewidth] (label1) at (img-00.north) {Ground Truth};
        \node[label, xshift=0.25\figurewidth] (label2) at (img-01.north) {Splatfacto};
        \node[label, xshift=0.25\figurewidth] (label3) at (img-02.north) {Splatfacto-W-T \cite{xu2024splatfactow}};
        \node[label, xshift=0.25\figurewidth] (label4) at (img-03.north) {SpotlessSplats \cite{sabour2024spotlesssplats}};
        \node[label, xshift=0.25\figurewidth] (label5) at (img-04.north) {\ours (ours)};

        \node[label,rotate=90] (scene1) at (img-00.west) {\itshape Android};
        \node[label,rotate=90] (scene2) at (img-10.west) {\itshape Statue};
    \end{tikzpicture}
    \caption{\textbf{Qualitative results on the RobustNeRF data set \cite{sabour2023robustnerf}.} In the {\it Android} and {\it Statue} scenes, \ours generates fewer artefacts than Splatfacto and reconstructs static objects and backgrounds accurately. 
    More qualitative examples in \cref{fig:qualitative-robustnerf-all} in the Appendix.
    }
    \label{fig:qualitative-robustnerf}
\end{figure*}
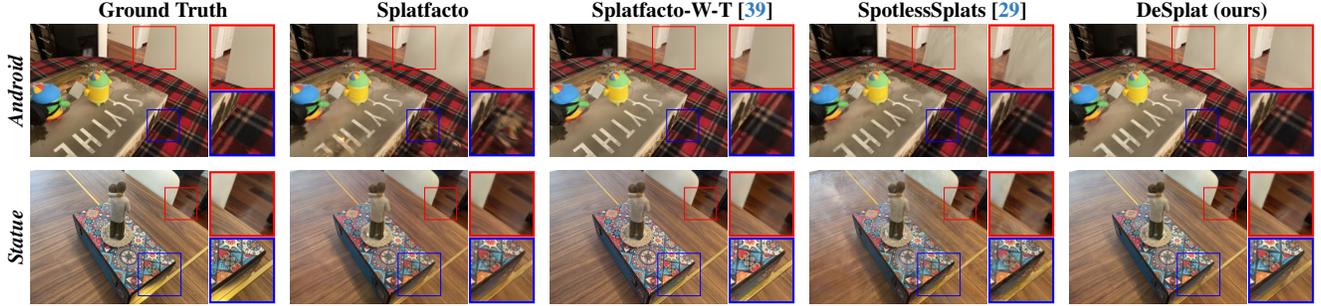

\begin{table*}[t!]
    \centering
    \caption{\textbf{Performance comparison between our method and the baselines on the RobustNeRF data set~\cite{sabour2023robustnerf}}. The \first{first}, \second{second}, and \third{third} best values are highlighted. $^*$ denotes that the reported baseline was ran by us, and `$-$' denotes that the metric is missing in the corresponding work. 
    The explicit scene separation from our \ours achieves the best performance on most scenes. 
    }
    \label{tab:robustnerf}
    \vspace*{-.1in}
    \setlength{\tabcolsep}{2pt}    
    \resizebox{1.0\linewidth}{!}{%
    \begin{tabular}{l|ccc|ccc|ccc|ccc|ccc}
\toprule
    & \multicolumn{3}{c|}{\textbf{\itshape Statue}} & \multicolumn{3}{c|}{\textbf{\itshape Android}} & \multicolumn{3}{c|}{\textbf{\itshape Crab~(1)}} & \multicolumn{3}{c|}{\textbf{\itshape Crab~(2)}} & \multicolumn{3}{c}{\textbf{\itshape BabyYoda}}  \\
    \textbf{Method} & \textbf{PSNR}$\uparrow$ & \textbf{SSIM}$\uparrow$ & \textbf{LPIPS}$\downarrow$ & \textbf{PSNR}$\uparrow$ & \textbf{SSIM}$\uparrow$ & \textbf{LPIPS}$\downarrow$ & \textbf{PSNR}$\uparrow$ & \textbf{SSIM}$\uparrow$ & \textbf{LPIPS}$\downarrow$ & \textbf{PSNR}$\uparrow$ & \textbf{SSIM}$\uparrow$ & \textbf{LPIPS}$\downarrow$ & \textbf{PSNR}$\uparrow$ & \textbf{SSIM}$\uparrow$ & \textbf{LPIPS}$\downarrow$  \\
    \midrule
    RobustNeRF \cite{sabour2023robustnerf} & \val{white}{20.89} & \val{white}{0.75} & \val{white}{0.28} & \val{white}{21.72} & \val{white}{0.65} & \val{white}{0.31} & \val{white}{30.75} & \val{white}{0.81} & \val{white}{0.21} & \val{white}{-} & \val{white}{-} & \val{white}{-} & \val{white}{30.87} & \val{white}{0.83} & \val{white}{0.20} \\
    NeRF On-the-go \cite{ren2024nerf} & \val{white}{21.58} & \val{white}{0.77} & \val{white}{0.24} & \val{white}{23.50} & \val{white}{0.75} & \val{white}{0.21} & \val{white}{-} & \val{white}{-} & \val{white}{-} & \val{white}{-} & \val{white}{-} & \val{white}{-} & \val{white}{29.96} & \val{white}{0.83} & \val{white}{0.24} \\
    NeRF-HuGS \cite{chen2024nerf} & \val{white}{21.00} & \val{white}{0.77} & \val{white}{0.13} & \val{white}{23.32} & \val{white}{0.76} & \val{white}{0.12} & \val{third}{34.16} & \val{third}{0.96} & \val{first}{0.03} & \val{white}{-} & \val{white}{-} & \val{white}{-} & \val{white}{30.70} & \val{white}{0.83} & \val{white}{0.12} \\
    \midrule
    Splatfacto$^*$ \cite{nerfstudio} & \val{third}{22.75} & \val{white}{0.87} & \val{white}{0.11} & \val{white}{24.46} & \val{white}{0.83} & \val{third}{0.09} & \val{white}{24.80} & \val{white}{0.92} & \val{white}{0.13} & \val{white}{31.20} & \val{white}{0.94} & \val{white}{0.11} & \val{third}{32.91} & \val{first}{0.96} & \val{third}{0.09} \\
    Splatfacto-W-A$^*$ \cite{xu2024splatfactow} & \val{white}{22.63} & \val{second}{0.88} & \val{second}{0.10} & \val{white}{24.18} & \val{white}{0.83} & \val{third}{0.09} & \val{white}{25.06} & \val{white}{0.92} & \val{white}{0.15} & \val{third}{31.93} & \val{third}{0.95} & \val{third}{0.09} & \val{white}{29.28} & \val{first}{0.96} & \val{third}{0.09} \\
    Splatfacto-W-T$^*$ \cite{xu2024splatfactow} & \val{second}{23.21} & \val{first}{0.89} & \val{first}{0.09} & \val{first}{25.39} & \val{first}{0.87} & \val{first}{0.07} & \val{white}{28.74} & \val{white}{0.95} & \val{white}{0.09} & \val{white}{24.41} & \val{white}{0.91} & \val{white}{0.19} & \val{white}{31.53} & \val{first}{0.96} & \val{second}{0.08} \\
    SLS-mlp \cite{sabour2024spotlesssplats} & \val{white}{22.69} & \val{white}{0.85} & \val{white}{0.12} & \val{second}{25.15} & \val{second}{0.86} & \val{third}{0.09} & \val{second}{35.85} & \val{first}{0.97} & \val{third}{0.08} & \val{second}{34.35} & \val{first}{0.96} & \val{first}{0.03} & \val{second}{33.60} & \val{first}{0.96} & \val{white}{0.10} \\
    \midrule
    \textbf{\ours (ours)} & \val{first}{23.40} & \val{second}{0.88} & \val{second}{0.10} & \val{third}{24.80} & \val{third}{0.85} & \val{second}{0.08} & \val{first}{36.15} & \val{first}{0.97} & \val{second}{0.04} & \val{first}{35.23} & \val{first}{0.96} & \val{second}{0.07} & \val{first}{35.60} & \val{first}{0.96} & \val{first}{0.06} \\
    \bottomrule
\end{tabular}   %
    }
\end{table*}

\section{Experiments}
\label{sec:experiments}
We conduct experiments on the data sets RobustNeRF \cite{sabour2023robustnerf}, On-the-go \cite{ren2024nerf}, and Photo Tourism \cite{snavely2006photo} to evaluate the efficacy and robustness of our method across varying occlusion and scene complexities. 
The experiments show \ours's ability to handle occluders and background separation, emphasizing performance in cluttered and open scenes under dynamic lighting conditions. In \cref{sec:exp-distractors}, we benchmark our method against state-of-the-art NeRF- and 3DGS-based baselines, assessing both qualitative and quantitative outcomes. \cref{sec:ablations} presents ablation studies to validate the impact of core model components on occluder handling and scene reconstruction. 

\paragraph{Data Sets} We evaluate \ours 
using the RobustNeRF \cite{sabour2023robustnerf}, On-the-go \cite{ren2024nerf}, and Photo Tourism \cite{snavely2006photo} data sets. 
For RobustNeRF, we evaluate on all five available scenes \textit{Statue}, \textit{Android}, \textit{Crab (1)}, \textit{Crab (2)}, and \textit{Yoda}, and downscale all images $8\times$~\cite{sabour2023robustnerf}. 
For On-the-go, we evaluate on the scenes \textit{Mountain}, \textit{Fountain}, \textit{Corner}, \textit{Patio}, \textit{Spot}, and \textit{Patio-High}, 
where all images are downscaled $8{\times}$ except the \textit{Patio} scene which is downscaled $4{\times}$~\cite{ren2024nerf}. Moreover, we use \textit{Arc de Triomphe} for visualization (see \cref{fig:handling-occluders}). 
To validate the effectiveness of \ours 
and its compatibility with prior approaches, we also evaluate on the \textit{Brandenburg Gate}, \textit{Sacre Coeur}, and \textit{Trevi Fountain} scenes from Photo Tourism. 

\paragraph{Baselines} 
We compare \ours against state-of-the-art 3DGS-based approaches for distractor-free rendering, such as 
SpotlessSplats \cite{sabour2024spotlesssplats} (SLS-mlp) and WildGaussians \cite{kulhanek2024wildgaussians}, as well as Splatfacto~\cite{nerfstudio, ye2024gsplat} which serves as our vanilla 3DGS baseline. 
For RobustNerf and On-the-go scenes, we also run Splatfacto-W-A with appearance embeddings and its extension with robust mask for distractor detection called Splatfacto-W-T~\cite{xu2024splatfactow} with their \textit{light} hyperparameter setup. 
On the RobustNeRF scenes, we include the NeRF methods RobustNeRF~\cite{sabour2023robustnerf}, NeRF On-the-go~\cite{ren2024nerf}, and NeRF-HuGS~\cite{chen2024nerf}. 
For Photo Tourism, we compare against SWAG~\cite{dahmani2024swag}, GS-W~\cite{zhang2024gaussian}, Wild-GS~\cite{xu2024wildgs}, Splatfacto-W~\cite{xu2024splatfactow}, and WildGaussians~\cite{kulhanek2024wildgaussians}. 
To ensure consistency, we show the official reported PSNR, SSIM, and LPIPS metrics from the corresponding works to compare against the performance of our \ours.

\paragraph{Implementation Details} 
We follow the hyperparameter settings for Splatfacto~\cite{nerfstudio} and train for 30k iterations on the RobustNeRF and On-the-go scenes. The static Gaussian points are configured using the standard initialization and update procedure for Splatfacto. 
We initialize $K=1000$ distractor points in every training image using the procedure described in \cref{sec:decomposed_3dgs}. 
The gradients of the distractor points are logged for ADC, which is performed for the distractor points in every image after all training images have been seen $S=10$ times. 
All parameters of the distractor points are optimized using Adam~\cite{kingma2015adam}.
Increasing the learning rate by $10\times$ was necessary for the quaternions and scales, while the learning rates for the means, opacities and colours are kept as default. 
We model the colour of distractors directly with three-dimensional RGB vectors instead of relying on SH coefficients, since the distractor points are per-view which alleviates the need for compensating for view-dependent effects using SH.
We also disable the opacity reset~\cite{kerbl20233d} for distractor Gaussians as this improved the separation of distractors from static objects. 
Additionally, we apply clamping of the RGB colours instead of using a sigmoid function~\cite{nerfstudio}, which enabled better learning of distractors with colours at the extremes white (1) and black (0). The regularization weights $\lambda_d$ and $\lambda_s$ are set to 0.01.

For the Photo Tourism data set, we integrate appearance learning as described in WildGaussians~\cite{kulhanek2024wildgaussians} and background modelling following Splatfacto-W~\cite{xu2024splatfactow}. Test-time optimization~\cite{martinbrualla2020nerfw, kulhanek2024wildgaussians} is employed to learn the appearance embeddings of test images. Due to the larger number of input images compared to the RobustNeRF and On-the-go data sets, we adjust the learning rate and other parameters accordingly. See \cref{app:experimental_settings} for more details and hyperparameter values.

\begin{figure*}[t!]
    \centering
    \setlength{\figurewidth}{0.135\textwidth}
    \begin{tikzpicture}[
        image/.style = {inner sep=0pt, outer sep=1pt, minimum width=\figurewidth, anchor=north west, text width=\figurewidth}, 
        node distance = 1pt and 1pt, every node/.style={font= {\tiny}}, 
        label/.style = {font={\footnotesize\bf\vphantom{p}},anchor=south,inner sep=0pt}, 
        spy using outlines={rectangle, magnification=2, size=0.36\figurewidth},
        ]

        \node [image] (img-00) {\includegraphics[width=\figurewidth, height=0.75\figurewidth, keepaspectratio=false, clip]{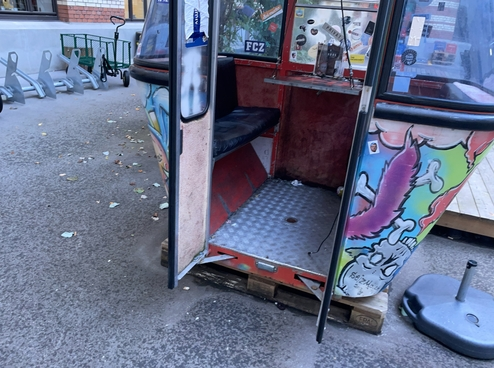}
        };
        \spy [red,magnification=1.5] on ($(img-00) + (-0.37\figurewidth, 0.025\figurewidth)$) in node [right] at ($(img-00.east) + (0,0.52*0.36\figurewidth)$);
        \spy [blue] on ($(img-00) + (0.17\figurewidth, -0.15\figurewidth)$) in node [right] at ($(img-00.east) - (0,0.52*0.36\figurewidth)$);

        \node [image,right=of img-00,xshift=0.42\figurewidth] (img-01) {\includegraphics[width=\figurewidth]{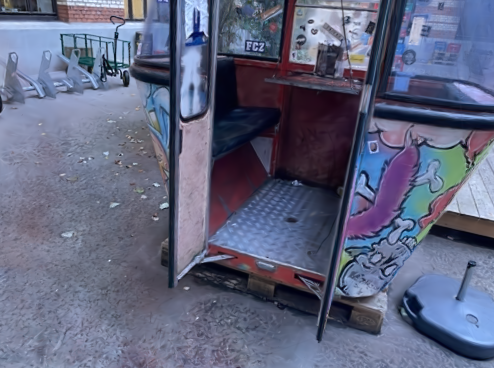}};

        \spy [red,magnification=1.5] on ($(img-01) + (-0.37\figurewidth, 0.025\figurewidth)$) in node [right] at ($(img-01.east) + (0,0.52*0.36\figurewidth)$);
        \spy [blue] on ($(img-01) + (0.17\figurewidth, -0.15\figurewidth)$) in node [right] at ($(img-01.east) - (0,0.52*0.36\figurewidth)$);

        \node [image,right=of img-01,xshift=0.42\figurewidth] (img-02) {\includegraphics[width=\figurewidth]{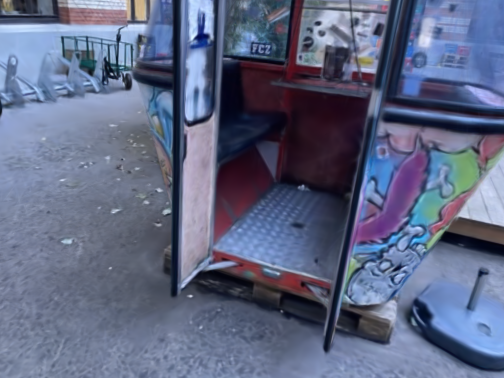}};

        \spy [red,magnification=1.5] on ($(img-02) + (-0.37\figurewidth, 0.025\figurewidth)$) in node [right] at ($(img-02.east) + (0,0.52*0.36\figurewidth)$);
        \spy [blue] on ($(img-02) + (0.17\figurewidth, -0.15\figurewidth)$) in node [right] at ($(img-02.east) - (0,0.52*0.36\figurewidth)$);

         \node [image,right=of img-02,xshift=0.42\figurewidth] (img-03) {\includegraphics[width=\figurewidth]{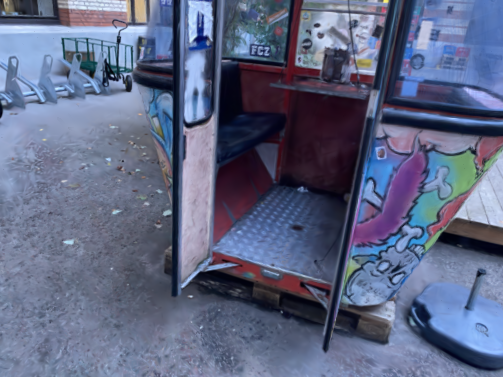}};

        \spy [red,magnification=1.5] on ($(img-03) + (-0.37\figurewidth, 0.025\figurewidth)$) in node [right] at ($(img-03.east) + (0,0.52*0.36\figurewidth)$);
        \spy [blue] on ($(img-03) + (0.17\figurewidth, -0.15\figurewidth)$) in node [right] at ($(img-03.east) - (0,0.52*0.36\figurewidth)$); 
        
        \node [image,right=of img-03,xshift=0.42\figurewidth] (img-04) {\includegraphics[width=\figurewidth]{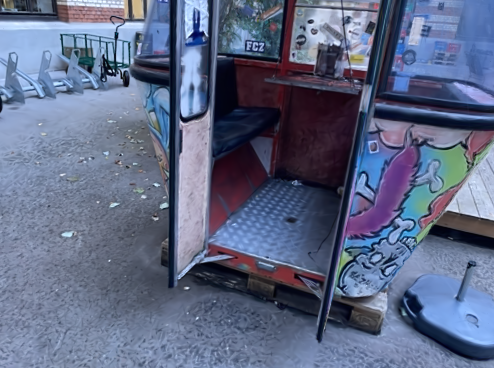}};

        \spy [red,magnification=1.5] on ($(img-04) + (-0.37\figurewidth, 0.025\figurewidth)$) in node [right] at ($(img-04.east) + (0,0.52*0.36\figurewidth)$);
        \spy [blue] on ($(img-04) + (0.17\figurewidth, -0.15\figurewidth)$) in node [right] at ($(img-04.east) - (0,0.52*0.36\figurewidth)$); 
        
        \node [image, below=0.05\figurewidth of img-00] (img-10) {\includegraphics[width=\figurewidth]{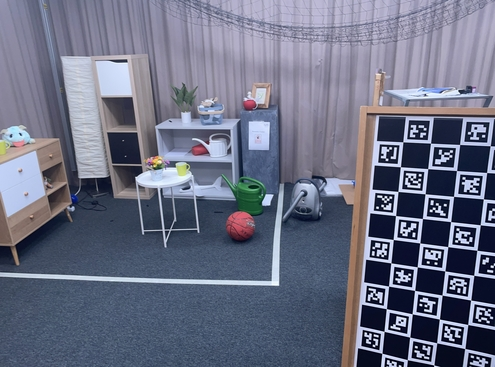}};
        \spy [red,magnification=2.5] on ($(img-10) + (-0.32\figurewidth,0.20\figurewidth)$) in node [right] at ($(img-10.east) + (0,0.52*0.36\figurewidth)$);  
        \spy [blue] on ($(img-10) + (-0.20\figurewidth, - 0.27\figurewidth)$) in node [right] at ($(img-10.east) - (0,0.52*0.36\figurewidth)$);

        \node [image,right=of img-10,xshift=0.42\figurewidth] (img-11) {\includegraphics[width=\figurewidth]{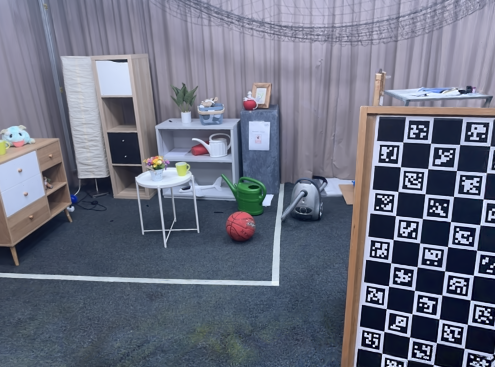}};
        \spy [red,magnification=2.5] on ($(img-11) + (-0.32\figurewidth,0.20\figurewidth)$) in node [right] at ($(img-11.east) + (0,0.52*0.36\figurewidth)$);  
        \spy [blue] on ($(img-11) + (-0.20\figurewidth, - 0.27\figurewidth)$) in node [right] at ($(img-11.east) - (0,0.52*0.36\figurewidth)$);

        \node [image,right=of img-11,xshift=0.42\figurewidth] (img-12) {\includegraphics[width=\figurewidth]{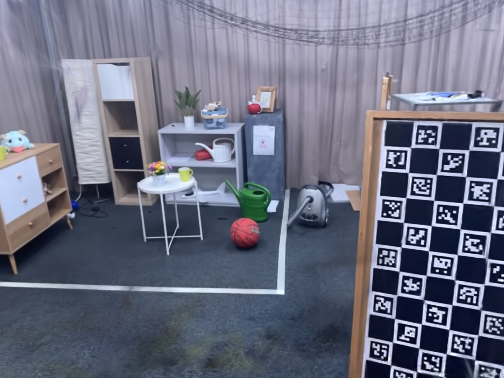}};
        \spy [red,magnification=2.5] on ($(img-12) + (-0.32\figurewidth,0.20\figurewidth)$) in node [right] at ($(img-12.east) + (0,0.52*0.36\figurewidth)$);  
        \spy [blue] on ($(img-12) + (-0.20\figurewidth, - 0.27\figurewidth)$) in node [right] at ($(img-12.east) - (0,0.52*0.36\figurewidth)$);

        \node [image,right=of img-12,xshift=0.42\figurewidth] (img-13) {\includegraphics[width=\figurewidth]{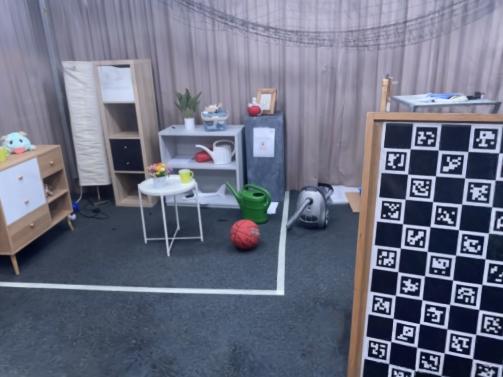}};
        \spy [red,magnification=2.5] on ($(img-13) + (-0.32\figurewidth,0.20\figurewidth)$) in node [right] at ($(img-13.east) + (0,0.52*0.36\figurewidth)$); 
        \spy [blue] on ($(img-13) + (-0.20\figurewidth, - 0.27\figurewidth)$) in node [right] at ($(img-13.east) - (0,0.52*0.36\figurewidth)$);

        \node [image,right=of img-13,xshift=0.42\figurewidth] (img-14) {\includegraphics[width=\figurewidth]{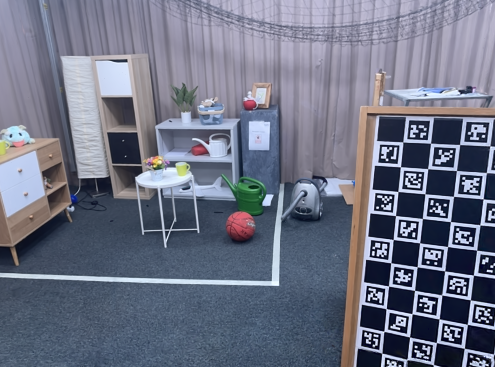}};
        \spy [red,magnification=2.5] on ($(img-14) + (-0.32\figurewidth,0.20\figurewidth)$) in node [right] at ($(img-14.east) + (0,0.52*0.36\figurewidth)$); 
        \spy [blue] on ($(img-14) + (-0.20\figurewidth, - 0.27\figurewidth)$) in node [right] at ($(img-14.east) - (0,0.52*0.36\figurewidth)$);
        
        \node[label, xshift=0.19\figurewidth] (label1) at (img-00.north) {Ground Truth};
        \node[label, xshift=0.19\figurewidth] (label2) at (img-01.north) {Splatfacto};
        \node[label, xshift=0.19\figurewidth] (label3) at (img-02.north) {WildGaussians \cite{kulhanek2024wildgaussians}};
        \node[label, xshift=0.19\figurewidth] (label4) at (img-03.north) {SpotlessSplats \cite{sabour2024spotlesssplats}};
        \node[label, xshift=0.19\figurewidth] (label5) at (img-04.north) {\ours (ours)};

        \node[label,rotate=90] (scene1) at (img-00.west) {\itshape Patio High};
        \node[label,rotate=90] (scene2) at (img-10.west) {\itshape Corner};
    \end{tikzpicture} 
    \caption{\textbf{Qualitative results on NeRF On-the-go ({\itshape Patio High} and {\itshape Corner}) data set.} 
    More examples in \cref{fig:qualitative-onthego-all} in the Appendix.
    }
    \label{fig:qualitative-onthego}
\end{figure*}
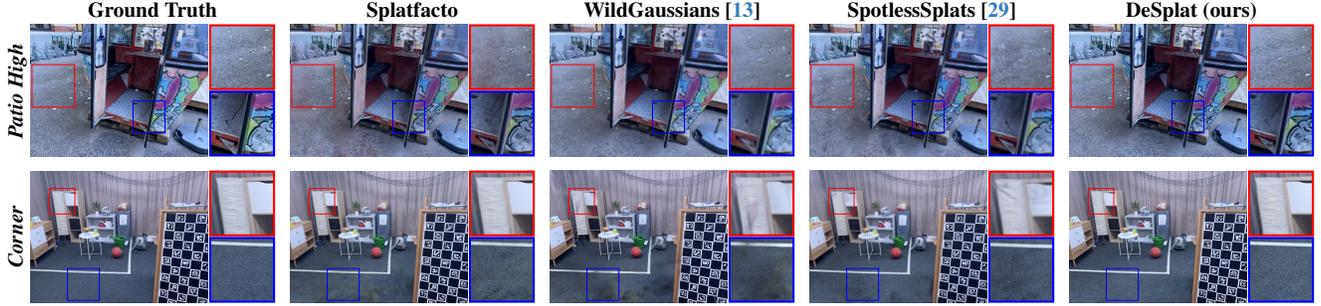

\begin{table*}[t!]
    \centering
    \caption{\textbf{Performance comparison between our method and the baselines on the On-the-go data set~\cite{ren2024nerf}}. The \first{first}, \second{second}, and \third{third} best values are highlighted. $^*$ denotes that the reported baseline was ran by us. 
    Our method performs the best or on par with the baselines for the scenes with medium and high occlusion rate but struggles with modelling outdoor scenes like the {\it Mountain} and {\it Fountain} scenes. 
    }
    \vspace*{-.1in}    
    \setlength{\tabcolsep}{2pt}    
    \resizebox{1.0\linewidth}{!}{%
    \begin{tabular}{l|ccc|ccc|ccc|ccc|ccc|ccc}
\toprule
    & \multicolumn{6}{c|}{\textbf{Low Occlusion}} & \multicolumn{6}{c|}{\textbf{Medium Occlusion}} & \multicolumn{6}{c}{\textbf{High Occlusion}}\\
    & \multicolumn{3}{c|}{\textbf{\itshape Mountain}} & \multicolumn{3}{c|}{\textbf{\itshape Fountain}} & \multicolumn{3}{c|}{\textbf{\itshape Corner}} & \multicolumn{3}{c|}{\textbf{\itshape Patio}} & \multicolumn{3}{c|}{\textbf{\itshape Spot}} & \multicolumn{3}{c}{\textbf{\itshape Patio-High}} \\
    \textbf{Method} & \textbf{PSNR}$\uparrow$ & \textbf{SSIM}$\uparrow$ & \textbf{LPIPS}$\downarrow$ & \textbf{PSNR}$\uparrow$ & \textbf{SSIM}$\uparrow$ & \textbf{LPIPS}$\downarrow$ & \textbf{PSNR}$\uparrow$ & \textbf{SSIM}$\uparrow$ & \textbf{LPIPS}$\downarrow$ & \textbf{PSNR}$\uparrow$ & \textbf{SSIM}$\uparrow$ & \textbf{LPIPS}$\downarrow$ & \textbf{PSNR}$\uparrow$ & \textbf{SSIM}$\uparrow$ & \textbf{LPIPS}$\downarrow$ & \textbf{PSNR}$\uparrow$ & \textbf{SSIM}$\uparrow$ & \textbf{LPIPS}$\downarrow$ \\
    \midrule
    NeRF On-the-go \cite{ren2024nerf} & \val{white}{20.15} & \val{white}{0.64} & \val{white}{0.26} & \val{white}{20.11} & \val{white}{0.61} & \val{white}{0.31} & \val{white}{24.22} & \val{white}{0.81} & \val{white}{0.19} & \val{white}{20.78} & \val{white}{0.75} & \val{white}{0.22} & \val{white}{23.33} & \val{white}{0.79} & \val{white}{0.19} & \val{white}{21.41} & \val{white}{0.72} & \val{white}{0.24} \\
    \midrule
    Splatfacto$^*$ \cite{nerfstudio} & \val{second}{21.08} & \val{second}{0.72} & \val{first}{0.16} & \val{third}{20.62} & \val{second}{0.68} & \val{second}{0.17} & \val{white}{25.37} & \val{third}{0.88} & \val{third}{0.09} & \val{white}{18.06} & \val{white}{0.74} & \val{white}{0.18} & \val{white}{23.73} & \val{third}{0.85} & \val{white}{0.15} & \val{white}{21.68} & \val{third}{0.78} & \val{third}{0.21} \\
    Splatfacto-W-A$^*$ \cite{xu2024splatfactow} & \val{white}{20.31} & \val{white}{0.69} & \val{white}{0.21} & \val{white}{18.74} & \val{white}{0.64} & \val{white}{0.21} & \val{white}{25.18} & \val{white}{0.86} & \val{white}{0.11} & \val{white}{17.92} & \val{white}{0.69} & \val{white}{0.20} & \val{white}{23.29} & \val{white}{0.82} & \val{white}{0.21} & \val{white}{20.68} & \val{white}{0.74} & \val{white}{0.26} \\
    Splatfacto-W-T$^*$ \cite{xu2024splatfactow} & \val{third}{20.94} & \val{white}{0.70} & \val{white}{0.20} & \val{white}{19.20} & \val{white}{0.65} & \val{white}{0.20} & \val{second}{26.15} & \val{second}{0.89} & \val{second}{0.08} & \val{white}{18.57} & \val{white}{0.73} & \val{white}{0.17} & \val{white}{23.80} & \val{third}{0.85} & \val{white}{0.15} & \val{white}{22.05} & \val{white}{0.77} & \val{white}{0.22} \\
    SLS-mlp \cite{sabour2024spotlesssplats} & \val{first}{22.53} & \val{first}{0.77} & \val{third}{0.18} & \val{first}{22.81} & \val{first}{0.80} & \val{first}{0.15} & \val{first}{26.43} & \val{first}{0.90} & \val{white}{0.10} & \val{first}{22.24} & \val{first}{0.86} & \val{first}{0.10} & \val{second}{25.76} & \val{first}{0.90} & \val{second}{0.12} & \val{first}{22.84} & \val{second}{0.83} & \val{second}{0.16} \\
    WildGaussians \cite{kulhanek2024wildgaussians} & \val{white}{20.43} & \val{white}{0.65} & \val{white}{0.26} & \val{second}{20.81} & \val{white}{0.66} & \val{white}{0.22} & \val{white}{24.16} & \val{white}{0.82} & \val{first}{0.05} & \val{second}{21.44} & \val{third}{0.80} & \val{third}{0.14} & \val{third}{23.82} & \val{white}{0.82} & \val{third}{0.14} & \val{third}{22.23} & \val{white}{0.73} & \val{third}{0.21} \\
    \midrule
    \textbf{\ours (ours)} & \val{white}{19.59} & \val{third}{0.71} & \val{second}{0.17} & \val{white}{20.27} & \val{second}{0.68} & \val{second}{0.17} & \val{third}{26.05} & \val{third}{0.88} & \val{third}{0.09} & \val{third}{20.89} & \val{second}{0.81} & \val{second}{0.11} & \val{first}{26.07} & \val{first}{0.90} & \val{first}{0.09} & \val{second}{22.59} & \val{first}{0.84} & \val{first}{0.12} \\
    \bottomrule
\end{tabular}
    }
    \label{tab:onthego}
\end{table*}

\subsection{Distractor-free 3D Reconstruction}
\label{sec:exp-distractors}

\noindent\textbf{Comparison on RobustNeRF Data Set~\cite{sabour2023robustnerf}}~\cref{tab:robustnerf} shows the performance metrics for our \ours and the baselines. 
We observe that \ours performs the best or on par with the baselines across all scenes, especially on the \textit{Yoda} and \textit{Crab} scenes where the distractors are removed and/or new
distractors are introduced for every training image. 
On \textit{Statue} and \textit{Android} scenes, \ours perform similarly to the best baselines Splatfacto-W-T and SLS-mlp without introducing MLPs for modelling varying appearances and lighting changes between the images. 

In \cref{fig:qualitative-robustnerf}, we show qualitative examples of \ours and the best baselines on the \textit{Statue} and \textit{Android} scenes. We observe that \ours removes distractors effectively to preserve fine details within the static scene elements and renders more accurate backgrounds, as highlighted in the zoom-in boxes. 
In the \textit{Statue} scene, \ours not only reconstructs details more accurately but also generates fewer artifacts compared to Splatfacto and SpotlessSplats.
The explicit scene separation by \ours achieves accurate reconstruction of the static elements in the scenes and can handle scenes with multiple distractors of various appearances effectively. 

\paragraph{Comparison on On-the-go Data Set~\cite{ren2024nerf}}
\cref{tab:onthego} shows the performance metrics for our \ours and the baselines, where \ours performs the best or on par with the baselines for the scenes with medium and high occlusion rates. 
However, \ours struggles with modelling outdoor scenes {\it Mountain} and {\it Fountain} scenes with low occlusion rates, where the distractor Gaussians often confuse the sky as a moving distractor between views. 
This typically results in small holes in the rendered images from static Gaussians which in turn penalizes the PSNR severely (see \cref{app:failure-cases} for examples). 
Despite this challenge, our method performs well on the SSIM and LPIPS metrics, which indicates that \ours manages to capture structural and perceptual aspects in its reconstructions of these scenes. 

\cref{fig:qualitative-onthego} shows visual examples of \ours and the best baselines on the \textit{Patio High} and \textit{Corner} scenes. 
In \textit{Patio High}, our \ours and WildGaussians effectively remove floaters arising from the moving people and toys in the training images, while Splatfacto and SpotlessSplats render more artefacts on the ground. 
By closer inspection of the door of the cabin, we observe that \ours manages to reconstruct fine details better than the other baselines in this scene. 
Similarly in the \textit{Corner} scene, 
\ours produces more crisp reconstructions of fine details of static objects compared to SpotLessSplats and WildGaussians. 
Despite being a pure splatting method, \ours often achieves accurate separation of people distractors in these real-world scenes.

\begin{figure}[t!]
  \centering\small
  \begin{subfigure}[t]{0.32\columnwidth}
    \centering
    \includegraphics[width=\textwidth]{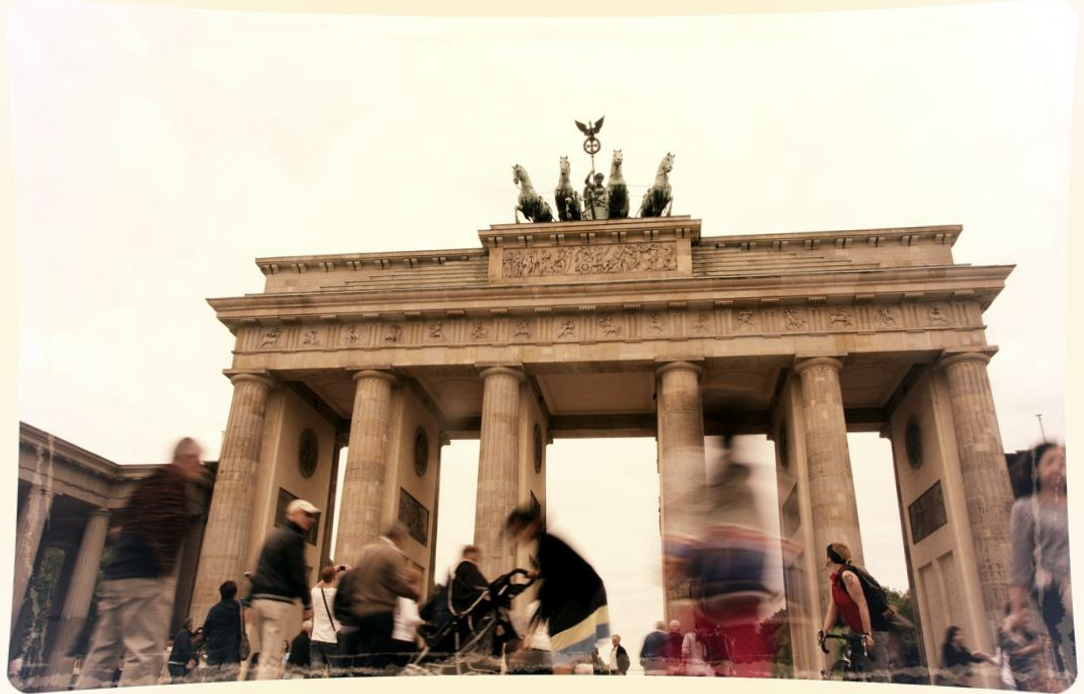}\\
    Ground Truth
  \end{subfigure}
  \hfill
  \begin{subfigure}[t]{0.32\columnwidth}
    \centering
    \includegraphics[width=\textwidth]{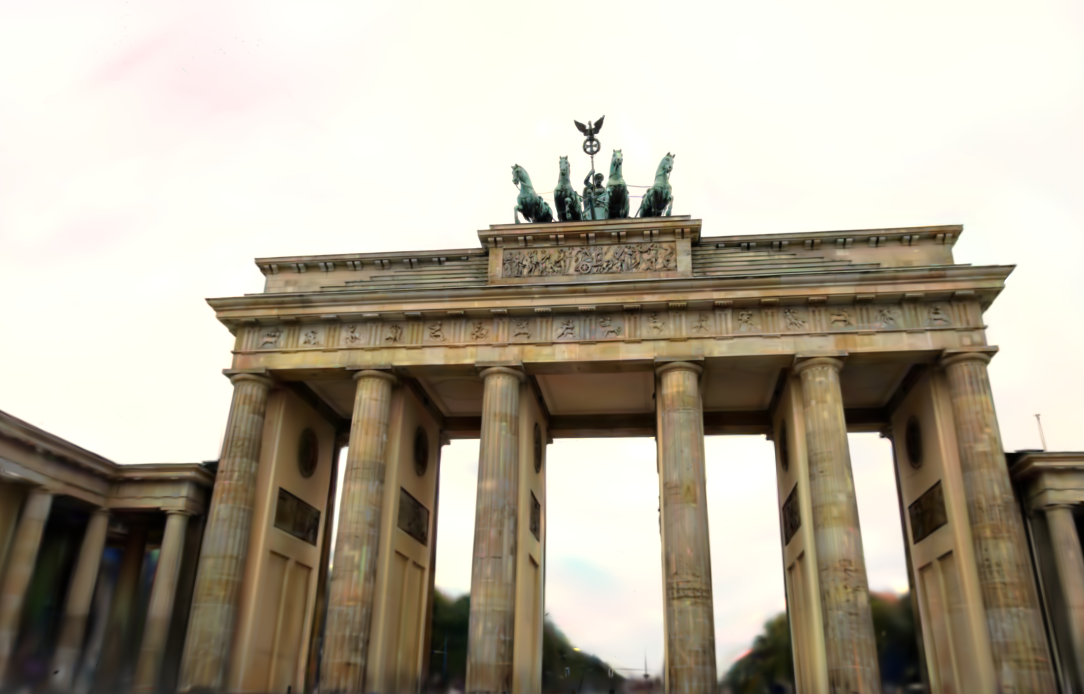}\\
    Static Component
  \end{subfigure}
  \hfill
  \begin{subfigure}[t]{0.32\columnwidth}
    \centering
    \includegraphics[width=\textwidth]{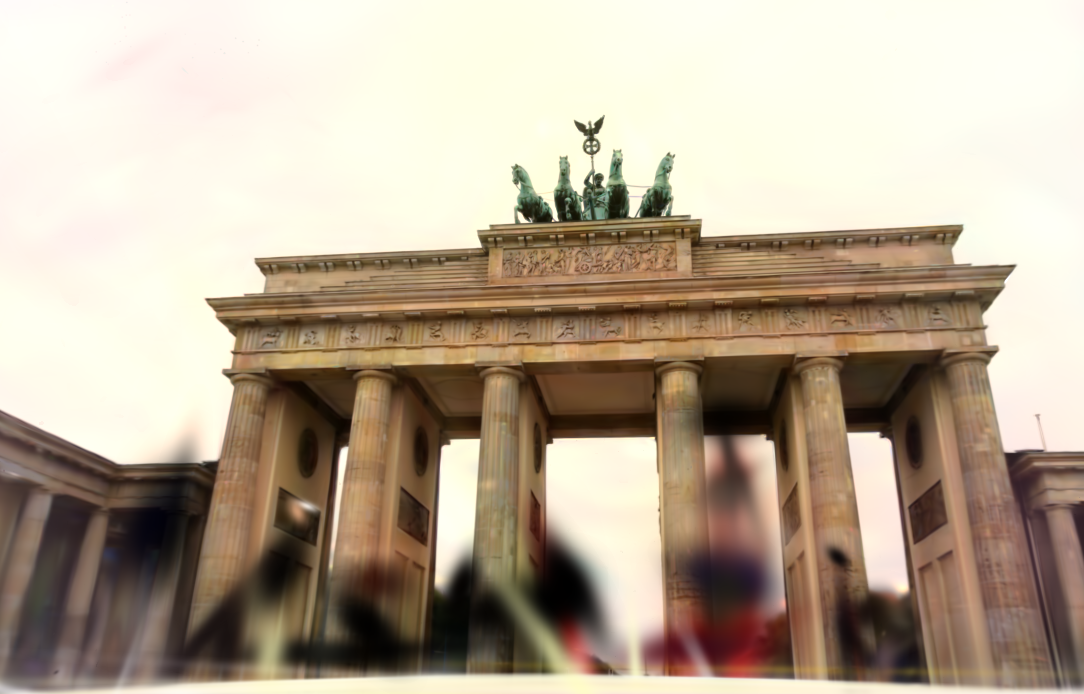}\\
    Composite
  \end{subfigure}
  \caption{\textbf{Qualitative results on a subset of
  \textit{Brandenburg Gate}~\cite{snavely2006photo} 
  with appearance modelling.} 
  We enable per-view appearance modelling following prior approaches \cite{bojanowski2019optimizinglatentspacegenerative, martinbrualla2020nerfw, kulhanek2024wildgaussians, sabour2024spotlesssplats} and our method still effectively separates the static and transient components, whilst learning inter-frame appearance variations.}
  \label{fig:phototourism-appearance}
\end{figure}

\paragraph{Comparison on Photo Tourism Data Set \cite{snavely2006photo}} Even in the challenging Photo Tourism data set, which features large variations in appearance and exposure between frames, 
our method effectively separates occluders from static objects while successfully leveraging appearance embeddings, as shown in \cref{fig:phototourism-appearance}. Quantitative results after test-time optimization, following the protocol described in WildGaussians~\cite{kulhanek2024wildgaussians}, are provided in the supplementary materials~\cref{app:experimental_results}.

\subsection{Ablation Studies}
\label{sec:ablations}
\begin{figure}[th!]
    \centering
    \setlength{\figurewidth}{0.135\textwidth}
    \begin{tikzpicture}[
        image/.style = {inner sep=0pt, outer sep=1pt, minimum width=\figurewidth, anchor=north west, text width=\figurewidth}, 
        node distance = 1pt and 1pt, every node/.style={font= {\tiny}}, 
        label/.style = {font={\footnotesize\bf\vphantom{p}},anchor=south,inner sep=0pt},
    ]

        
        \node [image] (img-00) {\includegraphics[width=\figurewidth]{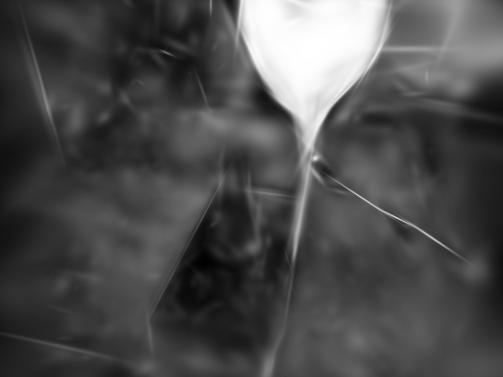}};
        \node [image, right=0.2cm of img-00] (img-01) {\includegraphics[width=\figurewidth]{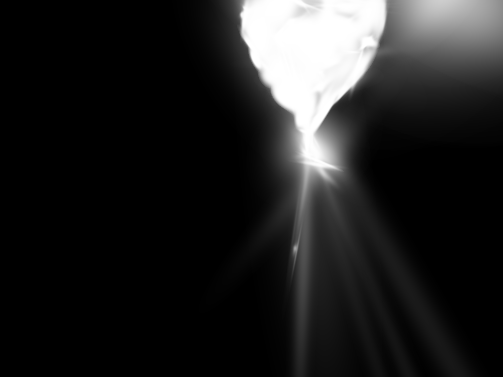}};
        \node [image, right=0.2cm of img-01] (img-02) {\includegraphics[width=\figurewidth]{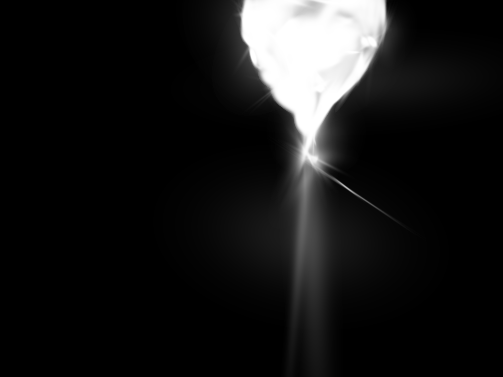}};
        
        \node [image, below=0.2cm of img-00] (img-10) {\includegraphics[width=\figurewidth]{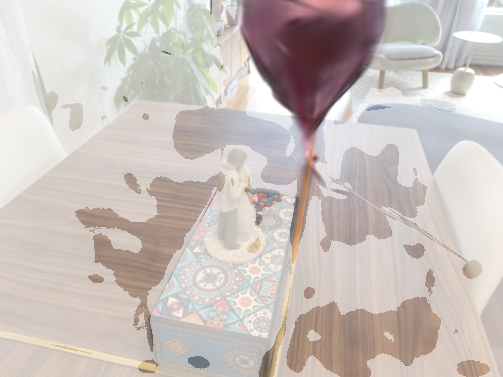}};
        \node [image, right=0.2cm of img-10] (img-11) {\includegraphics[width=\figurewidth]{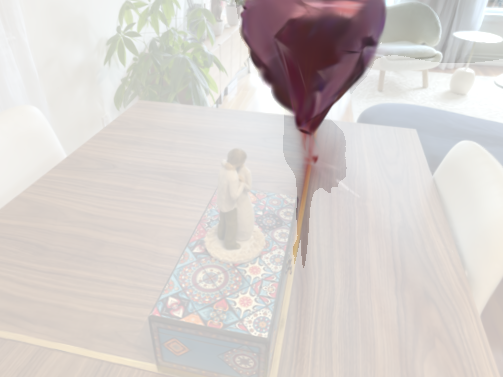}};
        \node [image, right=0.2cm of img-11] (img-12) {\includegraphics[width=\figurewidth]{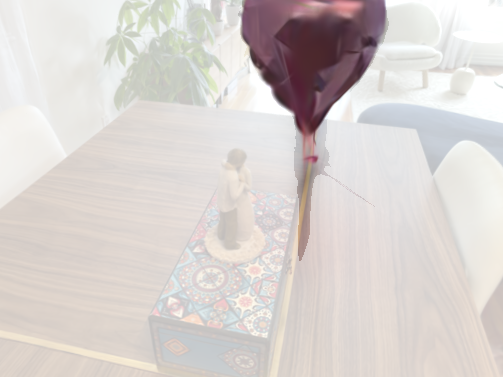}};

        \node[label,rotate=90] (scene1) at (img-00.west) {\itshape Alpha};
        \node[label,rotate=90] (scene2) at (img-10.west) {\itshape RGB w/ mask};
        
        \node[label] (label1) at (img-00.north) {w/o Reg. \& ADC};
        \node[label] (label1) at (img-01.north) {w/o Reg.};
        \node[label] (label2) at (img-02.north) {Ours};
        
    \end{tikzpicture}
    \caption{
    \textbf{Qualitative Ablation on \textit{Statue} \cite{sabour2023robustnerf}} 
    without both regularization (Reg.) and Adaptive Density Control (ADC), without Reg. 
    but with ADC, and Ours. The mask shows where distractor opacity exceeds 0.5.}
    \label{fig:qualitative-ablation}
\end{figure}

\noindent\textbf{Ablation of Our Components~~}
We conduct an ablation study on the \textit{Statue} and \textit{Android} scenes~\cite{sabour2023robustnerf} to evaluate different components of our method: enabling distractor Gaussians, applying ADC to them, and using our opacity regularization terms. \cref{tab:ablation} shows the metrics, with Splatfacto~\cite{nerfstudio} as the baseline (all components off). The results emphasize the importance of our choices. Qualitative results in \cref{fig:qualitative-ablation} show finer separation with ADC and regularization. 
We visualize the distractor regions by masking pixels where the opacity of distractor Gaussians exceeds 0.5. 
Without ADC, the performance is slightly lower compared to Splatfacto~\cite{nerfstudio}, as the redundant distractor Gaussians cannot be pruned, leading to challenges in balancing with static Gaussians. 
While regularization slightly reduces PSNR, it avoids over-explanation of static objects, resulting in better separation as seen in the renders (right-side image, \cref{fig:qualitative-ablation}).

In \cref{app:ablation}, we compare profiling results of memory usage, training times, and rendering speed
between \ours and other approaches. 
Compared to Splatfacto, the addition of distractor Gaussians results in only a slight increase in memory usage, whilst maintaining nearly identical rendering speeds. Compared with SpotlessSplats \cite{sabour2024spotlesssplats} and WildGaussians \cite{xu2024wildgs}, our method not only reduces the time required for training, but exhibits superior memory efficiency. 

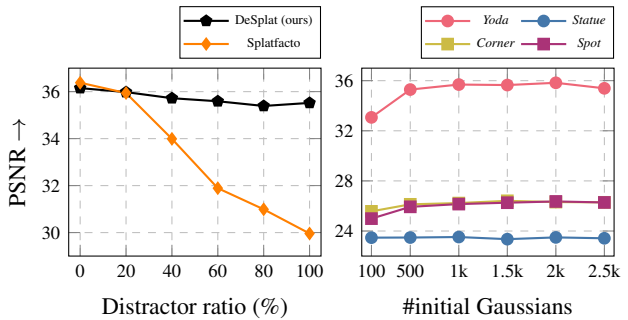
\begin{figure}[t!]
    \centering\small
    \setlength{\figurewidth}{.35\columnwidth}
    \setlength{\figureheight}{.3\columnwidth}
    \pgfplotsset{x tick label style={rotate=0,font=\scriptsize}, 
    ylabel={PSNR~$\rightarrow$},
    grid=both,
    grid style={dashed, thin, gray!50},   
    y tick label style={rotate=0, font=\scriptsize}, 
    legend style={font=\tiny, cells={align=left}},
    legend style={anchor=south east,at={(1,1.05)},inner sep=1pt},
    scale only axis,    
    }
    \begin{subfigure}[b]{.48\columnwidth}
        \centering
        \pgfplotsset{xlabel={\strut Distractor ratio (\%)},
        xmin=-.05, xmax=1.05,
        ymin=29, ymax=37,
        legend columns=-1,
        xtick={0,0.2,0.4,0.6,0.8,1},
        xticklabels={0,20,40,60,80,100},
        ytick={30, 32, 34, 36, 38, 40},
        yticklabels={30, 32, 34, 36, 38, 40},
        legend columns=1, 
        }
        \begin{tikzpicture}
            \begin{axis}[width=1.15\figurewidth,height=\figureheight]
            \addplot[color=black, thick, mark=pentagon*, error bars/.cd, y dir=both, y explicit] coordinates {
                (0, 36.15) (0.2, 35.98) (0.4, 35.72) (0.6, 35.59) (0.8, 35.39)(1.0, 35.52)
            };
            \addlegendentry{\ours (ours)}
            
            \addplot[color=orange, thick, mark=diamond*, error bars/.cd,
                    y dir=both,
                    y explicit] coordinates {
                (0, 36.38)  (0.2, 35.94) (0.4, 33.99) (0.6, 31.89) (0.8, 30.99) (1.0, 29.96)
            };
            \addlegendentry{Splatfacto}
            \end{axis}
     \end{tikzpicture}
  \end{subfigure}
  \hfill
  \begin{subfigure}[b]{.48\columnwidth}
     \centering
     \pgfplotsset{xlabel={\strut\#initial Gaussians},
        ylabel={},
        xmin=0, xmax=2600,
        ymin=22, ymax=37,
        xtick={100,500,1000,1500,2000,2500},
        xticklabels={100,500,1k,1.5k,2k,2.5k},
        ytick={20, 24, 28, 32, 36, 40},
        yticklabels={20, 24, 28, 32, 36, 40},
        legend columns=2
     }
     \begin{tikzpicture}
        \begin{axis}[width=1.15\figurewidth, height=\figureheight]
            \addplot[color=scRed, thick, mark=*] coordinates {
                (100, 33.07) (500, 35.29) (1000, 35.69) (1500, 35.65) (2000, 35.83) (2500, 35.39)
            };
            \addlegendentry{\it Yoda}
            \addplot[color=scBlue, thick, mark=*] coordinates {
                (100, 23.47) (500, 23.48) (1000, 23.52) (1500, 23.35) (2000, 23.49) (2500, 23.42)
            };
            \addlegendentry{\it Statue}
            \addplot[color=scYellow, thick, mark=square*] coordinates {
                (100, 25.57) (500, 26.13) (1000, 26.23) (1500, 26.41) (2000, 26.31) (2500, 26.29)
            };
            \addlegendentry{\it Corner}
            \addplot[color=scPurple, thick, mark=square*] coordinates {
                (100, 24.99) (500, 25.93) (1000, 26.15) (1500, 26.26) (2000, 26.36) (2500, 26.28)
            };
            \addlegendentry{\it Spot}
        \end{axis}
     \end{tikzpicture}
  \end{subfigure}\vspace*{-.1in}
  \caption{\textbf{Ablation study on sensitivity.} (left) PSNR for \ours and Splatfacto on \textit{Yoda} scene over various ratios of distractors. (right) PSNR for \ours with different number of initialized distractor Gaussians per view. \ours performs robustly over different distractor ratios with saturation after $1000$ initialized points.}
  \label{fig:ablation_ratio_and_initialization}
  \vspace{-1em}
\end{figure}

\paragraph{Sensitivity} Not all views in real-world scenes contain occluders. The RobustNeRF data set \cite{sabour2023robustnerf} includes both clean and cluttered data from the same viewpoints, enabling robustness testing by adjusting the ratio of cluttered images. We ablate the performance of our method with different clutter ratios in ~\cref{fig:ablation_ratio_and_initialization}. With ADC enabled for our distractor Gaussians, we demonstrate robustness to various clutter ratios. We notice a significant drop in performance for the Splatfacto baseline when the distractor ratio is increased, whilst our method maintains comparable metrics. In \cref{fig:ablation_ratio_and_initialization}, we also ablate the number of initialized distractor Gaussians with ADC enabled and find that we reach the best performance and saturation with only 1k Gaussians per view.

\begin{table}[t!]
    \centering
    \caption{\textbf{Quantitative ablation on \textit{Statue} and \textit{Android}~\cite{sabour2023robustnerf}}. We compare the impact of various components of our approach.
    }
    \vspace*{-.1in}
    \resizebox{1.0\linewidth}{!}{
        \begin{tabular}{ccc|ccc|ccc}
            \toprule
            \multirow{2}{*}{\begin{tabular}[c]{@{}c@{}}Distractor\\ Gaussians\end{tabular}} & \multirow{2}{*}{ADC} & \multirow{2}{*}{Regularization} & \multicolumn{3}{c|}{\textbf{\textit{Statue}}} & \multicolumn{3}{c}{\textbf{\textit{Android}}} \\
            &&&\textbf{PSNR} & \textbf{SSIM} & \textbf{LPIPS} & \textbf{PSNR} & \textbf{SSIM} & \textbf{LPIPS}\\
            \midrule
            \xmark & \xmark & \xmark & 22.75 & 0.87 & 0.11 & 24.46 & 0.83 & 0.09 \\
            \cmark & \xmark & \xmark & 21.37 & 0.85 & 0.10 & 22.92 & 0.81 & 0.10 \\
            \cmark & \cmark & \xmark & 23.52 & 0.88 & 0.10 & 24.69 & 0.84 & 0.08 \\
            \cmark & \cmark & \cmark & 23.40 & 0.88 & 0.10 & 24.80 & 0.84 & 0.08 \\
            \bottomrule
        \end{tabular}
    }
    \label{tab:ablation}
\end{table}

\section{Conclusions}
\label{sec:conclusions}
We have introduced \ours, a novel method that separates occluders from static scene elements using volume rendering of Gaussian primitives. Unlike previous approaches, \ours is entirely splatting-based without the need for external MLPs or pre-processing, 
lowering computational costs while achieving a clear separation of distractors and static content. 
We evaluate \ours on a variety of real-world data sets, showcasing that it is also compatible with prior approaches using appearance MLPs and background modelling, and achieve comparable results to state-of-the-art methods.

\paragraph{Limitations}
Our method leverages view consistency to extract as much information as possible from the images. However, if distractors appear static across most views, \ours may mistakenly detect them as part of the static scene. 
Additionally, a small colour difference between distractors and static objects can hinder proper separation. 
Another limitation is that \ours may struggle to define clear boundaries for occluders, potentially leading to a loss of detail and reduced image quality. In outdoor scenes with skies and varying lighting or weather conditions, distractor Gaussians may capture these elements, making separation more challenging.\looseness-1


\section*{Acknowledgements}
AS acknowledges funding from the Research Council of Finland (grants 339730 and 362408). 
JK acknowledges funding from the Research Council of Finland (grants 352788, 353138, and 362407).
MK and MT acknowledge funding from the Finnish Center for Artificial Intelligence (FCAI). 
We acknowledge CSC -- IT Center for Science, Finland, and the Aalto Science-IT project for the computational resources. 
We thank Martin Trapp for helpful discussions. 
Lastly, we thank the anonymous reviewers for their thoughtful feedback. 

{
    \small
    \bibliographystyle{ieeenat_fullname}
    \bibliography{main}
}



\appendix
\clearpage
\setcounter{page}{1}
\maketitlesupplementary

\renewcommand{\thetable}{A\arabic{table}}
\renewcommand{\thefigure}{A\arabic{figure}}

\section{Method Details}
\label{app:method}

Here, we provide additional details about \ours. In general, \ours follows the same settings as Splatfacto from Nerfstudio~\cite{nerfstudio}, utilizing a warm-up phase of 500 steps and image downscaling factor of two at the beginning of training. 
We present modifications that improve the separation of static elements and distractors next by disabling opacity reset and modifying the colour representation. Additionally, we describe how we combine \ours with the appearance modelling in~\cite{kulhanek2024wildgaussians} and the background model in~\cite{xu2024splatfactow} which we use in the Photo Tourism experiments.

\paragraph{Disabling Opacity Reset for Distractor Gaussians} ~\citet{kerbl20233d} introduced an opacity reset mechanism in 3DGS that removes inactive Gaussian points, \eg ones located close to camera views.
However, in \ours we only apply opacity reset on static Gaussians and disable it for the distractor Gaussians, as we experimentally found that opacity reset can cause confusion between occluders and static objects.

\paragraph{RGB Colour Representation for Distractor Gaussians} 
We modify the colour representation for the distractor Gaussians to model the colour in RGB space and use clamping to set its range instead of using a sigmoid function. 
Modelling the colours as RGB vectors instead of using SH coefficients alleviates the need to compensate for view-dependent effects using SH as the distractor Gaussians are per view. 
Additionally, applying clamping into the range $[0, 1]$ of the RGB colours instead of using a sigmoid function enabled better learning of distractors with colours at the extremes white (1) and black (0). 

\paragraph{Appearance Embeddings} 
We show that \ours can be combined with MLPs for modelling appearance variations with per-view image embeddings. This is commonly used for NeRF~\cite{martinbrualla2020nerfw} and 3DGS~\cite{kulhanek2024wildgaussians, sabour2024spotlesssplats, xu2024splatfactow, xu2024wildgs} in the Photo Tourism data set which consists of web images with varying weather conditions and lighting scenarios.
We follow the appearance modelling from \citet{kulhanek2024wildgaussians}, which 
uses per-image embeddings $\{ \ve_j \}_{j=1}^{N_{\text{train}}}$ with $N_{\text{train}}$ as the number of training images to handle varying appearances and illuminations, and per-Gaussian embeddings $\{ \vz_i \}_{i=1}^N$ to handle varying colours for each Gaussian under different appearances. 
The per-image embeddings $\ve_j \in \bbR^{d_{\ve}}$, per-Gaussian embeddings $\vz_i \in \bbR^{d_{\vz}}$, and the 0-th order SH coefficient $\Bar{\vc}_i$ are input to an MLP $f_{\bphi}$ as
\begin{equation}
    (\bbeta, \bgamma) = f_{\bphi}(\ve_j, \vz_i, \Bar{\vc}_i) ,
\end{equation}
where $\bbeta, \bgamma \in \bbR^3$ are the shift and scale of an affine transformation respectively, and $\bphi$ are the parameters of the MLP. 
The view-dependent colour of the $i$-th Gaussian $\vc_i$ is then modulated by
\begin{equation}
    \Tilde{\vc}_i = \bgamma \odot \vc_i + \bbeta, 
\end{equation}
where $\odot$ is an element-wise multiplication. The toned colour $\Tilde{\vc}_i$ is then passed to the rasterization. 
In \cref{app:experimental_results}, we show the results for \ours on Photo Tourism  where we apply the appearance modelling on the static Gaussians. 

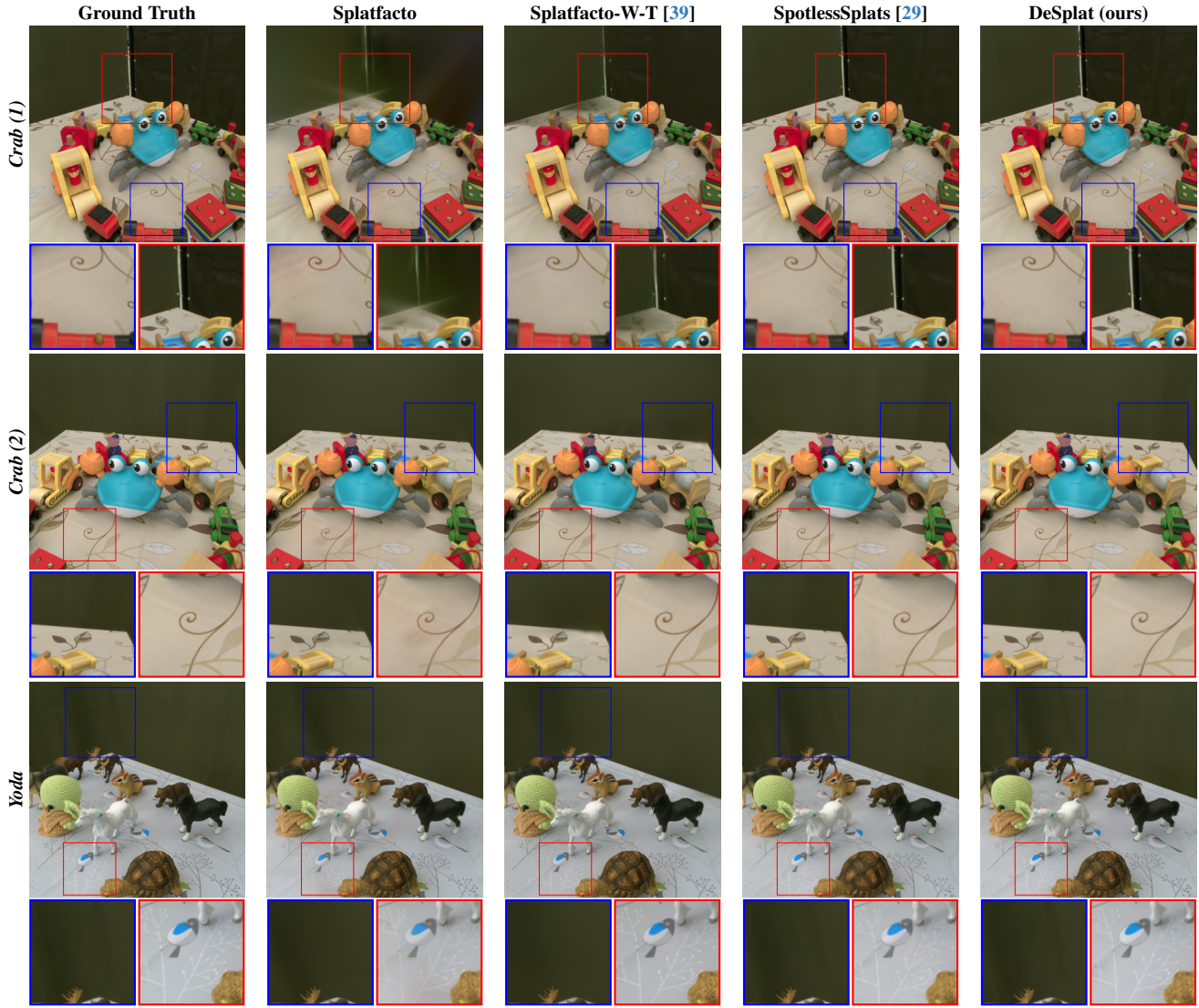
\begin{figure*}[ht!]
    \centering
    \setlength{\figurewidth}{0.18\textwidth}
    \begin{tikzpicture}[
        image/.style = {inner sep=0pt, outer sep=1pt, minimum width=\figurewidth, anchor=north west, text width=\figurewidth}, 
        node distance = 1pt and 1pt, every node/.style={font= {\tiny}}, 
        label/.style = {font={\footnotesize\bf\vphantom{p}},anchor=south,inner sep=0pt}, 
        spy using outlines={rectangle, magnification=2, size=0.485\figurewidth}
        ] 

        \node [image] (img-00) {\includegraphics[width=\figurewidth]{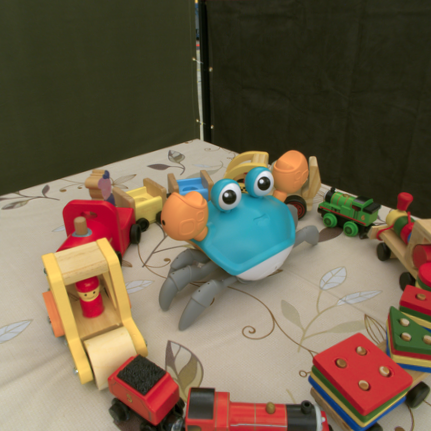}};
        \spy [red,magnification=1.5] on ($(img-00.north) + (0.0\figurewidth,-0.3\figurewidth)$) in node [right] at ($(img-00.south) + (0.01\figurewidth,-0.68*0.36\figurewidth)$); 
        \spy [blue] on ($(img-00) + (0.09\figurewidth, - 0.35\figurewidth)$) in node [left] at ($(img-00.south) - (0.01\figurewidth,0.68*0.36\figurewidth)$);

        \node [image,right=of img-00,xshift=0.07\figurewidth] (img-01) {\includegraphics[width=\figurewidth]{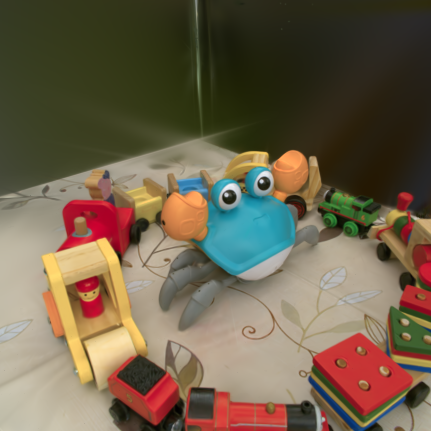}};
        \spy [red,magnification=1.5] on ($(img-01.north) + (0.0\figurewidth,-0.3\figurewidth)$) in node [right] at ($(img-01.south) + (0.01\figurewidth,-0.68*0.36\figurewidth)$); 
        \spy [blue] on ($(img-01) + (0.09\figurewidth, - 0.35\figurewidth)$) in node [left] at ($(img-01.south) - (0.01\figurewidth,0.68*0.36\figurewidth)$);

        \node [image,right=of img-01,xshift=0.07\figurewidth] (img-02) {\includegraphics[width=\figurewidth]{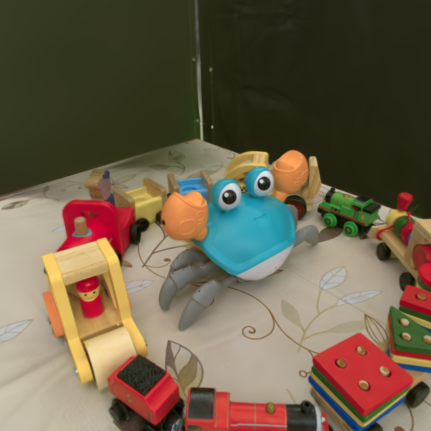}};
        \spy [red,magnification=1.5] on ($(img-02.north) + (0.0\figurewidth,-0.3\figurewidth)$) in node [right] at ($(img-02.south) + (0.01\figurewidth,-0.68*0.36\figurewidth)$); 
        \spy [blue] on ($(img-02) + (0.09\figurewidth, - 0.35\figurewidth)$) in node [left] at ($(img-02.south) - (0.01\figurewidth,0.68*0.36\figurewidth)$);
        
        \node [image,right=of img-02,xshift=0.07\figurewidth] (img-03) {\includegraphics[width=\figurewidth]{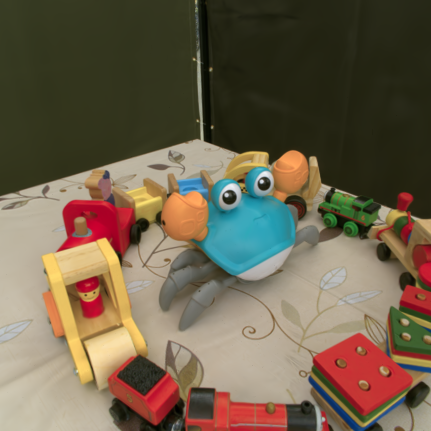}};
        \spy [red,magnification=1.5] on ($(img-03.north) + (0.0\figurewidth,-0.3\figurewidth)$) in node [right] at ($(img-03.south) + (0.01\figurewidth,-0.68*0.36\figurewidth)$); 
        \spy [blue] on ($(img-03) + (0.09\figurewidth, - 0.35\figurewidth)$) in node [left] at ($(img-03.south) - (0.01\figurewidth,0.68*0.36\figurewidth)$);

        \node [image,right=of img-03,xshift=0.07\figurewidth] (img-04) {\includegraphics[width=\figurewidth]{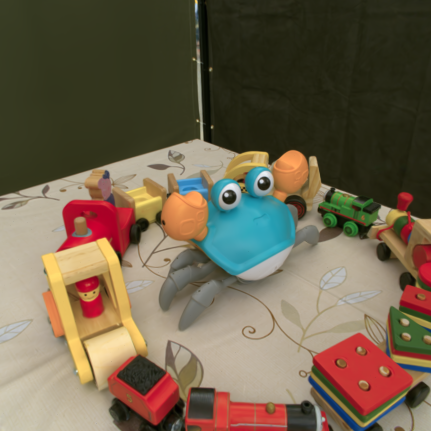}};
        \spy [red,magnification=1.5] on ($(img-04.north) + (0.0\figurewidth,-0.3\figurewidth)$) in node [right] at ($(img-04.south) + (0.01\figurewidth,-0.68*0.36\figurewidth)$); 
        \spy [blue] on ($(img-04) + (0.09\figurewidth, - 0.35\figurewidth)$) in node [left] at ($(img-04.south) - (0.01\figurewidth,0.68*0.36\figurewidth)$);

        \node [image, below=0.5\figurewidth of img-00] (img-10) {\includegraphics[width=\figurewidth]{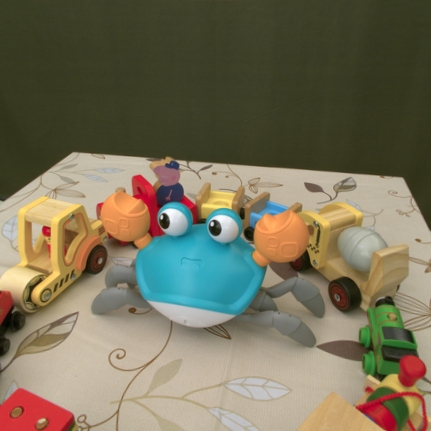}};
        \spy [red] on ($(img-10.north) + (-0.22\figurewidth,-0.85\figurewidth)$) in node [right] at ($(img-10.south) + (0.01\figurewidth,-0.68*0.36\figurewidth)$); 
        \spy [blue,magnification=1.5] on ($(img-10.north) + (0.3\figurewidth, - 0.4\figurewidth)$) in node [left] at ($(img-10.south) - (0.01\figurewidth,0.68*0.36\figurewidth)$);

        \node [image,right=of img-10,xshift=0.07\figurewidth] (img-11) {\includegraphics[width=\figurewidth]{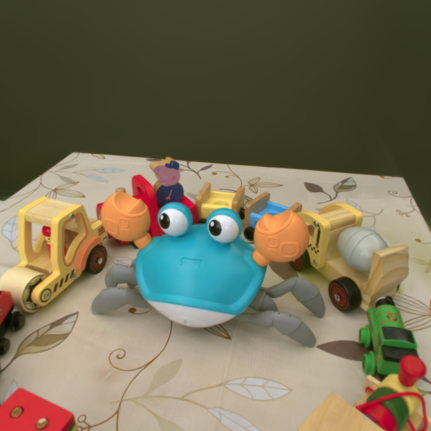}};
        \spy [red] on ($(img-11.north) + (-0.22\figurewidth,-0.85\figurewidth)$) in node [right] at ($(img-11.south) + (0.01\figurewidth,-0.68*0.36\figurewidth)$); 
        \spy [blue,magnification=1.5] on ($(img-11.north) + (0.3\figurewidth, - 0.4\figurewidth)$) in node [left] at ($(img-11.south) - (0.01\figurewidth,0.68*0.36\figurewidth)$);

        \node [image,right=of img-11,xshift=0.07\figurewidth] (img-12) {\includegraphics[width=\figurewidth]{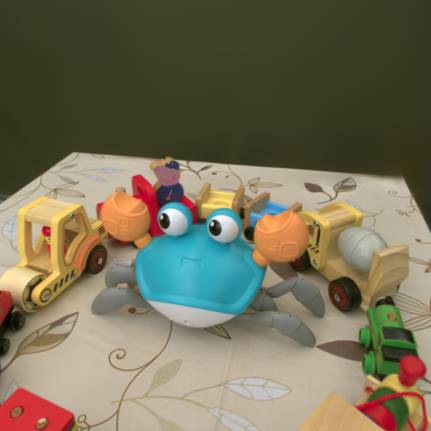}};
        \spy [red] on ($(img-12.north) + (-0.22\figurewidth,-0.85\figurewidth)$) in node [right] at ($(img-12.south) + (0.01\figurewidth,-0.68*0.36\figurewidth)$); 
        \spy [blue,magnification=1.5] on ($(img-12.north) + (0.3\figurewidth, - 0.4\figurewidth)$) in node [left] at ($(img-12.south) - (0.01\figurewidth,0.68*0.36\figurewidth)$);
        
        \node [image,right=of img-12,xshift=0.07\figurewidth] (img-13) {\includegraphics[width=\figurewidth]{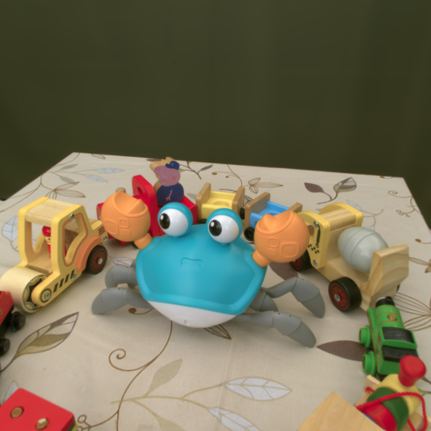}};
        \spy [red] on ($(img-13.north) + (-0.22\figurewidth,-0.85\figurewidth)$) in node [right] at ($(img-13.south) + (0.01\figurewidth,-0.68*0.36\figurewidth)$); 
        \spy [blue,magnification=1.5] on ($(img-13.north) + (0.3\figurewidth, - 0.4\figurewidth)$) in node [left] at ($(img-13.south) - (0.01\figurewidth,0.68*0.36\figurewidth)$);

        \node [image,right=of img-13,xshift=0.07\figurewidth] (img-14) {\includegraphics[width=\figurewidth]{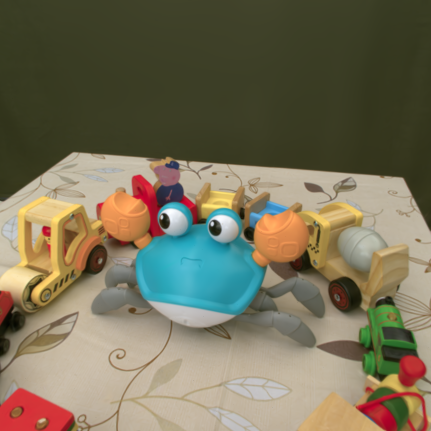}};
        \spy [red] on ($(img-14.north) + (-0.22\figurewidth,-0.85\figurewidth)$) in node [right] at ($(img-14.south) + (0.01\figurewidth,-0.68*0.36\figurewidth)$); 
        \spy [blue,magnification=1.5] on ($(img-14.north) + (0.3\figurewidth, - 0.4\figurewidth)$) in node [left] at ($(img-14.south) - (0.01\figurewidth,0.68*0.36\figurewidth)$);

        \node [image, below=0.5\figurewidth of img-10] (img-20) {\includegraphics[width=\figurewidth]{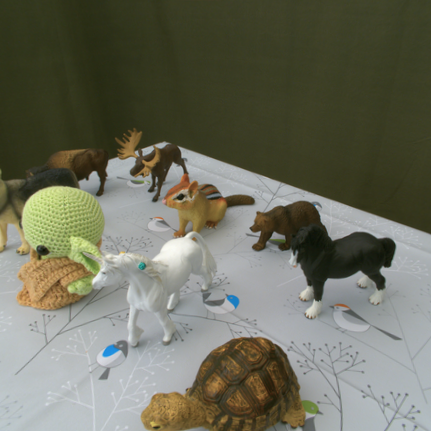}};
        \spy [red,magnification=2] on ($(img-20.north) + (-0.22\figurewidth,-0.88\figurewidth)$) in node [right] at ($(img-20.south) + (0.01\figurewidth,-0.68*0.36\figurewidth)$); 
        \spy [blue,magnification=1.5] on ($(img-20.north) + (-0.17\figurewidth, - 0.2\figurewidth)$) in node [left] at ($(img-20.south) - (0.01\figurewidth,0.68*0.36\figurewidth)$);

        \node [image,right=of img-20,xshift=0.07\figurewidth] (img-21) {\includegraphics[width=\figurewidth]{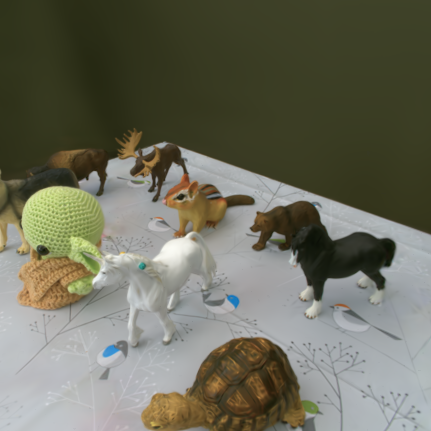}};
        \spy [red,magnification=2] on ($(img-21.north) + (-0.22\figurewidth,-0.88\figurewidth)$) in node [right] at ($(img-21.south) + (0.01\figurewidth,-0.68*0.36\figurewidth)$); 
        \spy [blue,magnification=1.5] on ($(img-21.north) + (-0.17\figurewidth, - 0.2\figurewidth)$) in node [left] at ($(img-21.south) - (0.01\figurewidth,0.68*0.36\figurewidth)$);

        \node [image,right=of img-21,xshift=0.07\figurewidth] (img-22) {\includegraphics[width=\figurewidth]{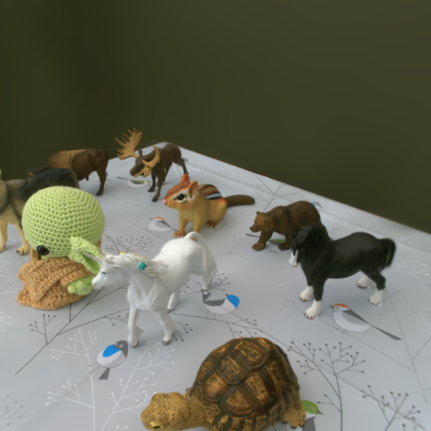}};
        \spy [red,magnification=2] on ($(img-22.north) + (-0.22\figurewidth,-0.88\figurewidth)$) in node [right] at ($(img-22.south) + (0.01\figurewidth,-0.68*0.36\figurewidth)$); 
        \spy [blue,magnification=1.5] on ($(img-22.north) + (-0.17\figurewidth, - 0.2\figurewidth)$) in node [left] at ($(img-22.south) - (0.01\figurewidth,0.68*0.36\figurewidth)$);
        
        \node [image,right=of img-22,xshift=0.07\figurewidth] (img-23) {\includegraphics[width=\figurewidth]{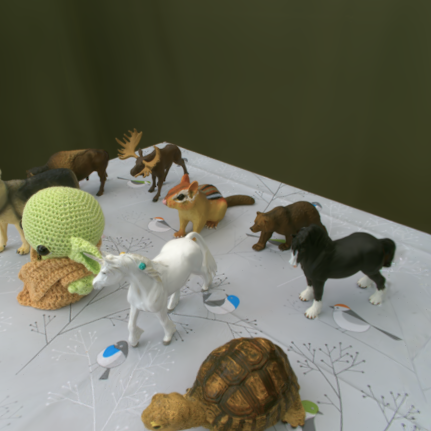}};
        \spy [red,magnification=2] on ($(img-23.north) + (-0.22\figurewidth,-0.88\figurewidth)$) in node [right] at ($(img-23.south) + (0.01\figurewidth,-0.68*0.36\figurewidth)$); 
        \spy [blue,magnification=1.5] on ($(img-23.north) + (-0.17\figurewidth, - 0.2\figurewidth)$) in node [left] at ($(img-23.south) - (0.01\figurewidth,0.68*0.36\figurewidth)$);

        \node [image,right=of img-23,xshift=0.07\figurewidth] (img-24) {\includegraphics[width=\figurewidth]{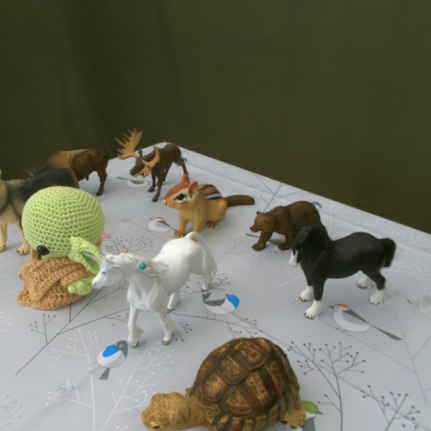}};
        \spy [red,magnification=2] on ($(img-24.north) + (-0.22\figurewidth,-0.88\figurewidth)$) in node [right] at ($(img-24.south) + (0.01\figurewidth,-0.68*0.36\figurewidth)$); 
        \spy [blue,magnification=1.5] on ($(img-24.north) + (-0.17\figurewidth, - 0.2\figurewidth)$) in node [left] at ($(img-24.south) - (0.01\figurewidth,0.68*0.36\figurewidth)$);
        
        \node[label] (label1) at (img-00.north) {Ground Truth};
        \node[label] (label2) at (img-01.north) {Splatfacto};
        \node[label] (label3) at (img-02.north) {Splatfacto-W-T \cite{xu2024splatfactow}};
        \node[label] (label4) at (img-03.north) {SpotlessSplats \cite{sabour2024spotlesssplats}};
        \node[label] (label5) at (img-04.north) {\ours (ours)};

        \node[label,rotate=90] (scene1) at (img-00.west) {\itshape Crab (1)};
        \node[label,rotate=90] (scene2) at (img-10.west) {\itshape Crab (2)};
        \node[label,rotate=90] (scene3) at (img-20.west) {\itshape Yoda };
    \end{tikzpicture}
    \caption{\textbf{Additional qualitative results on the RobustNeRF data set \cite{sabour2023robustnerf}.} In the {\it Crab~(1)}, \textit{Crab~(2)} and {\it Statue} scenes, \ours reconstructs static objects and backgrounds accurately. 
    }
    \label{fig:qualitative-robustnerf-all}
\end{figure*}

\paragraph{Background Model} 
For background modelling, we follow the approach introduced in Splatfacto-W \cite{xu2024splatfactow}. Instead of previous methods that use a unified colour for the background, we leverage the same per-image embeddings used for appearance modelling to predict the Spherical Harmonics (SH) coefficients $b$ for the background using an MLP. Prior to applying alpha blending, we render the background RGB colour $\vc_\text{BG}$ from the predicted SH coefficients:
\begin{equation}
b=f_\text{BG}(\ve_j).
\end{equation}
To encourage the opacity of both distractor and static Gaussians corresponding to these areas to be low, we also apply the opacity regularization from Splatfacto-W \cite{xu2024splatfactow} for both distractor Gaussians and static Gaussians:
\begin{equation}
\mathcal{L}_\text{bg} = \sum_{\br \in \mcP} |\alpha_d(\vr) + (1-\alpha_d(\vr))\cdot \alpha_s(\vr))|,
\end{equation}
where $\alpha(\vr)$ is the per-pixel accumulation of the Gaussians at pixel $\vr$. The set $\mcP$ is defined as $\mcP = \{(\vr) \mid M(\vr) > 0.6\}$. 

Similar to Splatfacto-W~\cite{xu2024splatfactow}, we also use a per-pixel mask $M$ to filter the area where the error of predicted background colour and ground truth $\epsilon(\vr)$ is smaller than a threshold $\mathcal{T_\epsilon}$, with a box filter $\mathcal{B}_{3\times3}$ to force the smoothness on the background
\begin{align}
    \Tilde{e}(\vr) & = \epsilon(\vr) \leq \mathcal{T_\epsilon}, \\
    M(\vr) & = (\Tilde{e}(\vr) \ast \mathcal{B}_{3\times3}) .
\end{align}
The final loss function of our method on Photo Tourism data set is:
\begin{equation}
\mathcal{L} = \mathcal{L}_\text{GS} + \lambda_d  \alpha_d + \lambda_\text{bg} \mathcal{L}_\text{bg}.
\end{equation}



\section{Additional Experimental Settings}
\label{app:experimental_settings}

\subsection{Data Sets}

\paragraph{RobustNeRF \cite{sabour2023robustnerf}} 
We run experiments on all five scenes \textit{Statue}, \textit{Android}, \textit{Crab (1)}, \textit{Crab (2)}, and \textit{Yoda}. All images are downscaled $8\times$ as instructed in~\cite{sabour2023robustnerf}. 
The scenes are indoors and are captured under different occluder settings. The \textit{Crab(2)} and \textit{Yoda} scenes include both clean and cluttered image variants from the same viewpoint, enabling us to perform ablation studies by varying the ratio of clean to cluttered training images (see \cref{fig:ablation_ratio_and_initialization}(left)). 

\begin{figure*}[t!]
    \centering
    \setlength{\figurewidth}{0.135\textwidth}
    \begin{tikzpicture}[
        image/.style = {inner sep=0pt, outer sep=1pt, minimum width=\figurewidth, anchor=north west, text width=\figurewidth}, 
        node distance = 1pt and 1pt, every node/.style={font= {\tiny}}, 
        label/.style = {font={\footnotesize\bf\vphantom{p}},anchor=south,inner sep=0pt}, 
        spy using outlines={rectangle, magnification=2, size=0.36\figurewidth},
        ]

        \node [image] (img-00) {\includegraphics[width=\figurewidth, height=0.75\figurewidth, keepaspectratio=false, clip]{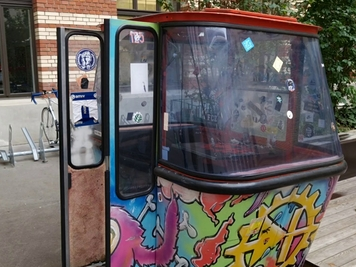}
        };
        \spy [red,magnification=1.5] on ($(img-00) + (-0.37\figurewidth, 0.025\figurewidth)$) in node [right] at ($(img-00.east) + (0,0.52*0.36\figurewidth)$);
        \spy [blue] on ($(img-00) - (0.28\figurewidth,0.28\figurewidth)$) in node [right] at ($(img-00.east) - (0,0.52*0.36\figurewidth)$);

        \node [image,right=of img-00,xshift=0.42\figurewidth] (img-01) {\includegraphics[width=\figurewidth]{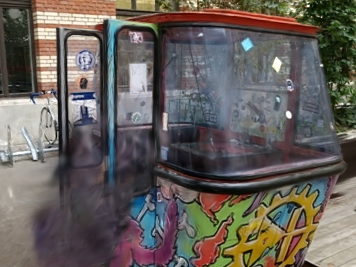}};

        \spy [red,magnification=1.5] on ($(img-01) + (-0.37\figurewidth, 0.025\figurewidth)$) in node [right] at ($(img-01.east) + (0,0.52*0.36\figurewidth)$);
        \spy [blue] on ($(img-01) - (0.28\figurewidth,0.28\figurewidth)$) in node [right] at ($(img-01.east) - (0,0.52*0.36\figurewidth)$);

        \node [image,right=of img-01,xshift=0.42\figurewidth] (img-02) {\includegraphics[width=\figurewidth]{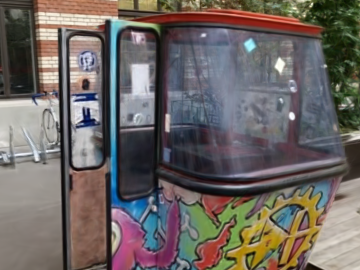}};

        \spy [red,magnification=1.5] on ($(img-02) + (-0.37\figurewidth, 0.025\figurewidth)$) in node [right] at ($(img-02.east) + (0,0.52*0.36\figurewidth)$);
        \spy [blue] on ($(img-02) - (0.28\figurewidth,0.28\figurewidth)$) in node [right] at ($(img-02.east) - (0,0.52*0.36\figurewidth)$);

        \node [image,right=of img-02,xshift=0.42\figurewidth] (img-03) {\includegraphics[width=\figurewidth]{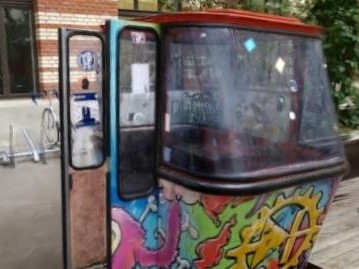}};

        \spy [red,magnification=1.5] on ($(img-03) + (-0.37\figurewidth, 0.025\figurewidth)$) in node [right] at ($(img-03.east) + (0,0.52*0.36\figurewidth)$);
        \spy [blue] on ($(img-03) - (0.28\figurewidth,0.28\figurewidth)$) in node [right] at ($(img-03.east) - (0,0.52*0.36\figurewidth)$); 
        
        \node [image,right=of img-03,xshift=0.42\figurewidth] (img-04) {\includegraphics[width=\figurewidth]{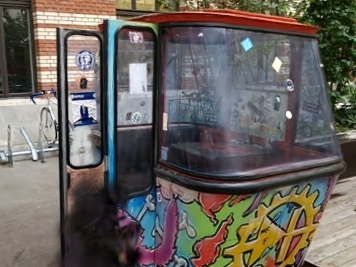}};

        \spy [red,magnification=1.5] on ($(img-04) + (-0.37\figurewidth, 0.025\figurewidth)$) in node [right] at ($(img-04.east) + (0,0.52*0.36\figurewidth)$);
        \spy [blue] on ($(img-04) - (0.28\figurewidth,0.28\figurewidth)$) in node [right] at ($(img-04.east) - (0,0.52*0.36\figurewidth)$);

        \node [image, below=0.05\figurewidth of img-00] (img-10) {\includegraphics[width=\figurewidth]{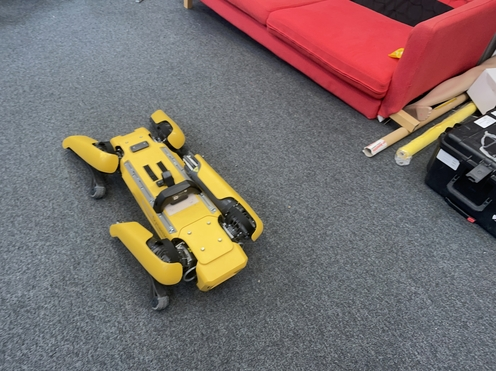}};
        \spy [red] on ($(img-10) + (0.13\figurewidth,0.26\figurewidth)$) in node [right] at ($(img-10.east) + (0,0.52*0.36\figurewidth)$);  
        \spy [blue] on ($(img-10) + (-0.35\figurewidth, - 0.27\figurewidth)$) in node [right] at ($(img-10.east) - (0,0.52*0.36\figurewidth)$);

        \node [image,right=of img-10,xshift=0.42\figurewidth] (img-11) {\includegraphics[width=\figurewidth]{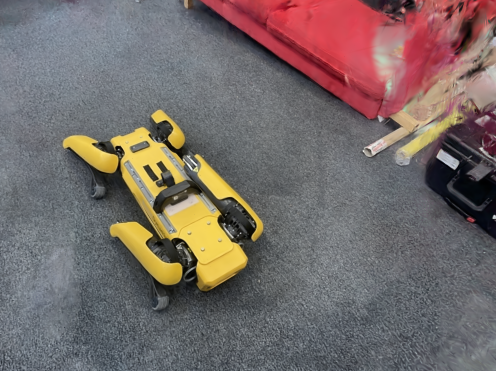}};
        \spy [red] on ($(img-11) + (0.13\figurewidth,0.26\figurewidth)$) in node [right] at ($(img-11.east) + (0,0.52*0.36\figurewidth)$);  
        \spy [blue] on ($(img-11) + (-0.35\figurewidth, - 0.27\figurewidth)$) in node [right] at ($(img-11.east) - (0,0.52*0.36\figurewidth)$);

        \node [image,right=of img-11,xshift=0.42\figurewidth] (img-12) {\includegraphics[width=\figurewidth]{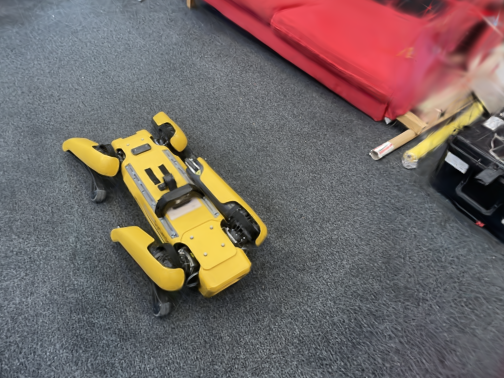}};
        \spy [red] on ($(img-12) + (0.13\figurewidth,0.26\figurewidth)$) in node [right] at ($(img-12.east) + (0,0.52*0.36\figurewidth)$);  
        \spy [blue] on ($(img-12) + (-0.35\figurewidth, - 0.27\figurewidth)$) in node [right] at ($(img-12.east) - (0,0.52*0.36\figurewidth)$);

        \node [image,right=of img-12,xshift=0.42\figurewidth] (img-13) {\includegraphics[width=\figurewidth]{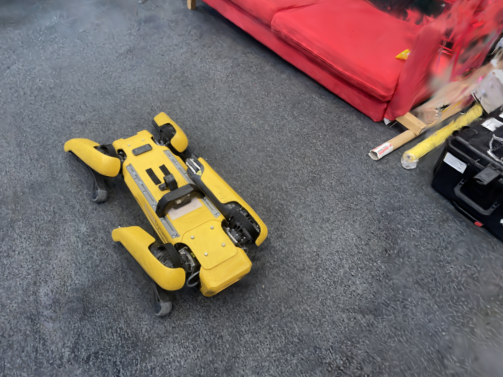}};
        \spy [red] on ($(img-13) + (0.13\figurewidth,0.26\figurewidth)$) in node [right] at ($(img-13.east) + (0,0.52*0.36\figurewidth)$); 
        \spy [blue] on ($(img-13) + (-0.35\figurewidth, - 0.27\figurewidth)$) in node [right] at ($(img-13.east) - (0,0.52*0.36\figurewidth)$);

        \node [image,right=of img-13,xshift=0.42\figurewidth] (img-14) {\includegraphics[width=\figurewidth]{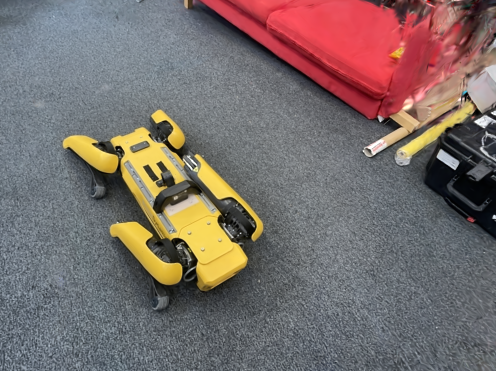}};
        \spy [red] on ($(img-14) + (0.13\figurewidth,0.26\figurewidth)$) in node [right] at ($(img-14.east) + (0,0.52*0.36\figurewidth)$); 
        \spy [blue] on ($(img-14) + (-0.35\figurewidth, - 0.27\figurewidth)$) in node [right] at ($(img-14.east) - (0,0.52*0.36\figurewidth)$);

        
        \node [image, below=0.05\figurewidth of img-10] (img-20) {\includegraphics[width=\figurewidth]{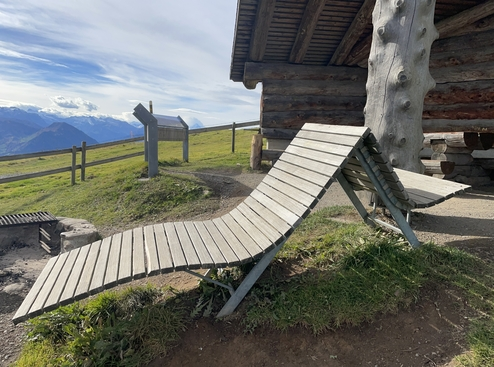}};
        \spy [red,magnification=2.5] on ($(img-20) + (0.0\figurewidth,0.18\figurewidth)$) in node [right] at ($(img-20.east) + (0,0.52*0.36\figurewidth)$);  
        \spy [blue] on ($(img-20) + (0.40\figurewidth, + 0.23\figurewidth)$) in node [right] at ($(img-20.east) - (0,0.52*0.36\figurewidth)$);

        \node [image,right=of img-20,xshift=0.42\figurewidth] (img-21) {\includegraphics[width=\figurewidth]{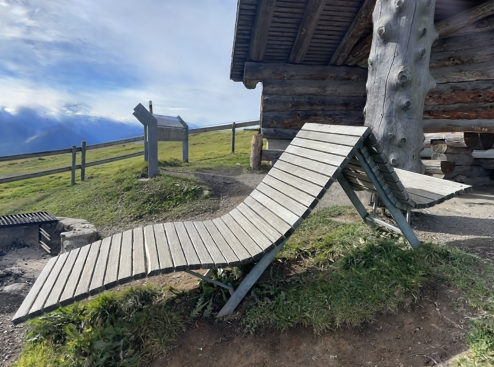}};
        \spy [red,magnification=2.5] on ($(img-21) + (0.0\figurewidth,0.18\figurewidth)$) in node [right] at ($(img-21.east) + (0,0.52*0.36\figurewidth)$);  
        \spy [blue] on ($(img-21) + (0.40\figurewidth, + 0.23\figurewidth)$) in node [right] at ($(img-21.east) - (0,0.52*0.36\figurewidth)$);

        \node [image,right=of img-21,xshift=0.42\figurewidth] (img-22) {\includegraphics[width=\figurewidth]{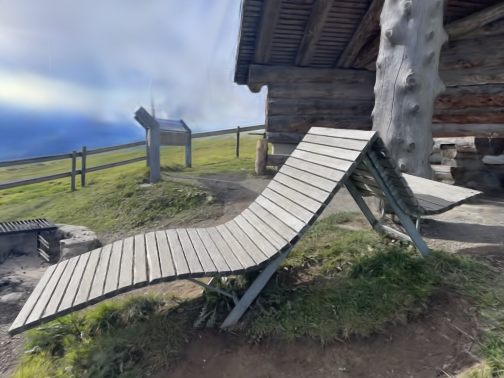}};
        \spy [red,magnification=2.5] on ($(img-22) + (0.0\figurewidth,0.18\figurewidth)$) in node [right] at ($(img-22.east) + (0,0.52*0.36\figurewidth)$);  
        \spy [blue] on ($(img-22) + (0.40\figurewidth, + 0.23\figurewidth)$) in node [right] at ($(img-22.east) - (0,0.52*0.36\figurewidth)$);

        \node [image,right=of img-22,xshift=0.42\figurewidth] (img-23) {\includegraphics[width=\figurewidth]{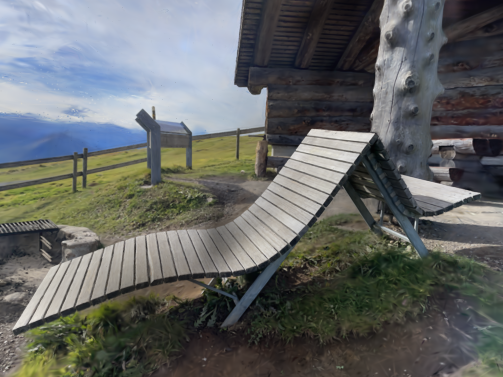}};
        \spy [red,magnification=2.5] on ($(img-23) + (0.0\figurewidth,0.18\figurewidth)$) in node [right] at ($(img-23.east) + (0,0.52*0.36\figurewidth)$); 
        \spy [blue] on ($(img-23) + (0.40\figurewidth, + 0.23\figurewidth)$) in node [right] at ($(img-23.east) - (0,0.52*0.36\figurewidth)$);

        \node [image,right=of img-23,xshift=0.42\figurewidth] (img-24) {\includegraphics[width=\figurewidth]{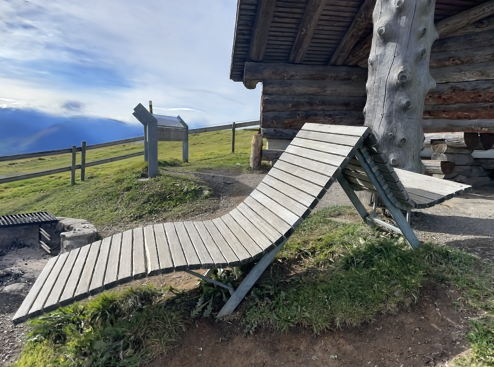}};
        \spy [red,magnification=2.5] on ($(img-24) + (0.0\figurewidth,0.18\figurewidth)$) in node [right] at ($(img-24.east) + (0,0.52*0.36\figurewidth)$); 
        \spy [blue] on ($(img-24) + (0.40\figurewidth, + 0.23\figurewidth)$) in node [right] at ($(img-24.east) - (0,0.52*0.36\figurewidth)$);


        \node [image, below=0.05\figurewidth of img-20] (img-30) {\includegraphics[width=\figurewidth]{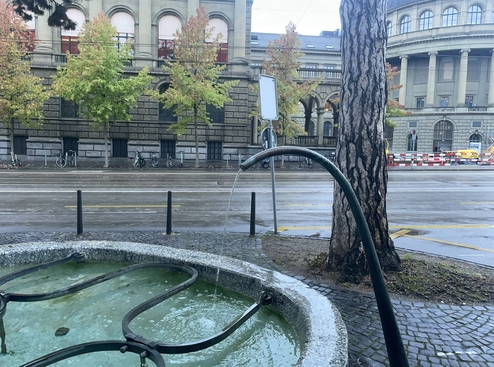}};
        \spy [red,magnification=2.5] on ($(img-30) + (0.38\figurewidth,0.20\figurewidth)$) in node [right] at ($(img-30.east) + (0,0.52*0.36\figurewidth)$);  
        \spy [blue] on ($(img-30) + (0.35\figurewidth, 0\figurewidth)$) in node [right] at ($(img-30.east) - (0,0.52*0.36\figurewidth)$);

        \node [image,right=of img-30,xshift=0.42\figurewidth] (img-31) {\includegraphics[width=\figurewidth]{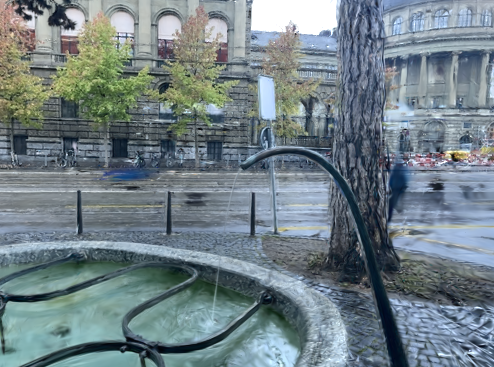}};
        \spy [red,magnification=2.5] on ($(img-31) + (0.38\figurewidth,0.20\figurewidth)$) in node [right] at ($(img-31.east) + (0,0.52*0.36\figurewidth)$);  
        \spy [blue] on ($(img-31) + (0.35\figurewidth, 0\figurewidth)$) in node [right] at ($(img-31.east) - (0,0.52*0.36\figurewidth)$);

        \node [image,right=of img-31,xshift=0.42\figurewidth] (img-32) {\includegraphics[width=\figurewidth]{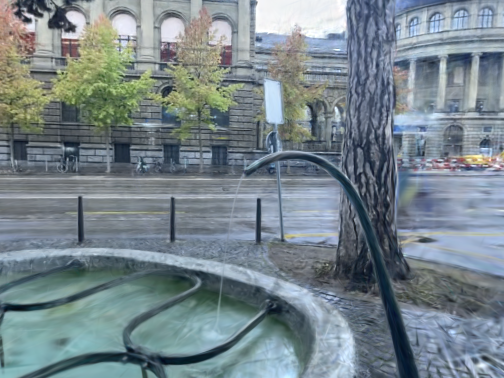}};
        \spy [red,magnification=2.5] on ($(img-32) + (0.38\figurewidth,0.20\figurewidth)$) in node [right] at ($(img-32.east) + (0,0.52*0.36\figurewidth)$);  
        \spy [blue] on ($(img-32) + (0.35\figurewidth, 0\figurewidth)$) in node [right] at ($(img-32.east) - (0,0.52*0.36\figurewidth)$);

        \node [image,right=of img-32,xshift=0.42\figurewidth] (img-33) {\includegraphics[width=\figurewidth]{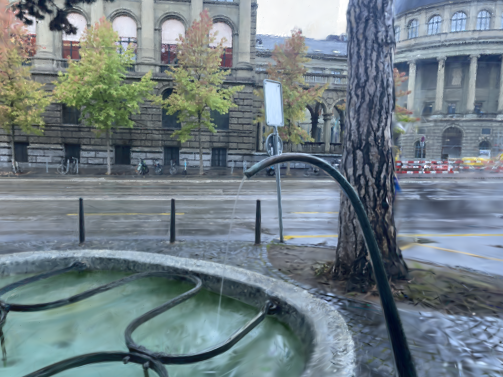}};
        \spy [red,magnification=2.5] on ($(img-33) + (0.38\figurewidth,0.20\figurewidth)$) in node [right] at ($(img-33.east) + (0,0.52*0.36\figurewidth)$); 
        \spy [blue] on ($(img-33) + (0.35\figurewidth, 0\figurewidth)$) in node [right] at ($(img-33.east) - (0,0.52*0.36\figurewidth)$);

        \node [image,right=of img-33,xshift=0.42\figurewidth] (img-34) {\includegraphics[width=\figurewidth]{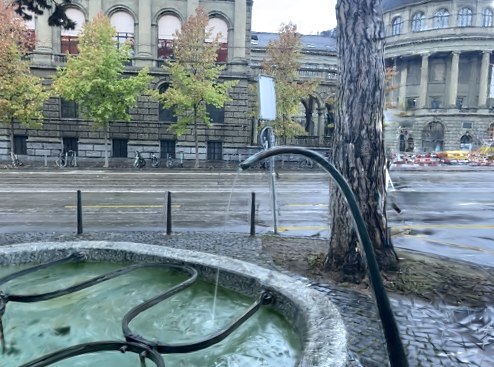}};
        \spy [red,magnification=2.5] on ($(img-34) + (0.38\figurewidth,0.20\figurewidth)$) in node [right] at ($(img-34.east) + (0,0.52*0.36\figurewidth)$); 
        \spy [blue] on ($(img-34) + (0.35\figurewidth, 0\figurewidth)$) in node [right] at ($(img-34.east) - (0,0.52*0.36\figurewidth)$);
        
        \node[label, xshift=0.25\figurewidth] (label1) at (img-00.north) {Ground Truth};
        \node[label, xshift=0.25\figurewidth] (label1) at (img-01.north) {Splatfacto};
        \node[label, xshift=0.25\figurewidth] (label2) at (img-02.north) {WildGaussians \cite{kulhanek2024wildgaussians}};
        \node[label, xshift=0.25\figurewidth] (label2) at (img-03.north) {SpotlessSplats \cite{sabour2024spotlesssplats}};
        \node[label, xshift=0.25\figurewidth] (label2) at (img-04.north) {\ours (ours)};

        \node[label,rotate=90] (scene1) at (img-00.west) {\itshape Patio};
        \node[label,rotate=90] (scene2) at (img-10.west) {\itshape Spot};
        \node[label,rotate=90] (scene3) at (img-20.west) {\itshape Mountain};
        \node[label,rotate=90] (scene4) at (img-30.west) {\itshape Fountain};
    \end{tikzpicture}
    \caption{\textbf{Additional qualitative results on On-the-go data set \cite{ren2024nerf}.} 
    }
    \label{fig:qualitative-onthego-all}
\end{figure*}
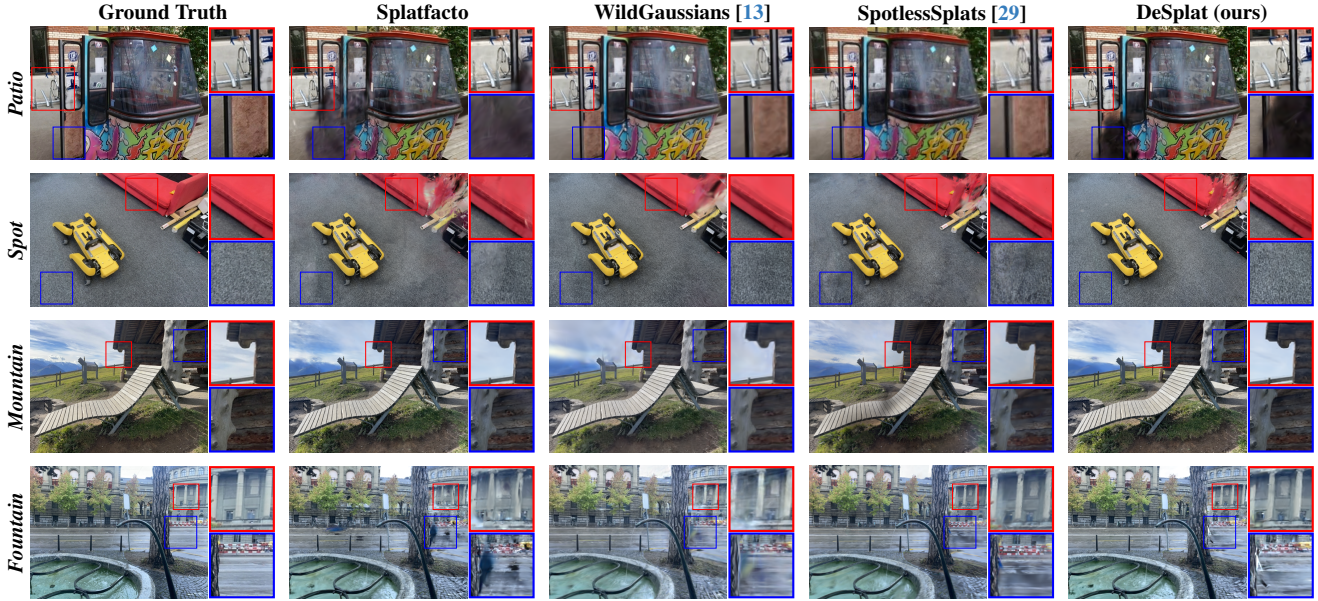

\begin{figure*}[h!]
    \centering
    \setlength{\figurewidth}{0.23\textwidth}
    \begin{tikzpicture}[
        image/.style = {inner sep=0pt, outer sep=1pt, minimum width=\figurewidth, anchor=north west, text width=\figurewidth}, 
        node distance = 1pt and 1pt, every node/.style={font= {\tiny}}, 
        label/.style = {font={\footnotesize\bf\vphantom{p}},anchor=south,inner sep=0pt}, 
        spy using outlines={rectangle, red, magnification=1.5, size=0.75cm, connect spies},
        ] 
        
        \node [image] (img-00) {\includegraphics[width=\figurewidth]{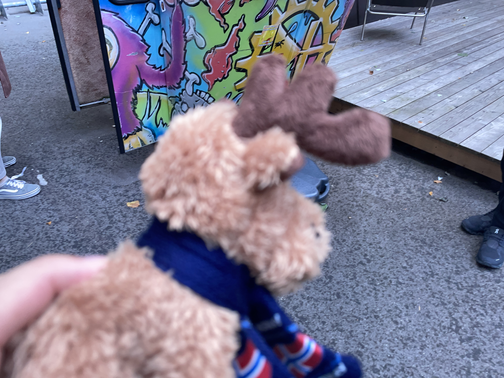}};
        \node [image,right=of img-00, xshift=0.06\figurewidth] (img-01) {\includegraphics[width=\figurewidth]{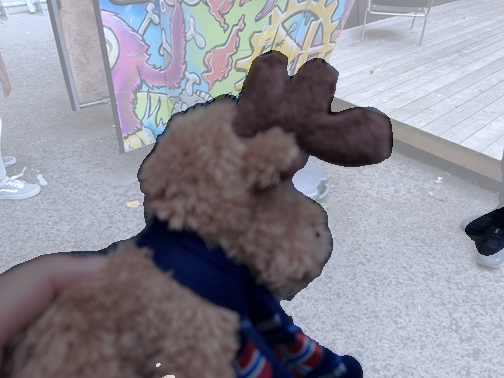}};
        \node [image,right=of img-01, xshift=0.06\figurewidth] (img-02) {\includegraphics[width=\figurewidth]{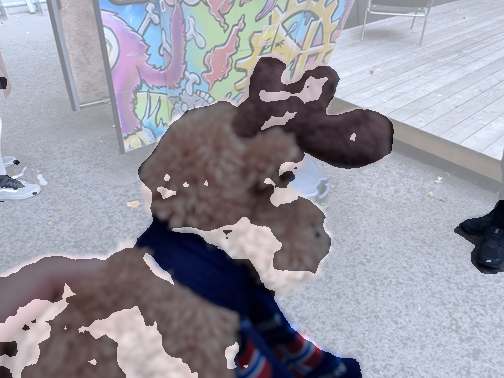}};
        \node [image,right=of img-02, xshift=0.06\figurewidth] (img-03) {\includegraphics[width=\figurewidth]{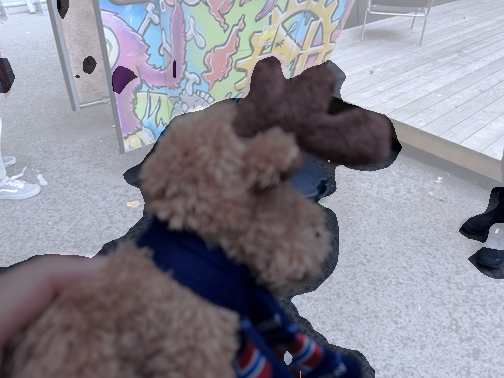}};

        \node [image, below=of img-00, yshift=-0.06\figurewidth] (img-10) {\includegraphics[width=\figurewidth]{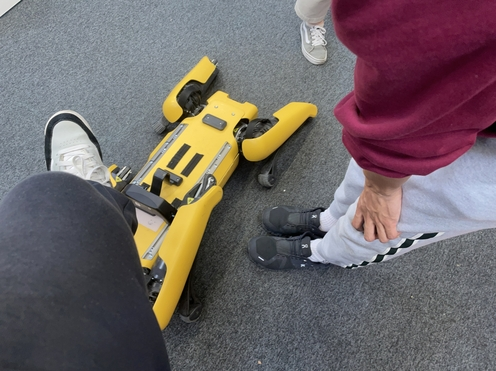}};
        \node [image,right=of img-10, xshift=0.06\figurewidth] (img-11) {\includegraphics[width=\figurewidth]{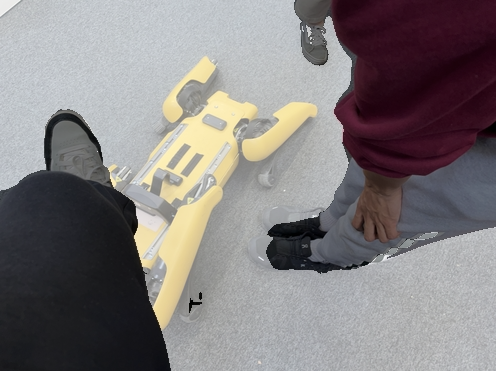}};
        \node [image,right=of img-11, xshift=0.06\figurewidth] (img-12) {\includegraphics[width=\figurewidth]{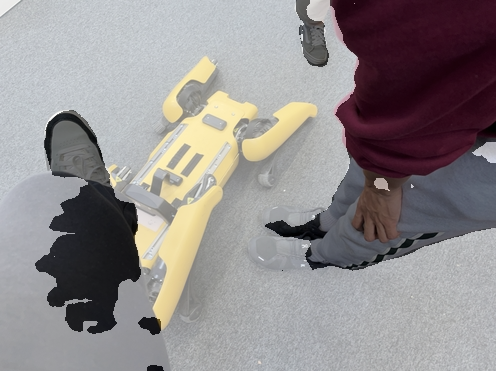}};
        \node [image,right=of img-12, xshift=0.06\figurewidth] (img-13) {\includegraphics[width=\figurewidth]{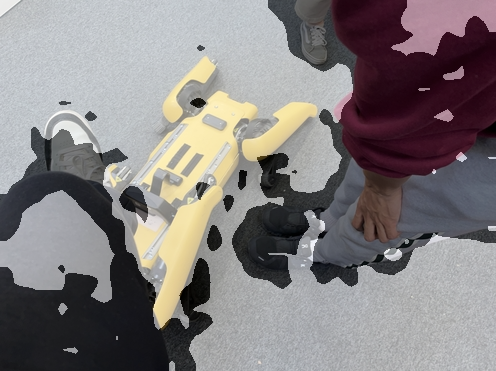}};
        
        \node[label] (label1) at (img-00.north) {Training Image};
        \node[label] (label2) at (img-01.north) {\ours (ours)};
        \node[label] (label3) at (img-02.north) {SLS-mlp};
        \node[label] (label4) at (img-03.north) {WildGaussians};

        \node[label,rotate=90] (scene1) at (img-00.west) {\itshape Patio High};
        \node[label,rotate=90] (scene2) at (img-10.west) {\itshape Spot};

    \end{tikzpicture}
    \vspace{-2mm}
    \caption{\textbf{Example distractor masks for different methods on On-the-go data set \cite{ren2024nerf}.}}
    \label{fig:mask}
    \vspace{-1em}
\end{figure*}

\paragraph{On-the-go \cite{ren2024nerf}} 
We run experiments on the scenes \textit{Mountain}, \textit{Fountain}, \textit{Corner}, \textit{Patio}, \textit{Spot}, and \textit{Patio-High} that are commonly used for reporting quantitative performance metrics~\cite{kulhanek2024wildgaussians,ren2024nerf,sabour2024spotlesssplats}. 
These scenes are further categorized according to three different occlusion rates: low (\textit{Mountain}, \textit{Fountain}), medium (\textit{Corner}, \textit{Patio}), and high (\textit{Spot}, \textit{Patio-High}) occlusion. 
All images are downscaled $8{\times}$, except the \textit{Patio} scene which is downscaled $4{\times}$. 
Furthermore, we use \textit{Arc de Triomphe} for visualization purposes (see \cref{fig:handling-occluders}). 

\paragraph{Photo Tourism \cite{snavely2006photo}} 
We run experiments on the scenes  \textit{Brandenburg Gate}, \textit{Sacre Coeur}, and \textit{Trevi Fountain} that are commonly used for reporting quantitative performance metrics~\cite{dahmani2024swag,kulhanek2024wildgaussians,xu2024splatfactow,xu2024wildgs,zhang2024gaussian}. 
The images for these scenes are complex, exhibiting different resolutions and varying illumination and weather effects, which necessitates view-dependent appearance modelling~\cite{martinbrualla2020nerfw}. 
We follow the evaluation protocol from \citet{martinbrualla2020nerfw} where a per-image embedding is optimized on the left half of the evaluation image and then evaluated on its right half. 
Furthermore, we follow the test-time optimization used by \citet{kulhanek2024wildgaussians} to optimize the per-image embeddings for the evaluation images. 
\cref{tab:phototourism} shows the reported metrics for the three scenes.  


\paragraph{Mip-NeRF 360 \cite{barron2022mipnerf360}}
We run experiments on seven scenes \textit{bicycle}, \textit{bonsai}, \textit{counter}, \textit{garden}, \textit{kitchen}, \textit{room} and \textit{stump} on Mip-NeRF 360 data set, as a comparison for performance on data sets that do not contain occluders. All indoor images are downscaled 2$\times$, and outdoor images are downscaled 4$\times$. 

\subsection{Baselines}

For all baselines, except Splatfacto and Splatfacto-W, we compare the performance of \ours against the PSNR, SSIM, and LPIPS metrics that were reported in their corresponding works to ensure consistency. 
For Splatfacto and Splatfacto-W, we run experiments for these using \texttt{gsplat} \cite{ye2024gsplat} version 1.0.0 and \texttt{nerfstudio} \cite{nerfstudio} version 1.1.4. 

For the qualitative results, we run experiments with SpotLessSplats~\cite{sabour2024spotlesssplats} and WildGaussians~\cite{kulhanek2024wildgaussians} to obtain visualizations (\eg, see \cref{fig:qualitative-onthego}). We run SpotLessSplats using the reimplementation~\footnote{\url{https://github.com/lilygoli/SpotLessSplats/tree/main}} in the \texttt{gsplat} library (version 1.1.1). For WildGaussians, we use the official codebase and trained model checkpoint files for evaluation and qualitative images.

\subsection{Implementation Details} 

\begin{table*}[th!]
    \scriptsize
    \centering
    \caption{\textbf{Performance comparison between our method and the baselines on the Photo Tourism data set}~\cite{snavely2006photo}. The \first{first}, \second{second}, and \colorbox{yellow!50!white}{third} values are highlighted. \ours perform significantly better than Splatfacto, indicating that the explicit scene separation is useful for these scenes. 
    }
    \vspace*{-.1in}
        \begin{tabular}{l|ccc|ccc|ccc}
        \toprule
        & \multicolumn{3}{c|}{\textbf{\itshape Brandenburg Gate}} & \multicolumn{3}{c|}{\textbf{\itshape Sacre Coeur}} & \multicolumn{3}{c}{\textbf{\itshape Trevi Fountain}} \\
        \textbf{Method} & \textbf{PSNR}$\uparrow$ & \textbf{SSIM}$\uparrow$ & \textbf{LPIPS}$\downarrow$ & \textbf{PSNR}$\uparrow$ & \textbf{SSIM}$\uparrow$ & \textbf{LPIPS}$\downarrow$ & \textbf{PSNR}$\uparrow$ & \textbf{SSIM}$\uparrow$ & \textbf{LPIPS}$\downarrow$ \\
        \midrule
        Splatfacto \cite{nerfstudio} & \val{white}{19.50} & \val{white}{0.885} & \val{white}{0.183} & \val{white}{17.15} & \val{white}{0.831} & \val{white}{0.210} & \val{white}{17.63} & \val{white}{0.696} & \val{white}{0.334} \\
        Splatfacto-W \cite{xu2024splatfactow} & \val{white}{26.87} & \val{second}{0.932} & \val{third}{0.124} & \val{white}{22.53} & \val{third}{0.876} & \val{white}{ 0.158} & \val{white}{22.66} & \val{white}{0.769} & \val{white}{0.224} \\
        Splatfacto-W-A \cite{xu2024splatfactow} & \val{white}{27.50} & \val{white}{0.930} & \val{white}{0.130} & \val{white}{22.62} & \val{third}{0.876} & \val{white}{0.156} & \val{white}{22.81} & \val{white}{0.770} & \val{white}{0.228} \\
        Splatfacto-W-T \cite{xu2024splatfactow} & \val{white}{26.16} & \val{white}{0.925} & \val{white}{0.131} & \val{third}{22.78} & \val{first}{0.878} & \val{third}{0.155} & \val{white}{22.88} & \val{white}{0.772} & \val{white}{0.228} \\
        SWAG \cite{dahmani2024swag} & \val{white}{26.33} & \val{white}{0.929} & \val{white}{0.139} & \val{white}{21.16} & \val{white}{0.860} & \val{white}{0.185} & \val{white}{23.10} & \val{first}{0.815} & \val{third}{0.208} \\
        GS-W \cite{zhang2024gaussian} & \val{second}{27.96} & \val{second}{0.932} & \val{first}{0.086} & \val{second}{23.24} & \val{white}{0.863} & \val{second}{0.130} & \val{white}{22.91} & \val{third}{0.801} & \val{first}{0.156} \\
        Wild-GS \cite{xu2024wildgs} & \val{first}{29.65} & \val{first}{0.933} & \val{second}{0.095} & \val{first}{24.99} & \val{first}{0.878} & \val{first}{0.127} & \val{first}{24.45} & \val{second}{0.808} & \val{second}{0.162} \\
        WildGaussians \cite{kulhanek2024wildgaussians} & \val{third}{27.77} & \val{white}{0.927} & \val{white}{0.133} & \val{white}{22.56} & \val{white}{0.859} & \val{white}{0.177} & \val{second}{23.63} & \val{white}{0.766} & \val{white}{0.228} \\
        \midrule
        \textbf{\ours (ours) - A} & \val{white}{26.72} & \val{white}{0.918} & \val{white}{0.132} & \val{white}{22.28} & \val{white}{0.876} & \val{white}{0.159} & \val{white}{23.06} & \val{white}{0.774} & \val{white}{0.229} \\
        \textbf{\ours (ours)} & \val{white}{25.04} & \val{white}{0.920} & \val{white}{0.142} & \val{white}{20.14} & \val{white}{0.868} & \val{white}{0.178} & \val{third}{23.31} & \val{white}{0.775} & \val{white}{0.226}\\
        \bottomrule
    \end{tabular}
    
    \label{tab:phototourism}
\end{table*}

\begin{figure*}[t]
    \centering
    \setlength{\figurewidth}{0.19\textwidth}
    \begin{tikzpicture}[
        image/.style = {inner sep=0pt, outer sep=1pt, minimum width=\figurewidth, anchor=north west, text width=\figurewidth}, 
        node distance = 1pt and 1pt, every node/.style={font= {\tiny}}, 
        label/.style = {font={\footnotesize\bf\vphantom{p}},anchor=south,inner sep=0pt}, 
        spy using outlines={rectangle, red, magnification=1.5, size=0.75cm, connect spies},
        ] 
        
        \node [image] (img-00) {\includegraphics[width=\figurewidth]{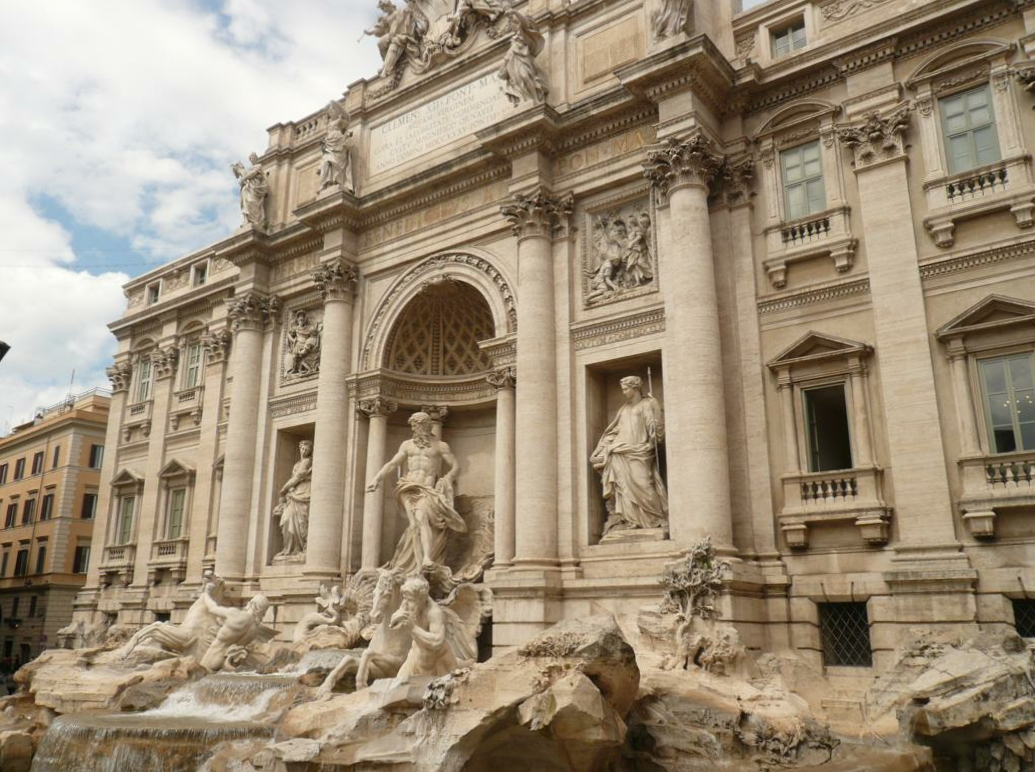}};
        \node [image,right=of img-00,xshift=1ex] (img-01) {\includegraphics[width=\figurewidth]{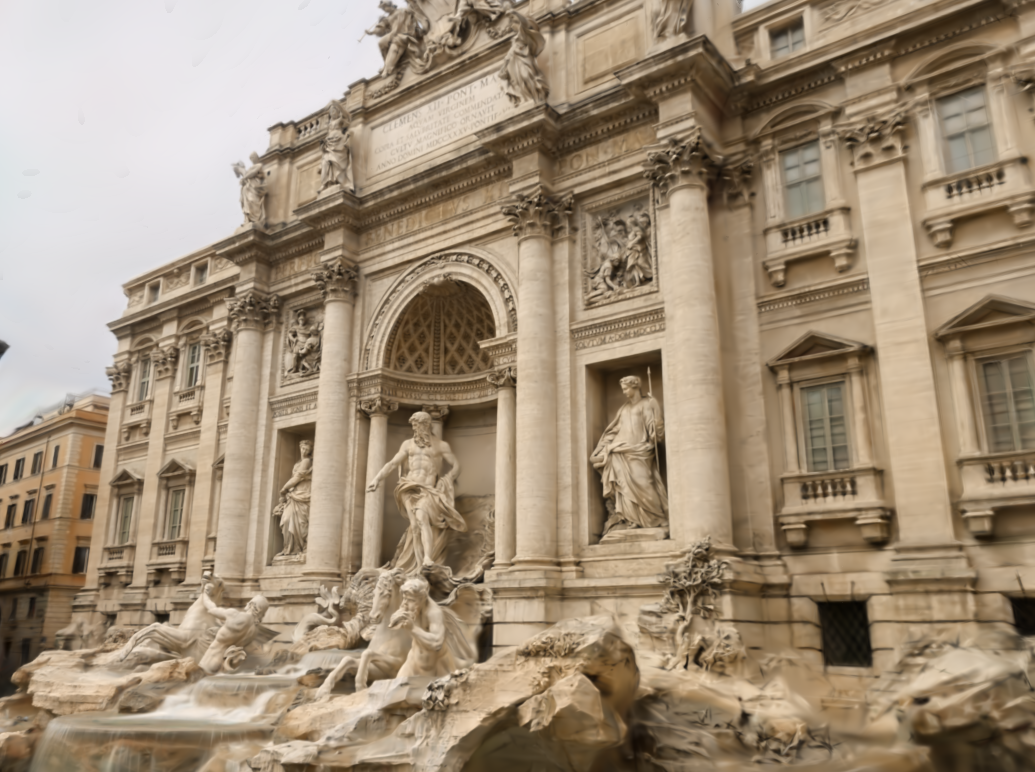}};
        \node [image,right=of img-01] (img-02) {\includegraphics[width=\figurewidth]{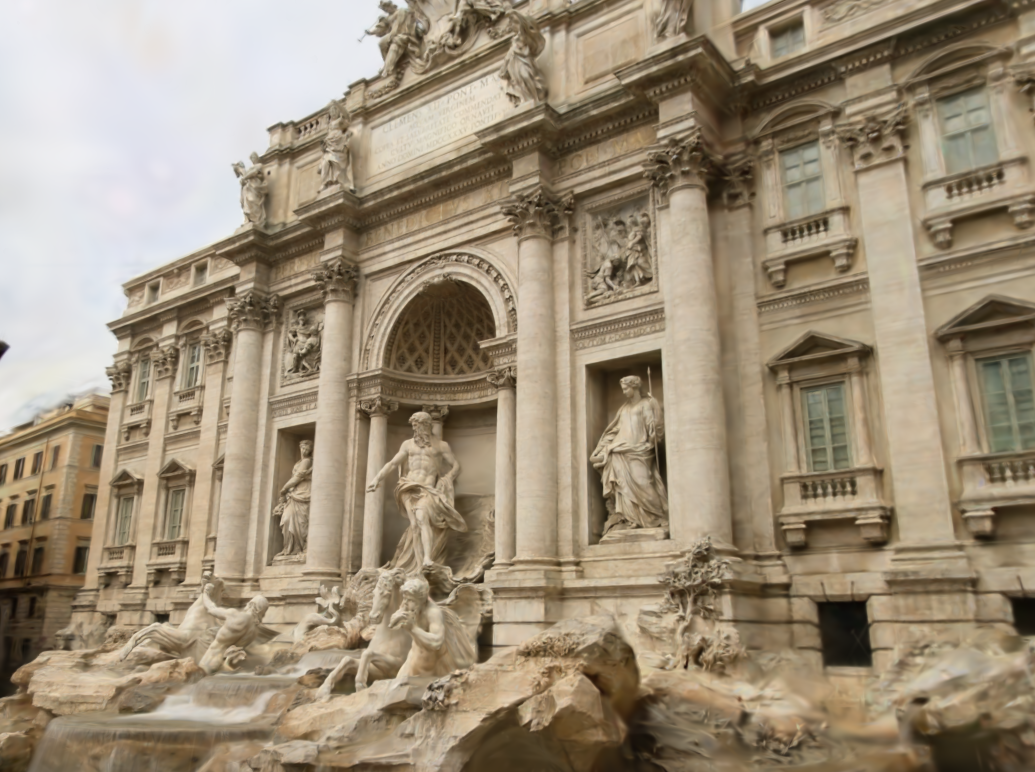}};
        \node [image,right=of img-02] (img-03) {\includegraphics[width=\figurewidth]{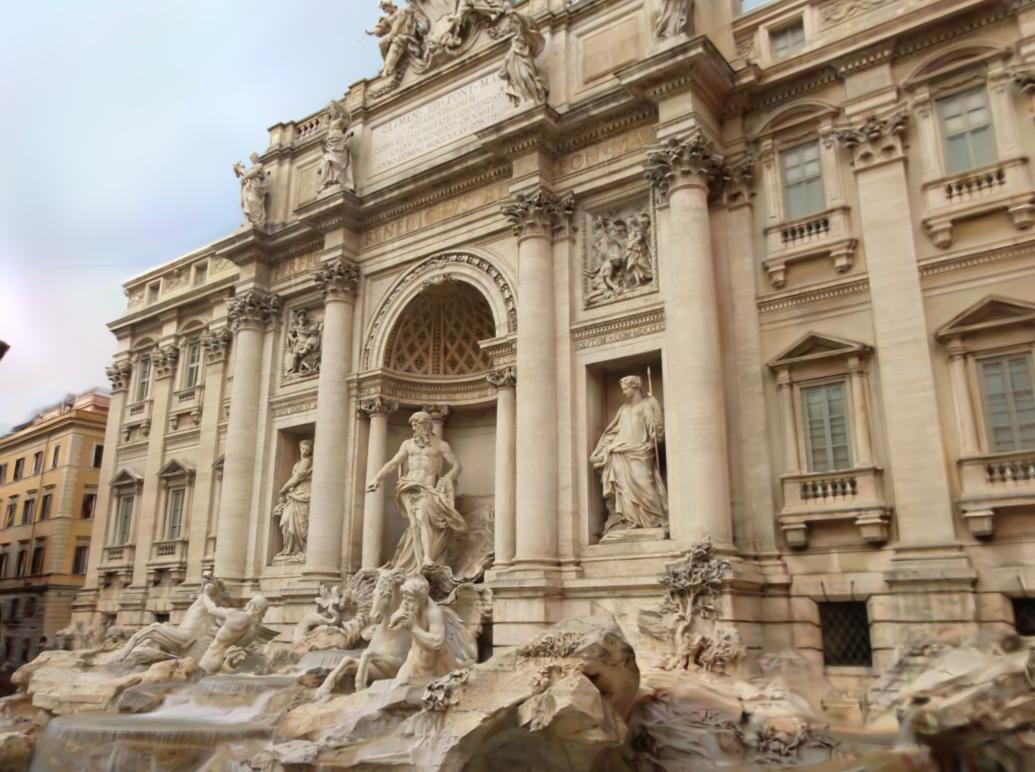}};
        \node [image,right=of img-03] (img-04) {\includegraphics[width=\figurewidth]{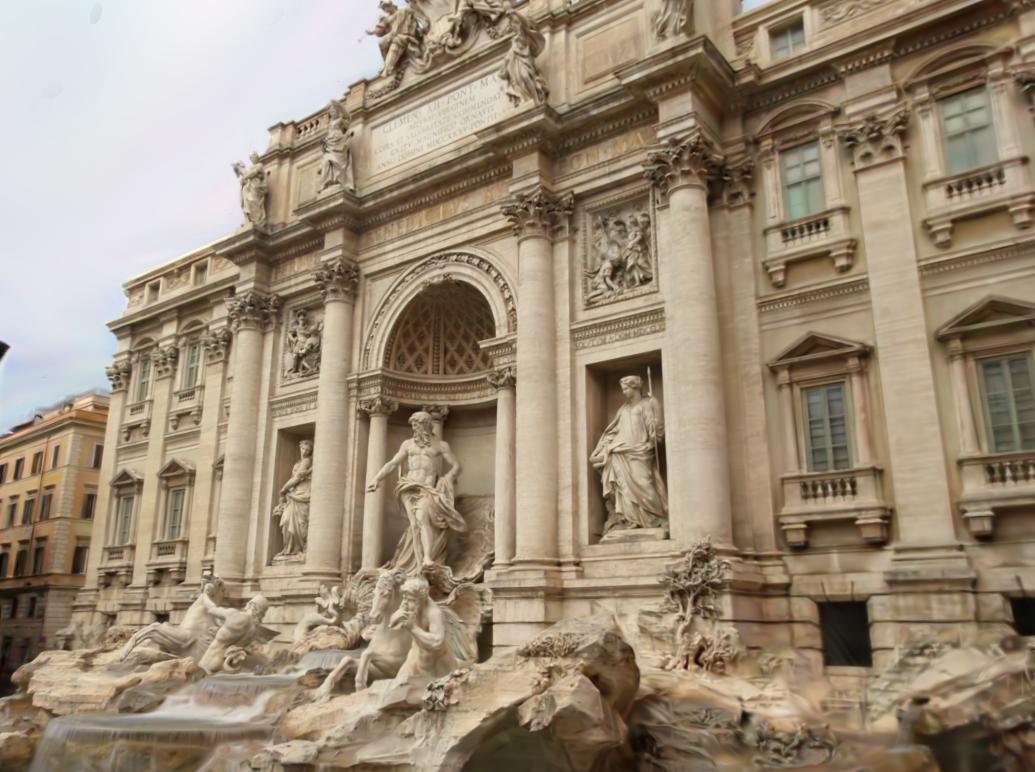}};
        
        \node [image,below=of img-00] (img-10) {\includegraphics[width=\figurewidth]{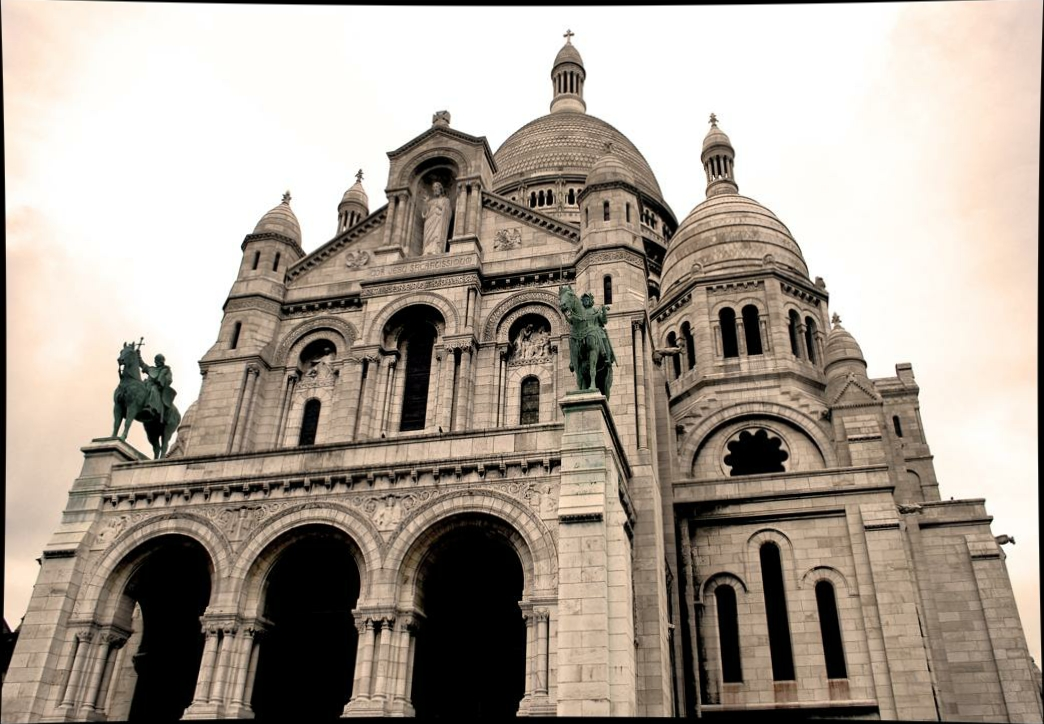}};
        \node [image,right=of img-10,xshift=1ex] (img-11) {\includegraphics[width=\figurewidth]{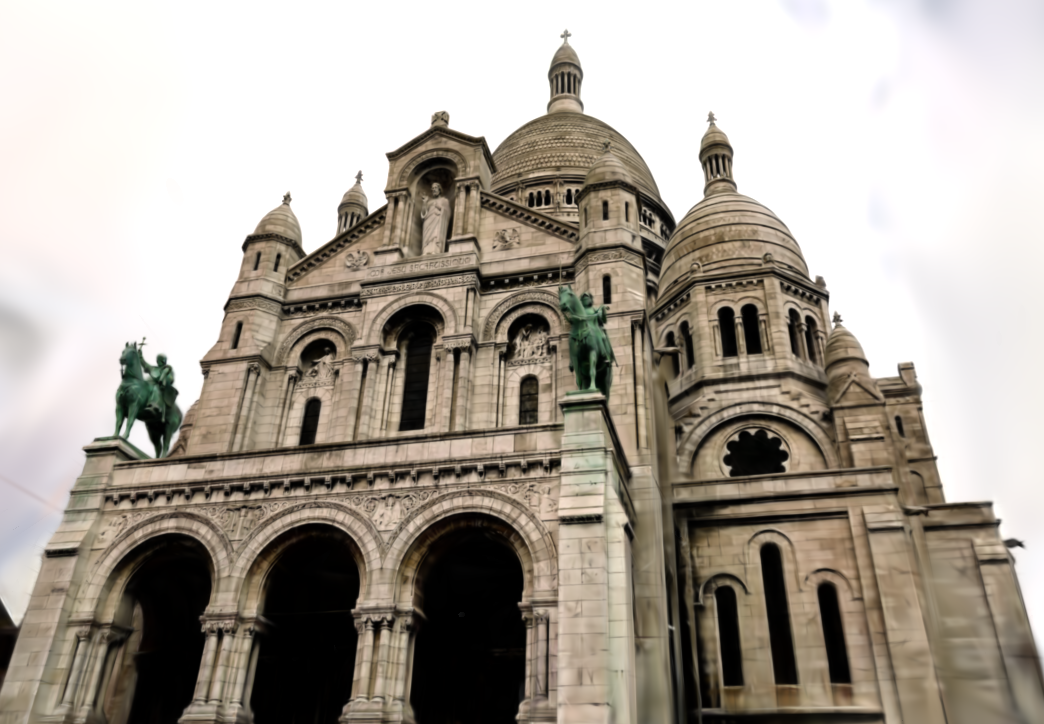}};
        \node [image,right=of img-11] (img-12) {\includegraphics[width=\figurewidth]{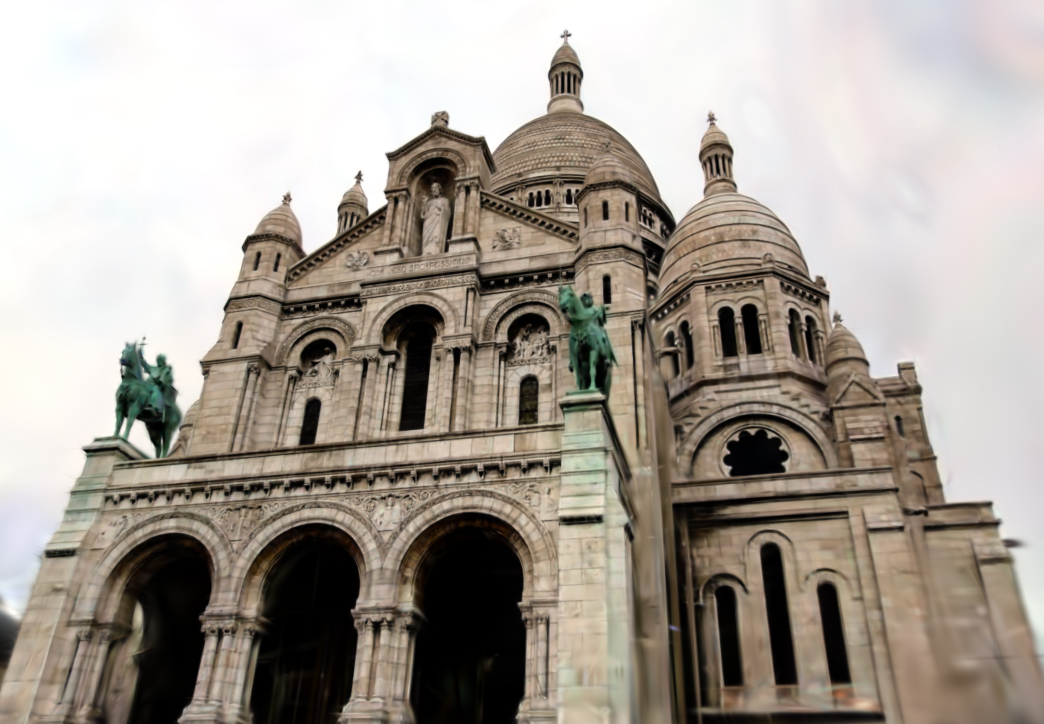}};
        \node [image,right=of img-12] (img-13) {\includegraphics[width=\figurewidth]{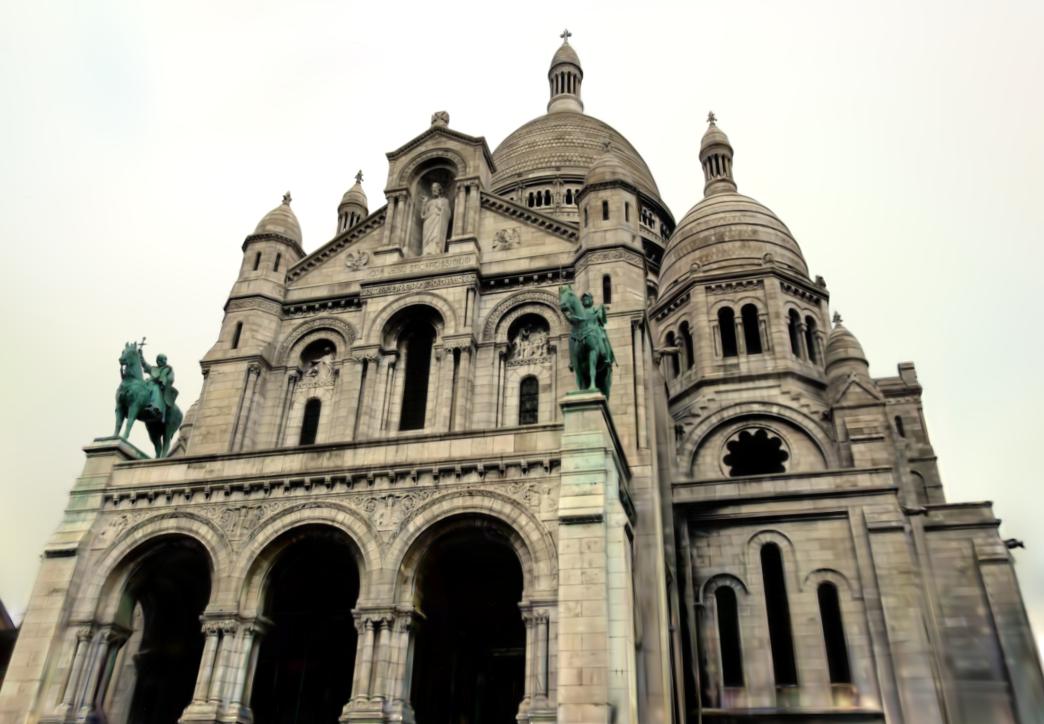}};
        \node [image,right=of img-13] (img-14) {\includegraphics[width=\figurewidth]{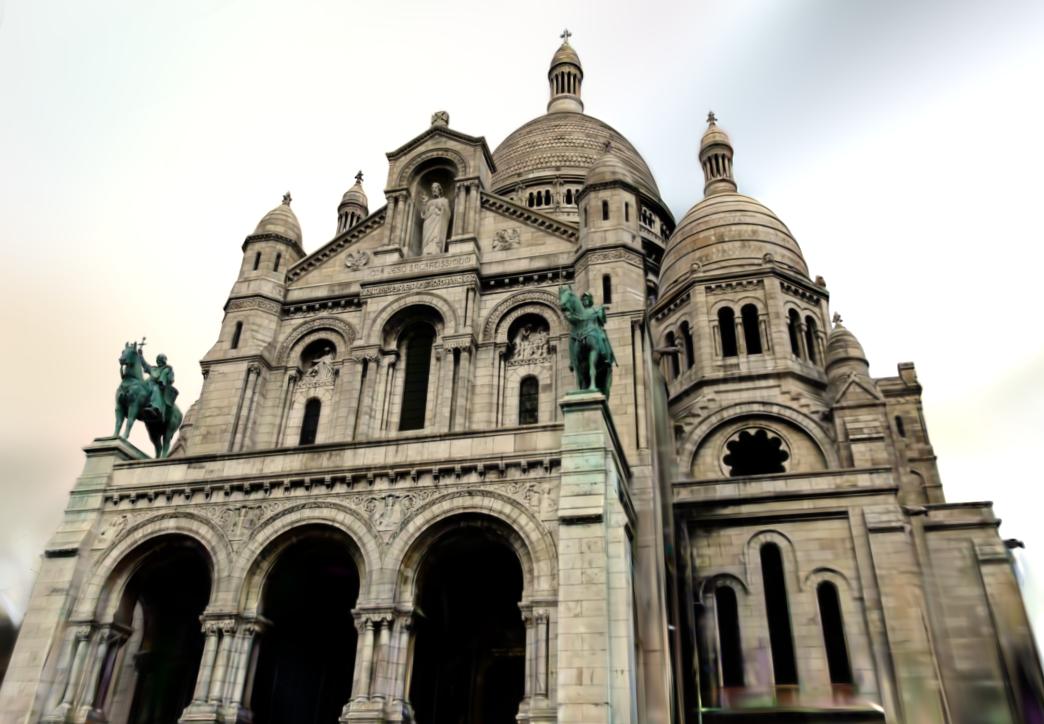}};
        
        \node [image,below=of img-10] (img-20) {\includegraphics[width=\figurewidth]{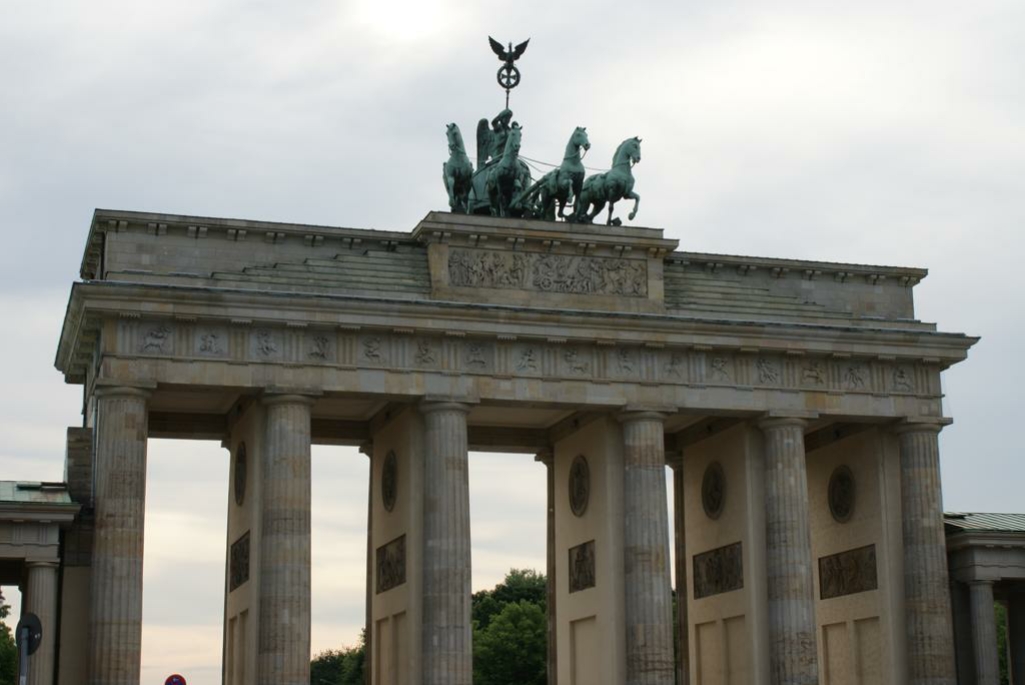}};
        \node [image,right=of img-20,xshift=1ex] (img-21) {\includegraphics[width=\figurewidth]{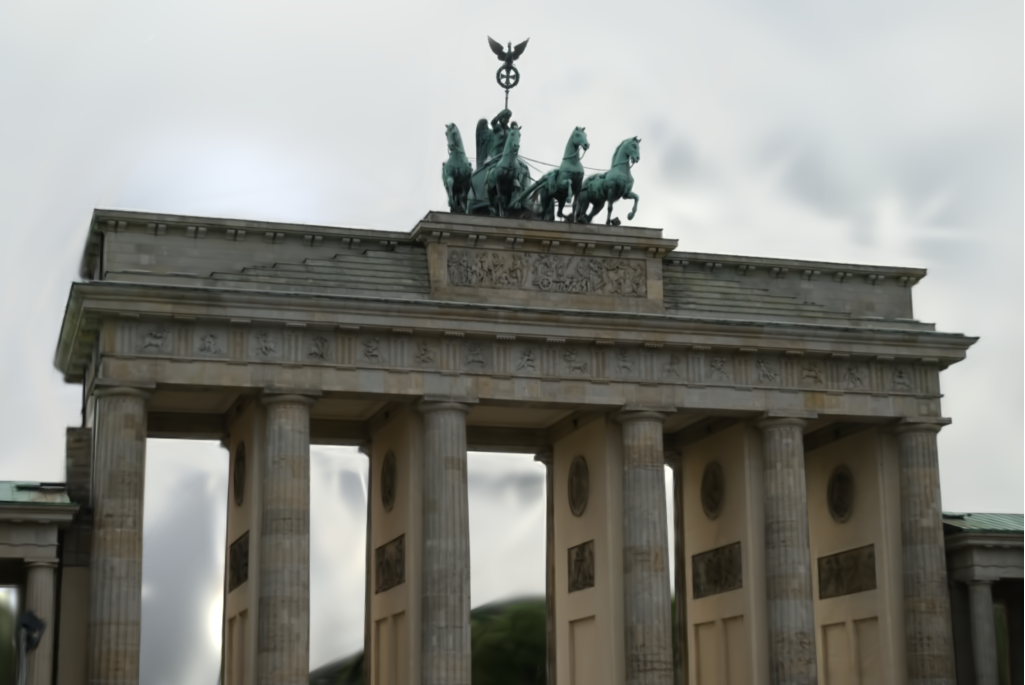}};
        \node [image,right=of img-21] (img-22) {\includegraphics[width=\figurewidth]{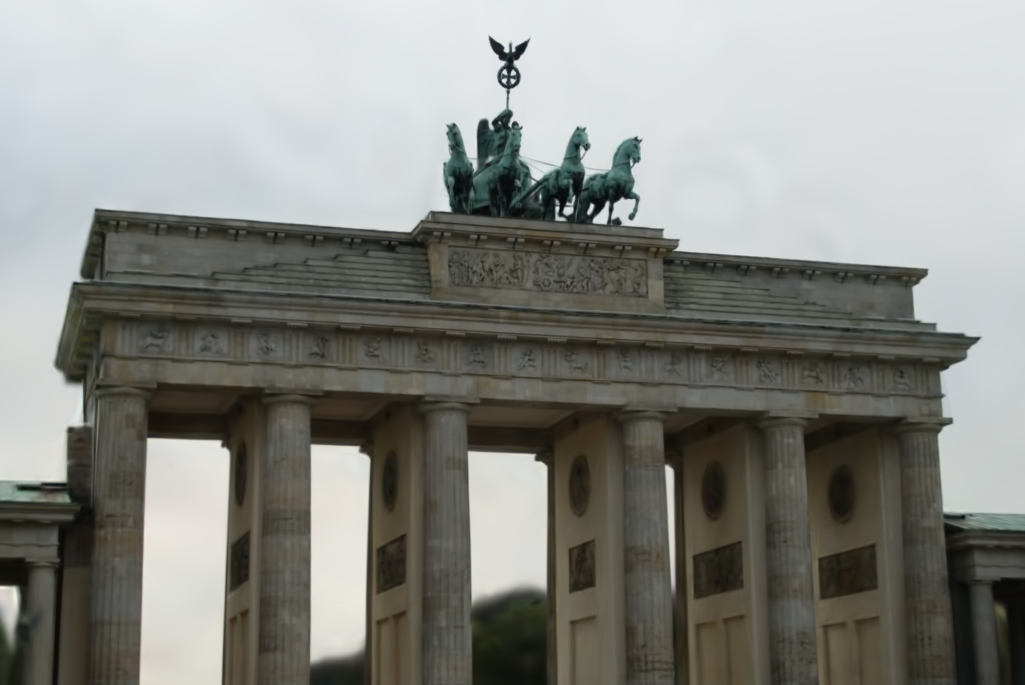}};
        \node [image,right=of img-22] (img-23) {\includegraphics[width=\figurewidth]{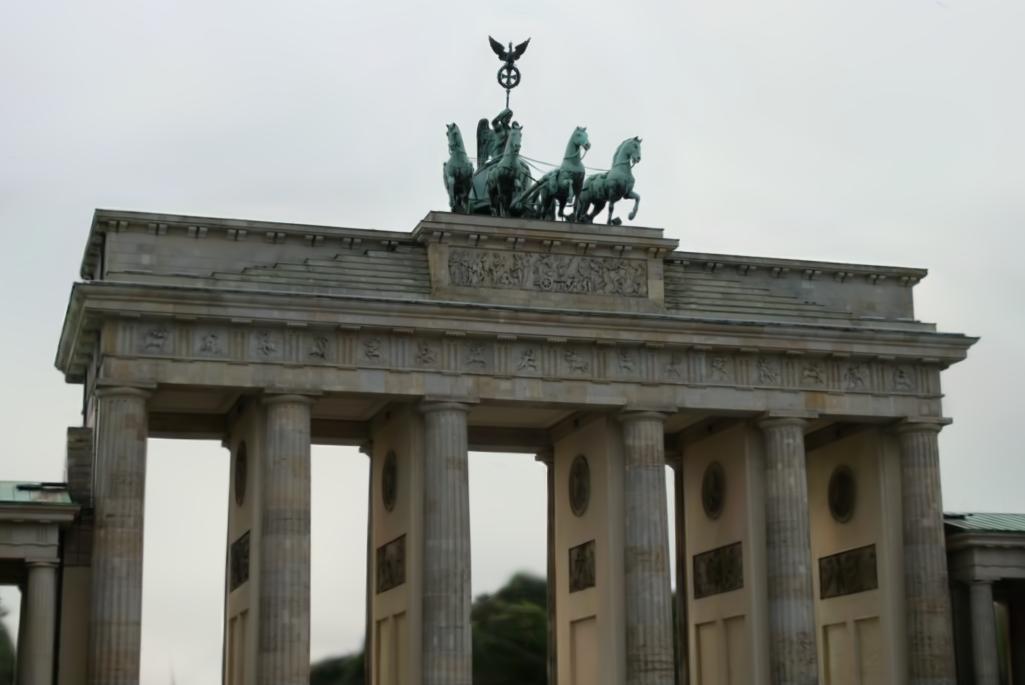}};
        \node [image,right=of img-23] (img-24) {\includegraphics[width=\figurewidth]{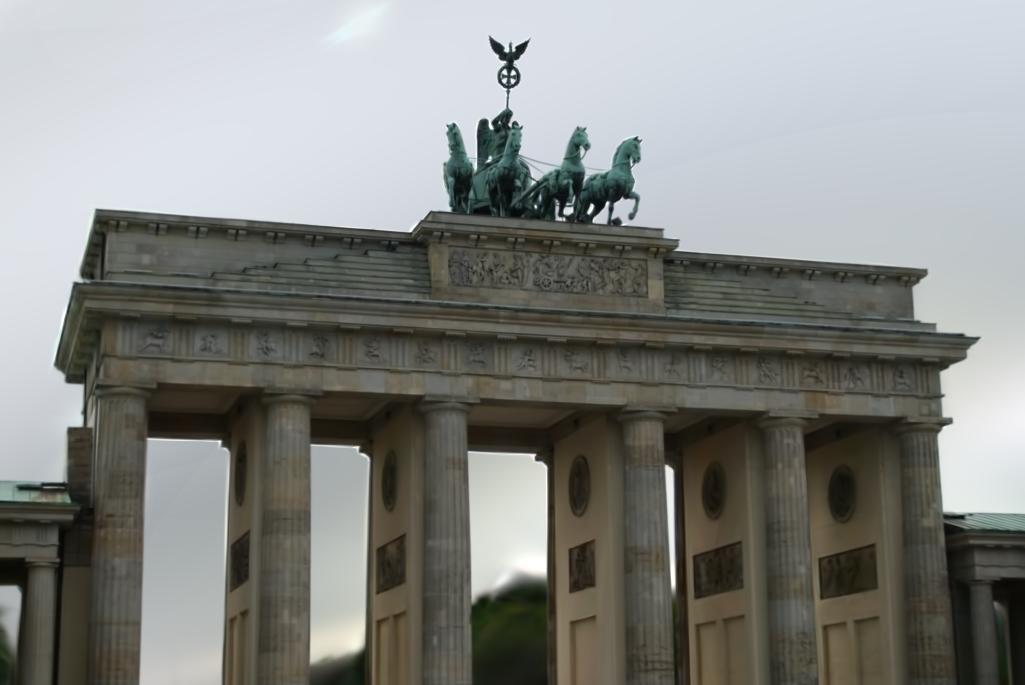}};

        \node[label] (label1) at (img-00.north) {Ground Truth};
        \node[label] (label1) at (img-01.north) {Splatfacto-W};
        \node[label] (label2) at (img-02.north) {WildGaussians};
        \node[label] (label3) at (img-03.north) {\ours (ours) - A};
        \node[label] (label4) at (img-04.north) {\ours (ours)};

        \node[label,rotate=90] (scene1) at (img-00.west) {\itshape Trevi Fountain};
        \node[label,rotate=90] (scene2) at (img-10.west) {\itshape Sacre Coeur};
        \node[label,rotate=90] (scene3) at (img-20.west) {\itshape Brandenburg Gate};
    \end{tikzpicture}
    \caption{\textbf{Qualitative results on the Photo Tourism data set} \cite{snavely2006photo}. including  Our method demonstrates high rendering quality in the \textit{Trevi Fountain}, \textit{Brandenburg Gate}, and \textit{Sacre Coeur} scenes.
    }
    \label{fig:results-phototourism}
\end{figure*}

We base our implementation of \ours on Nerfstudio codebase~\cite{nerfstudio, ye2024gsplat} (\texttt{nerfstudio} version 1.1.4 and \texttt{gsplat} version 1.0.0). 
For a fair comparison of computational efficiency, all ablation experiments are conducted on a single Tesla V100 GPU. Next, we present the hyperparameters used for the data sets. 

\paragraph{RobustNeRF, On-the-go and Mip-NeRF 360} 
We use the same hyperparameter settings for \ours on the scenes from RobustNeRF~\cite{sabour2023robustnerf}, On-the-go~\cite{ren2024nerf} and Mip-NeRF 360~\cite{barron2022mipnerf360} data sets. 
More specifically, we follow the default hyperparameter setting for Splatfacto~\cite{nerfstudio} and train for 30k iterations. Densification of distractor Gaussians stops after 15k iteration.
We initialize $1000$ distractor points in every training image (see \cref{sec:decomposed_3dgs} for details) and initialize the static points using the standard method based on the COLMAP point cloud, as employed in Splatfacto.
All parameters of static and distractor points are optimized using Adam optimizer~\cite{kingma2015adam}.
For the distractor points, we set the learning rates for the quaternions, scales and RGB colours to 0.01, 0.05 and 0.025 respectively, while the learning rates are kept as the Splatfacto default for the means and opacities. The regularization on the alpha-blending weights are set to $\lambda_d = 0.01$ and $\lambda_s=0.01$ in the loss in \cref{eq:loss-total}. 

\paragraph{Photo Tourism} 
We train each scene for 200k iterations. 
We employ the appearance modelling from WildGaussians~\cite{kulhanek2024wildgaussians} which uses an MLP with 2 hidden layers of size 128. 
The per-image embedding $\ve_j \in \bbR^{d_{\ve}}$ has dimension of $d_{\ve} = 32$, while the per-Gaussian embedding $\vz_i \in \bbR^{d_{\vz}}$ has dimension $d_{\vz} = 24$ and is initialized using Fourier frequencies with 4 components. 
All parameters above are optimized using Adam optimizer~\cite{kingma2015adam} with learning rate $0.0005$ for the MLP, $0.001$ for the per-image embeddings $\ve_j$ and $0.005$ for the per-Gaussian embeddings $\vz_i$. 

\begin{table*}[th!]
    \scriptsize
    \centering
    \caption{\textbf{Performance comparison on the Mip-NeRF 360 data set~\cite{barron2022mipnerf360}}. 
    }
    \vspace*{-.1in}
    \resizebox{1.0\linewidth}{!}{%
        \begin{tabular}{l|ccc|ccc|ccc|ccc|ccc|ccc|ccc}
    \toprule
        & \multicolumn{9}{c|}{outdoor} & \multicolumn{12}{c}{indoor}\\
        & \multicolumn{3}{c|}{\textbf{\itshape bicycle}} & \multicolumn{3}{c|}{\textbf{\itshape garden}} & \multicolumn{3}{c|}{\textbf{\itshape stump}} & \multicolumn{3}{c|}{\textbf{\itshape room}} & \multicolumn{3}{c|}{\textbf{\itshape counter}} & \multicolumn{3}{c|}{\textbf{\itshape kitchen}} & \multicolumn{3}{c}{\textbf{\itshape bonsai}}\\
        \textbf{Method} & \textbf{PSNR}$\uparrow$ & \textbf{SSIM}$\uparrow$ & \textbf{LPIPS}$\downarrow$ & \textbf{PSNR}$\uparrow$ & \textbf{SSIM}$\uparrow$ & \textbf{LPIPS}$\downarrow$ & \textbf{PSNR}$\uparrow$ & \textbf{SSIM}$\uparrow$ & \textbf{LPIPS}$\downarrow$ & \textbf{PSNR}$\uparrow$ & \textbf{SSIM}$\uparrow$ & \textbf{LPIPS}$\downarrow$ & \textbf{PSNR}$\uparrow$ & \textbf{SSIM}$\uparrow$ & \textbf{LPIPS}$\downarrow$ & \textbf{PSNR}$\uparrow$ & \textbf{SSIM}$\uparrow$ & \textbf{LPIPS}$\downarrow$ & \textbf{PSNR}$\uparrow$ & \textbf{SSIM}$\uparrow$ & \textbf{LPIPS}$\downarrow$ \\
        \midrule
        Splatfacto & 24.99 & 0.761 & 0.171 & 27.24 & 0.853 & 0.096 & 26.30 & 0.771 & 0.164 & 31.87 & 0.921 & 0.164 & 28.69 & 0.899 & 0.166 & 31.30 & 0.920 & 0.107 & 31.72 & 0.938 & 0.137 \\
        \textbf{\ours} & 24.80 & 0.756 & 0.171 & 27.03 & 0.852 & 0.096 & 25.54 & 0.751 & 0.176 & 31.51 & 0.920 & 0.164 & 28.29 & 0.897 & 0.166 & 30.93 & 0.919 & 0.105 & 31.42 & 0.936 & 0.138 \\
        \bottomrule
    \end{tabular}
    
    }
    \label{tab:mipnerf360_v2}
\end{table*}

For background modelling, we use the background MLP from~\cite{xu2024splatfactow} consisting of 3 hidden layers with 128 hidden units each. The Spherical Harmonics (SH) degree of the background model is set to 4. The learning rates for the background model encoder and the 0th-order SH prediction head are 0.002, while the learning rate for the remaining SH orders is set to 0.0001. The Adam optimizer is used, along with an Exponential decay scheduler with decay rates of 0.0001, 0.0002, and 0.00001 separately. The regularization weights are configured as follows: $\lambda_s = 0$, $\lambda_d = 0.01$, and $\lambda_\text{bg} = 0.15$. The threshold $\mathcal{T_\epsilon}$ for regularization is set to 0.003. All other settings are kept consistent with the parameters used for the On-the-go and RobustNerf data sets. 

We also perform an ablation study using only the appearance embedding model while disabling the background model, which is denoted as \ours-A. In this ablation, the learning rates and MLP structure for the appearance embedding model are kept the same as described above, while the regularization weights are set as follows: $\lambda_s = 0.01$, $\lambda_d = 0.01$, and $\lambda_\text{bg} = 0$.

We follow the evaluation protocol~\cite{martinbrualla2020nerfw, kulhanek2024wildgaussians} and learn per-image appearance embeddings for test images. During evaluation, we train per-image appearance embedding for 128 iterations using the Adam optimizer with a learning rate of 0.01. All other parameters remain frozen during this period. 
We learn the embedding on the left half of the image, and then evaluate metrics on the right half.


\section{Additional Experimental Results}
\label{app:experimental_results}

\subsection{More Qualitative Results}

\paragraph{Comparison on RobustNeRF data set} We present qualitative results for the \textit{Crab~(1)}, \textit{Crab~(2)}, and \textit{Yoda} scenes from the RobustNeRF data set in \cref{fig:qualitative-robustnerf-all}. Our method achieves state-of-the-art results across all three scenes. However, we observe that our occluder removal does not always outperform all baselines. In some cases, where other baselines successfully remove occluders completely, our method occasionally leaves small artefacts. Despite this, our approach excels in reconstructing the background with exceptional clarity. For more details see \cref{fig:qualitative-robustnerf-all}.

\paragraph{Comparison on On-the-go data set} We also compare our method with other baselines on the \textit{Patio}, \textit{Spot}, \textit{Mountain}, and \textit{Fountain} scenes from the On-the-go data set. In the \textit{Patio} scene, our \ours does not perform well, as it fails to remove all occluders effectively. This is due to a person wearing a black jacket standing in one place for a long period, resulting in black artefacts. For detailed analysis, see the failure analysis section \cref{app:failure-cases}. In the \textit{Spot} scene, \ours produces fewer artefacts and reconstructs more detailed background information. As shown in \cref{fig:qualitative-onthego-all}, our method achieves finer reconstruction on the floor, and the wrinkles on the sofa closely resemble the ground truth. In the \textit{Mountain} scene, our method better reconstructs the global colour and texture, especially for the wooden hut. Outside of the zoomed-in areas, white artefacts in the bottom-right corner and black artefacts on the chair, which are visible in SpotlessSplats, are not present in our results. However, our method has difficulty reconstructing the sky, as moving clouds are sometimes incorrectly identified as dynamic elements. More details can be found in \ref{app:failure-cases}. In the \textit{Fountain} scene, while none of the baselines can fully remove artefacts near the trunk, our method successfully reconstructs the column of the background architecture.

We also compare the ability of \ours, SpotLessSplats (using Stable Diffusion features) and WildGaussians (using DINOv2 features) to detect distractors on the On-the-go data set, as shown in \cref{fig:mask}.
Since we do not use masks for training, we follow the same criteria as SLS for rendering the mask, selecting areas where the distractor's opacity is greater than 0.5 as the mask region. Despite the lack of semantic supervision, our method demonstrates a higher capability in rendering accurate masks in certain scenes compared to our baselines.

\paragraph{Comparison on Photo Tourism data set} 
 We evaluate our method on the \textit{Brandenburg Gate}, \textit{Sacre Coeur}, and \textit{Trevi Fountain} scenes from the Photo Tourism data set, comparing it with the baselines. As shown in \cref{tab:phototourism}, after incorporating appearance embeddings, our \ours outperforms some baselines and achieves results close to the state-of-the-art. However, the metrics slightly degrade when incorporating background modelling, which aligns with the results of Splatfacto-W. Removing background Gaussians results in the loss of high-frequency details in the background, which reduces image quality. Moreover, it remains challenging to fully separate the background sky from the foreground, adding complexity to scene reconstruction. The gap between our method and the baselines is smallest for the \textit{Trevi Fountain} scene, as in this scene, the distribution of distractors in the training images is relatively concentrated, making it easier for our method to learn dynamic elements effectively. From \cref{fig:results-phototourism}, our method produces visually appealing results with appearance embeddings, reconstructing fine details and achieving a smoother background compared to Splatfacto-W and WildGaussians.


\paragraph{Comparison on Mip-NeRF 360 data set}
The results on the Mip-NeRF 360 dataset demonstrate that \ours can be applied to scenes without distractors due to its adaptive density control of distractor points.
As shown in \cref{tab:mipnerf360_v2}, our method performs comparably to Splatfacto across all scenes in the Mip-NeRF 360 dataset, with only a slight performance drop. This decline is primarily caused by the complexity of certain textures, such as sharp light variations on grass, reflections on the metal bowl in the \textit{counter} scene, and fine details like flower petals in the \textit{bonsai} scene. These factors make it more challenging to distinguish distractors from static objects.

\subsection{Computational Efficiency}
\label{app:ablation}

In \cref{tab:memory-ablation-size} and \cref{tab:memory-ablation-time}, we compare the memory usage (MB) and training time of \ours with Splatfacto~\cite{nerfstudio}, Splatfacto-W~\cite{xu2024splatfactow}, SLS-mlp, SLS-mlp with utilization-based pruning (UBP) \cite{sabour2024spotlesssplats}, and WildGaussians \cite{kulhanek2024wildgaussians}. We also report the memory usage required to store the distractor Gaussians for \ours. 
Our rendering speed (FPS) is compared with that of Splatfacto, as reported in \cref{tab:memory-ablation-fps} using \texttt{ns-eval} in Nerfstudio~\cite{nerfstudio}, which calculates FPS based on both rendering time and evaluation time (including PSNR, SSIM, and LPIPS). We observe that \ours achieves a rendering speed comparable to Splatfacto, as \ours remains a purely splatting-based method. 
Since other baselines, such as SLS and WildGaussians, use different frameworks or rasterization libraries, it is difficult to make direct comparisons in terms of rendering speed.
While \ours introduces a slight memory overhead compared to Splatfacto, only the static Gaussians need to be retained for novel view synthesis. Training time increases by a few minutes due to the additional memory usage. 
Finally, we observe that \ours achieves a rendering speed similar to Splatfacto, is more memory-efficient, and requires less training time than other distractor-free methods.

\begin{table*}[h!]
    \centering
    \small
    \vspace*{-4pt}
    \caption{\textbf{Comparison of memory size.} Comparison of memory size (MB) for \ours, Splatfacto, Splatfacto-W, SLS-mlp, and WildGaussians on the RobustNeRF~\cite{sabour2023robustnerf} and On-the-go~\cite{ren2024nerf} data sets. To ensure consistency, we use Splatfacto-W-light with all functions enabled.}
    \vspace{-3mm}
    \resizebox{1.0\linewidth}{!}{
    \begin{tabular}{l|ccccc|cccccc|c}
        \toprule
        &\multicolumn{5}{c|}{\textbf{RobustNeRF}} &\multicolumn{6}{c|}{\textbf{On-the-go}} & Average\\
        \textbf{Method} & \textbf{\textit{Statue}} & \textbf{\textit{Android}} & \textbf{\textit{Crab~(1)}} & \textbf{\textit{Crab~(2)}} & \textbf{\textit{BabyYoda}} & \textbf{\textit{Mountain}} & \textbf{\textit{Fountain}} & \textbf{\textit{Corner}} & \textbf{\textit{Patio}} & \textbf{\textit{Spot}} & \textbf{\textit{Patio-High}} & \\
        \midrule
        Splatfacto \cite{nerfstudio} & 38.19 & 51.95 & 25.38 & 26.86 & 24.19 & 123.63 & 145.82 & 67.45 & 73.81 & 70.76 & 106.83 & 68.62\\
        Splatfacto-W$^*$ \cite{xu2024splatfactow} & 42.86 & 53.49 & 40.13 & 37.57 & 23.95 & 139.90 & 165.23 & 64.01 & 74.81 & 85.04 & 114.11 & 76.46\\
        SLS-mlp \cite{sabour2024spotlesssplats} & 110.14 & 129.38 & 34.41 & 47.40 & 54.50 & 108.85 & 264.34 & 142.25 & 128.09 & 107.20 & 192.79 & 119.94\\
        SLS-mlp (+UBP) \cite{sabour2024spotlesssplats} & 42.24 & 56.79 & 15.11 & 18.90 & 23.59 & 53.48 & 109.61 & 45.70 & 58.06 & 39.56 & 60.45 & 47.59 \\
        WildGaussians \cite{kulhanek2024wildgaussians} & - & - & - & - & - & 215.64 & 480.52 & 232.98 & 140.14 & 124.37 & 230.31 & 237.33\\
        \midrule
        \textbf{\ours (static)} & 46.99 & 67.80 & 35.89 & 39.40 & 38.75 & 118.27 & 163.73 & 67.03 & 55.69 & 61.28 & 102.52 & 72.49\\ 
        \textbf{\ours (distractors)} & 16.82 & 18.15 & 6.35 & 7.23 & 11.24 & 23.26 & 12.51 & 21.03 & 38.51 & 40.33 & 35.65 & 21.01\\
        \bottomrule
    \end{tabular}%
    }
    \label{tab:memory-ablation-size}
\end{table*}

\begin{table*}[h!]
    \centering
    \small
    \vspace*{-4pt}
    \caption{\textbf{Comparison of training time.} Comparison of training time for \ours, Splatfacto, Splatfacto-W, SLS-mlp, and WildGaussians on the RobustNeRF~\cite{sabour2023robustnerf} and On-the-go~\cite{ren2024nerf} data sets. To ensure consistency, we use Splatfacto-W-light with all functions enabled.}
    \vspace{-3mm}
    \resizebox{1.0\linewidth}{!}{
    \begin{tabular}{l|ccccc|cccccc|c}
        \toprule
        &\multicolumn{5}{c|}{\textbf{RobustNeRF}} &\multicolumn{6}{c|}{\textbf{On-the-go}} & Average\\
        \textbf{Method} & \textbf{\textit{Statue}} & \textbf{\textit{Android}} & \textbf{\textit{Crab~(1)}} & \textbf{\textit{Crab~(2)}} & \textbf{\textit{BabyYoda}} & \textbf{\textit{Mountain}} & \textbf{\textit{Fountain}} & \textbf{\textit{Corner}} & \textbf{\textit{Patio}} & \textbf{\textit{Spot}} & \textbf{\textit{Patio-High}} & \\
        \midrule
        Splatfacto \cite{nerfstudio} & 9:32 & 11:42 & 9:02 & 10:43 & 10:17 & 9:46 & 10:34 & 8:56 & 8:55 & 9:02 & 11:11 & 9:58 \\
        Splatfacto-W$^*$ \cite{xu2024splatfactow} & 24:25 & 24:08 & 23:44 & 25:26 & 25:24 & 24:13 & 24:53 & 15:59 & 17:38 & 19:05 & 20:44 & 22:19 \\
        SLS-mlp \cite{sabour2024spotlesssplats}& 25:21 & 24:52 & 21:57 & 25:33 & 25:02 & 27:39 & 29:06 & 22:47 & 19:31 & 22:16 & 21:41 & 24:09\\
        SLS-mlp (+UBP) \cite{sabour2024spotlesssplats} & 24:09 & 23:13 & 21:59 & 25:16 & 25:55 & 26:22 & 27:14 & 19:54 & 18:16 & 19:56 & 21:08 & 23:02\\
        WildGaussians \cite{kulhanek2024wildgaussians} & - & - & - & - & - & 58:06 & 1:20:30 & 58:41 & 52:10 & 1:00:46 & 58:49 & 1:01:30 \\
        \midrule
        \textbf{\ours (ours)} & 13:40 & 13:33 & 13:03 & 14:01 & 13:26 & 11:36 & 14:20 & 10:39 & 12:11 & 12:43 & 13:10 & 12:56 \\ 
        \bottomrule
    \end{tabular}%
    }
    \label{tab:memory-ablation-time}
\end{table*}

\begin{table*}[h!]
    \centering
    \small
    \vspace*{-4pt}
    \caption{\textbf{Rendering Speed.} Comparison of rendering speed (FPS) between \ours and Splatfacto.}
    \vspace{-3mm}
    \resizebox{1.0\linewidth}{!}{
    \begin{tabular}{l|ccccc|cccccc|c}
        \toprule
        \textbf{Method} &\multicolumn{5}{c|}{\textbf{RobustNeRF}} &\multicolumn{6}{c|}{\textbf{On-the-go}} & Average\\
        \midrule
        Splatfacto \cite{nerfstudio} & \phantom{1}91.04 & \phantom{1}72.98 & 102.85 & 103.03 & \phantom{1}99.83 & 102.87 & 90.32 & 101.58 & 101.60 & 89.85 & \phantom{1}79.13 & \phantom{1}94.01\\
        \textbf{\ours} & 104.98 & 101.86 & 109.16 & 112.02 & 114.11 & 110.16 & 89.12 & 136.13 & 108.24 & 96.73 & 101.16 & 107.61\\
        \bottomrule
    \end{tabular}%
    }
    \label{tab:memory-ablation-fps}
\end{table*}

\begin{figure*}[t]
    \centering
    \setlength{\figurewidth}{0.20\textwidth}
    \begin{tikzpicture}[
        image/.style = {inner sep=0pt, outer sep=1pt, minimum width=\figurewidth, anchor=north west, text width=\figurewidth}, 
        node distance = 1pt and 1pt, every node/.style={font= {\tiny}}, 
        label/.style = {font={\footnotesize\bf\vphantom{p}},anchor=south,inner sep=0pt},
        highlight1/.style = {draw=orange, thick, rectangle, minimum width=0.25\figurewidth, minimum height=0.15\figurewidth},
        highlight2/.style = {draw=orange, thick, rectangle, minimum width=0.3\figurewidth, minimum height=0.35\figurewidth}
    ]

        \node [image] (img-00) {\includegraphics[width=\figurewidth]{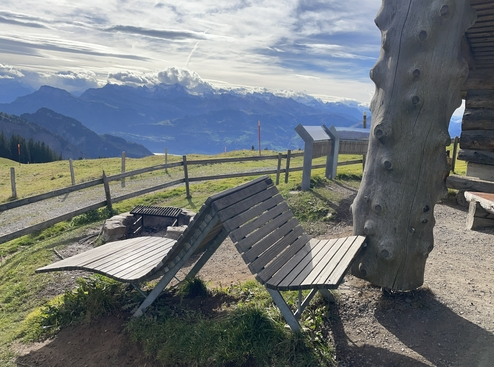}};
        \node [image, right=0.2cm of img-00] (img-01) {\includegraphics[width=\figurewidth]{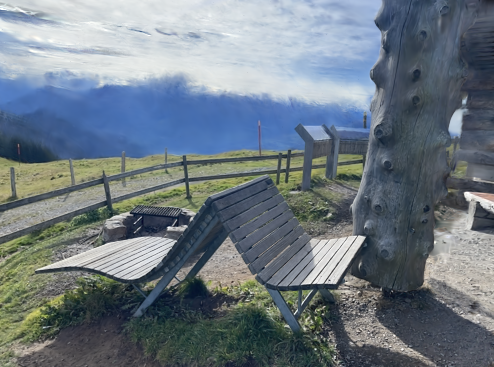}};
        \node [image, right=0.2cm of img-01] (img-02) {\includegraphics[width=\figurewidth]{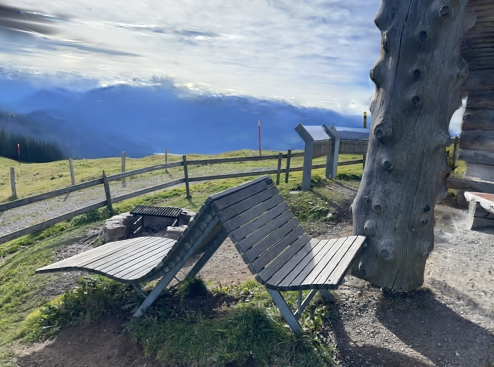}};

        \node[highlight1] at ($(img-00.west) + (0.15\figurewidth, 0.29\figurewidth)$) {};
        \node[highlight1] at ($(img-01.west) + (0.15\figurewidth, 0.29\figurewidth)$) {};
        \node[highlight1] at ($(img-02.west) + (0.15\figurewidth, 0.29\figurewidth)$) {};
        
        \node [image, below=0.2cm of img-00] (img-10) {\includegraphics[width=\figurewidth]{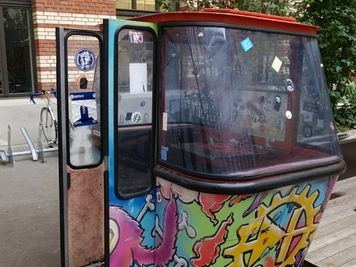}};
        \node [image, right=0.2cm of img-10] (img-11) {\includegraphics[width=\figurewidth]{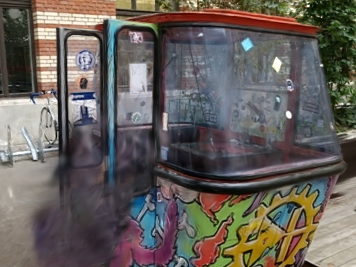}};
        \node [image, right=0.2cm of img-11] (img-12) {\includegraphics[width=\figurewidth]{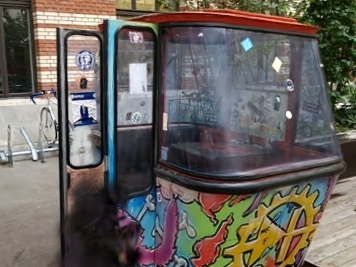}};

        \node[highlight2] at ($(img-10.west) + (0.28\figurewidth, -0.19\figurewidth)$) {};
        \node[highlight2] at ($(img-11.west) + (0.28\figurewidth, -0.19\figurewidth)$) {};
        \node[highlight2] at ($(img-12.west) + (0.28\figurewidth, -0.19\figurewidth)$) {};
        
        \node[label,rotate=90] (scene1) at (img-00.west) {\itshape Mountain};
        \node[label,rotate=90] (scene2) at (img-10.west) {\itshape Patio};
        
        \node[label] (label1) at (img-00.north) {Ground Truth};
        \node[label] (label1) at (img-01.north) {Splatfacto \cite{nerfstudio}};
        \node[label] (label2) at (img-02.north) {Ours};
        
    \end{tikzpicture}
    \caption{\textbf{Failed Cases in the \textit{Mountain} and \textit{Patio} Scenes.} The upper-left corner of the \textit{Mountain} scene reveals a visible hole, while the black artifacts in the \textit{Patio} scene appear denser compared with Splatfacto.
    } 
    \label{fig:failedcase}
\end{figure*}

\subsection{Failure Cases}
\label{app:failure-cases}

Our \ours performs well, particularly in scenarios with minimal changes in lighting and weather, and where occluders vary across camera views, such as in the \textit{Yoda} and \textit{Crab} scenes. 
However, its performance decreases in outdoor data sets with complex lighting and weather variations, because without the help of appearance embedding, the occluders may also explain part of the background. For example, in \textit{Mountain} scene, the cloud in sky is not identical in every frame, leading some of the distractor Gaussians to explain the cloud and sky. Another limitation arises when occluders persist across many frames. Since our method separates distractors based on photometric inconsistencies between views, occluders that remain in the same or similar positions across multiple frames are misclassified as static objects. A notable example is the \textit{Patio} scene. Consequently,  as illustrated in \cref{fig:failedcase} for reference.

\section{Future Directions}
\label{app:future-directions}

Investigating how to combine appearance modelling with the decomposed Gaussians for separation and better controllability is a key next step to improve \ours's applicability for large-scale 3D reconstruction tasks~\cite{snavely2006photo}.   
Since \ours is a pure splatting method, an interesting future direction is to incorporate semantic features from foundation models for improving the separation between static objects and distractors, which has shown to effectively remove distractors in 3DGS~\cite{kulhanek2024wildgaussians,sabour2024spotlesssplats}. 
Additionally, informing the initialization step of the distractor Gaussians with plausible spatial locations and shapes of the occluders, and assigning Gaussian sets per occluder could potentially yield a more fine-grained scene decomposition. 
Finally, extending \ours to handle dynamics in videos is an interesting future direction, since \ours may confuse distractors as being static if the distractor do not move significantly across the views which is typical for videos. 
We hope that this work can spur more research for explicit 3D scene decomposition based on 3DGS. 

\end{document}